\documentclass{article}

 \usepackage[main, final]{neurips_2026}

\usepackage[utf8]{inputenc} 
\usepackage[T1]{fontenc}    
\usepackage{hyperref}       
\usepackage{url}            
\usepackage{booktabs}       
\usepackage{amsfonts}       
\usepackage{nicefrac}       
\usepackage{microtype}      
\usepackage{xcolor}         
\usepackage{amsmath,amssymb,amsthm}
\usepackage{mathtools}
\usepackage{geometry}
\usepackage{enumitem}
\usepackage{booktabs}

\usepackage{multirow}    
\usepackage{array}       
\usepackage{colortbl}    
\usepackage{amsmath}     
\usepackage{caption}     
\usepackage[T1]{fontenc} 
\usepackage{fontawesome5}
\usepackage{graphicx}
\usepackage[T1]{fontenc}
\usepackage[dvipsnames]{xcolor}  

\usepackage{algorithm}
\usepackage{algpseudocode}
\usepackage{xcolor}

\algrenewcommand\algorithmiccomment[1]{\hfill\textcolor{teal!70!black}{\textit{\(\triangleright\) #1}}}

\makeatletter
\renewcommand{\ALG@beginalgorithmic}{\small}
\makeatother

\definecolor{ggreen}{RGB}{0,130,65}
\definecolor{rred}{RGB}{198,30,30}
\definecolor{rowA}{RGB}{250,250,250}
\definecolor{rowB}{RGB}{255,255,255}
\definecolor{basecol}{RGB}{235,240,248}

\newcommand{\gd}[1]{{\color{ggreen}#1}}
\newcommand{\rd}[1]{{\color{rred}#1}}

\theoremstyle{plain}

\theoremstyle{definition}

\theoremstyle{remark}

\usepackage{tcolorbox}

\newtcolorbox{notebox}{
    colback=limecolor!10,
    colframe=black,
    boxrule=0.3pt,
    arc=0pt,
    left=3pt, right=3pt, top=1.5pt, bottom=1.5pt,
    boxsep=1pt,
    before skip=2pt,
    after skip=2pt,
    fontupper=\small
}
\tcbuselibrary{theorems, skins}

\newtcolorbox{theorembox}[1][]{
    colback=gray!5,
    colframe=black,
    boxrule=0.4pt,
    arc=0pt,
    left=3pt, right=3pt, top=1.5pt, bottom=1.5pt,
    boxsep=2pt,
    before skip=2pt,
    after skip=2pt,
    fonttitle=\small,
    title=#1,
    fontupper=\small
}

\makeatletter
\newcommand{\customlabel}[2]{%
    \def\@currentlabel{#2}%
    \label{#1}%
}

\renewcommand{\arraystretch}{1.28}
\title{Where You Backpropagate Matters: Token Hypothesis for Memory-Efficient Fine-Tuning}

\author{
  Md Kowsher$^{1,2}$\thanks{This work was not conducted at Meta.} \qquad Chen Chen$^{2}$ \\
  $^{1}$Meta, \quad $^{2}$University of Central Florida \\[6pt]
  \faGlobe\ \href{https://kowsher.github.io/Token-Hypothesis/}{Project Page}
  \qquad
  \faGithub\ \href{https://github.com/Kowsher/Token-Hypothesis}{Code}
}

\begin{document}

\maketitle
\begin{abstract}
Fine-tuning large transformers routes gradients through every token position, paying the cost in activation memory that scales linearly with sequence length and dominates the GPU budget. Existing efficient methods reduce the parameter side---low-rank adapters, prefix tuning, weight-decomposed updates---but leave the activation footprint untouched. We present the Token Hypothesis, a memory-efficient fine-tuning procedure motivated by a Neural Tangent Kernel analysis of how fine-tuning signal is distributed across token positions. We prove that every position is a low-dimensional capacity bottleneck whose width does not grow with the model, that positions with aligned hidden-state geometry are interchangeable in bidirectional models, and that causal positions form a monotone capacity chain in which later positions strictly dominate earlier ones. The analysis provides an assumption-qualified criterion for estimating how many positions are needed and which ones, using quantities computable from forward passes alone: in bidirectional models, small arbitrary subsets are predicted to suffice when position-wise kernels are aligned, while in causal models, later positions are predicted to carry the richest fine-tuning signal. We translate this into a procedure that backpropagates through only a small subset of positions while running the full forward pass over the rest. Empirically, the procedure matches or exceeds full-sequence fine-tuning in nearly all settings across six task families---commonsense reasoning, mathematical reasoning, visual question answering, video question answering, GLUE, and image classification---and five backbones spanning bidirectional and causal architectures. It reduces activation memory substantially, translating to up to $\sim 19\%$ lower total GPU memory ($\sim 31$~GB) on visual question answering while exceeding full-sequence accuracy, and composes cleanly with parameter-efficient methods such as LoRA and DoRA.

\end{abstract}

\section{Introduction}

\begin{figure}[ht!]
\centering
\includegraphics[width=\linewidth]{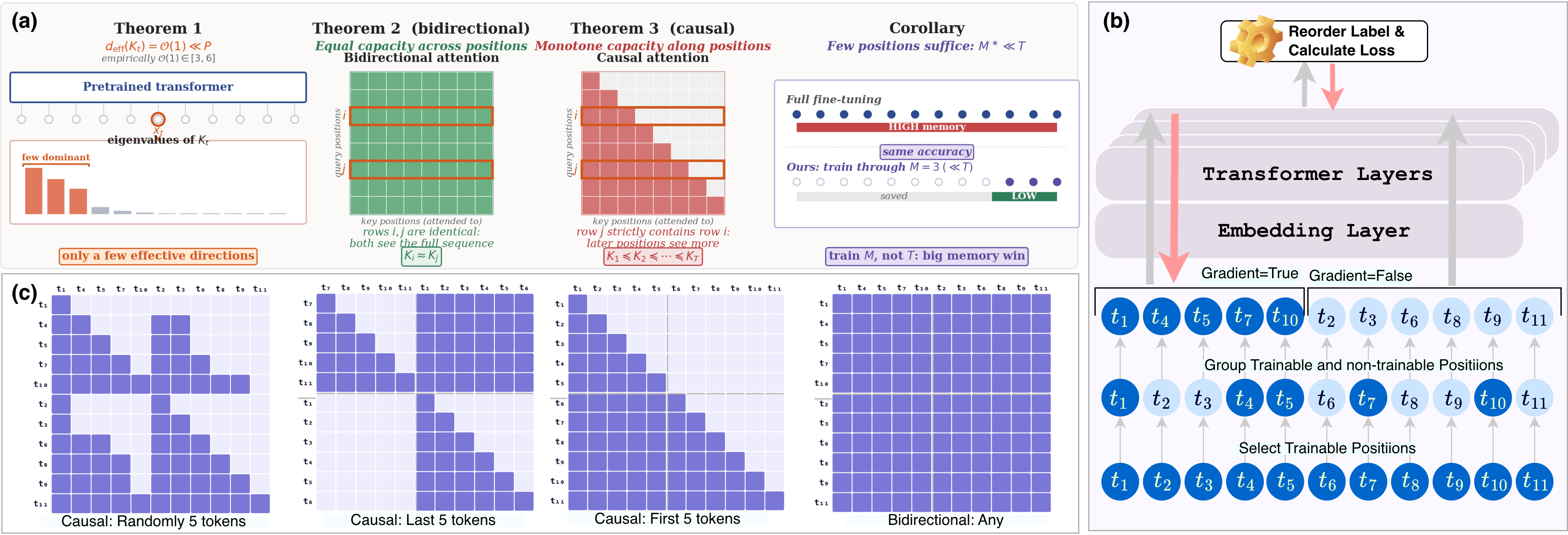}
\caption{\textbf{The Token Hypothesis and our training procedure.}
\textbf{(a)} Three theorems and a corollary characterize how fine-tuning signal is distributed across token positions: capacity per position is bounded, equal across positions in bidirectional models, and monotonically increasing in causal models---so a small subset $M \ll T$ of positions suffices to match full fine-tuning.
\textbf{(b)} We select $M$ trainable positions, group them against the non-trainable ones, and reorder the sequence so that only the trainable group receives gradients while the rest pass forward without storing activations.
\textbf{(c)} Attention masks for four selection strategies, shown after reordering tokens so trainable positions are contiguous:
\emph{(i) Random} selects $M$ positions uniformly at random each epoch;
\emph{(ii) Last-$M$} selects the final $M$ positions of the sequence (optimal for causal models);
\emph{(iii) First-$M$} selects the initial $M$ positions (baseline for causal models);
\emph{(iv) Any subset} selects any $M$ positions in bidirectional models, where all positions are interchangeable.
Darker rows mark trainable queries.}

\label{fig:token_hypothesis}
\end{figure}

Fine-tuning a pretrained transformer backpropagates the loss through every token position of every training sequence~\cite{radford2018improving}. The resulting memory cost has three sources: \textbf{(i)} model parameters, \textbf{(ii)} gradients and optimizer states, and \textbf{(iii)} intermediate activations~\cite{zeng2026tokenseek}. Most prior work on parameter-efficient fine-tuning targets the first two through low-rank adapters~\cite{hu2022lora}, prefix tuning~\cite{li2021prefix}, and related techniques~\cite{kowsher2025rocoft}, yet activations dominate GPU memory in practice~\cite{zeng2026tokenseek, simoulin2024memory}. Activations scale linearly with sequence length $T$: fine-tuning an 8B-parameter model at $T=2048$ consumes over $100$~GB of activation memory alone.

A natural way to reduce this cost is to backpropagate through only a subset of token positions, and recent methods~\cite{simoulin2024memory, zeng2026tokenseek, qin2025sstoken} have pursued increasingly sophisticated criteria for \emph{which} tokens to select. We argue this framing skips the prior question: what determines a single position's capacity to drive fine-tuning, and how does that capacity depend on architecture? Once these are answered, both \emph{how many} and \emph{which} positions to train follow directly. We answer both theoretically and empirically through what we call the \emph{Token Hypothesis} through a Neural Tangent Kernel (NTK) analysis. Training through a single token position confines every parameter update to a low-dimensional subspace whose size is set at initialization and does not grow with the model. Concretely, let $\theta \in \mathbb{R}^P$ denote the pretrained parameters initialized at $\theta_0$, and let $o_t(x; \theta)$ denote the model output at position $t$ for input $x$. Given a fine-tuning dataset $\mathcal{D} = \{(x^{(n)}, y^{(n)})\}_{n=1}^{N}$, define the position-$t$ feature map $\phi_t(x) := \nabla_\theta o_t(x; \theta_0) \in \mathbb{R}^P$ and Gram matrix $K_t \in \mathbb{R}^{N \times N}$ with $K_t[n,m] := \langle \phi_t(x^{(n)}), \phi_t(x^{(m)}) \rangle$. Fine-tuning through position $t$ alone then confines updates to the span of $\{\phi_t(x^{(n)})\}_{n=1}^{N}$, a subspace of dimension at most $\mathrm{rank}(K_t) \leq N$, independent of $P$. The effective rank $d_{\mathrm{eff}}(K_t) := \mathrm{tr}(K_t) / \|K_t\|_{\mathrm{op}}$, which counts how many gradient directions actually carry signal, is tighter still: across RoBERTa-base, SmolLM2-135M, and LLaMA-3-8B, we measure $d_{\mathrm{eff}}(K_t) \in [1.5, 5.2]$ at every tested position and dataset size (Figures~\ref{fig:thm1}, \ref{fig:app-thm1-smoll}, \ref{fig:app-thm1-llama}). This single-position bottleneck raises three questions, which we resolve in turn.

\emph{Are all bottlenecks equally wide?} In bidirectional models, two positions $i$ and $j$ that hold fixed tokens across samples and share aligned hidden-state geometry produce nearly identical fine-tuned predictors, with the gap controlled by quantities computable from forward passes alone (Theorem~\ref{thm:equal}). Choosing among such positions therefore reduces to sampling. \emph{What about positions with different information access?} In a causal model, under a contraction condition relating adjacent hidden states, the position-specific Gram matrices form a monotone chain $K_1 \preceq K_2 \preceq \cdots \preceq K_T$ (up to a vanishing slack term), so later positions carry at least as rich a gradient signal as earlier ones (Theorem~\ref{thm:monotone}). The natural trainable subset is therefore the last $M$ positions. \emph{How many positions are needed?} The minimum number $M^{\ast}$ of positions whose combined Gram matrix spans the same image as the full sequence is governed by the architecture (Corollary~\ref{cor:optimal-M}). In bidirectional models, any small subset works whenever the position-wise Gram matrices are close in operator norm. In causal models the answer depends on the ordering: under strict prefix inclusion, the last single position spans the entire prediction space, while the forward ordering requires all $T$. Training the last $M$ positions is therefore both architecturally principled and provably efficient. Every quantity in these bounds is measurable from the pretrained model with no fine-tuning (Figure~\ref{fig:token_hypothesis}(a)).

These results yield a memory-efficient fine-tuning procedure. Given a sequence of length $T$ and a target $M \leq T$, we select a subset $\mathcal{S} \subseteq \{1, \ldots, T\}$ of size $M$---any subset for bidirectional models, the last $M$ positions for causal models---and backpropagate only through $\mathcal{S}$ while running the full forward pass over all $T$ positions. Positions outside $\mathcal{S}$ are evaluated inside a \texttt{no\_grad} context (Algorithm \ref{alg:wrapper}), so their activations are never materialized. Peak activation memory is reduced by a fraction $1 - M/T$, the forward representation is unchanged, and inference is unaffected. The procedure composes with LoRA~\cite{hu2022lora} and DoRA~\cite{liu2024dora}, adding activation-memory savings on top of the gradient and optimizer-state savings these methods already provide (Figure~\ref{fig:token_hypothesis}(b)).

Empirically, a small budget of trainable positions---$M{=}25$--$100$ for bidirectional models and $M{=}100$--$500$ for causal models---matches or exceeds full-sequence fine-tuning in nearly all settings across commonsense reasoning, mathematical reasoning, visual question answering, video reasoning, GLUE, and image classification, spanning RoBERTa-base, LLaMA-3-8B, Qwen3-VL-8B, LLaVA-OneVision-7B, and ViT-L/16. The empirical optimum lies above the worst-case $M^{\ast}$ from Corollary~\ref{cor:optimal-M}, reflecting a practical preference for stable optimization over the minimal spanning subset.

\textbf{Contributions.}
\textbf{(i)} We show that each token position is a capacity bottleneck with $\mathrm{rank}(K_t) \leq \min(N,P)$, independent of parameter count, and empirically observe $d_{\mathrm{eff}}(K_t) \in [1.5, 5.2]$ across tested models.
\textbf{(ii)} We establish equal-capacity behavior for aligned bidirectional positions and monotone capacity for causal positions, yielding architecture-dependent selection rules.
\textbf{(iii)} We characterize $M^{\ast}$ as the smallest position budget whose combined Gram matrix spans the full prediction space, with all required quantities computable from forward passes.
\textbf{(iv)} We turn this analysis into a fine-tuning procedure that matches or exceeds full-sequence fine-tuning across six task families and five architectures, while substantially reducing activation memory and lowering total GPU memory by up to $19\%$ in measured long-context settings.

\textbf{Related Work.} We briefly discuss related work here and provide a detailed literature review in Appendix~\ref{app:related-work}. Most prior work on memory-efficient fine-tuning either reduces the number of trainable parameters~\cite{hu2022lora, liu2024dora, li2021prefix, kowsher2025rocoft}, reversibility, or quantization~\cite{chen2016training, gomez2017reversible, dettmers2023qlora, luo2025streambp}, leaving the number of positions in the backward graph untouched. The line closest to ours selects a subset of token positions to backpropagate through~\cite{simoulin2024memory, zeng2026tokenseek}, but does so heuristically---random selection or attention-and-gradient scoring with a mandatory pre-stage and per-model hyperparameters---without explaining why a small subset suffices or how the optimal size depends on architecture. We use the NTK regime~\cite{jacot2018neural, malladi2023kernel, afzal2026linearization} as the theoretical lens for our analysis in large pretrained models. Within this regime, we ask a question prior work has not: how does the fine-tuning capacity of a \emph{single token position} depend on architecture, and how few positions suffice to span the same achievable prediction space as the full sequence? The answer turns token selection from a theoretically  heuristic into an empirical recipe.

\section{Preliminaries}
\label{sec:prelim}

Let $\small f_\theta : \mathcal{V}^T \to \mathbb{R}^{T \times C}$ be a large pretrained transformer with vocabulary $\mathcal{V}$ and $D$ layers, producing per-position outputs $o_t(x; \theta) := f_\theta(x)_t \in \mathbb{R}^C$ from token and positional embeddings $h_t^{(0)}(x) = E_{\mathrm{tok}}[x_t] + E_{\mathrm{pos}}[t]$. In a \emph{bidirectional} model $o_t$ depends on all positions; in a \emph{causal} model it depends only on the prefix $(x_1, \ldots, x_t)$. For a subset $\mathcal{S} \subseteq \{1, \ldots, T\}$ of \emph{trainable positions}, only positions in $\mathcal{S}$ contribute to parameter updates through the masked loss
\begin{equation*}
\label{eq:masked-loss}
\mathcal{L}_\mathcal{S}(\theta) \;:=\; \frac{1}{|\mathcal{S}|\, N} \sum_{n=1}^{N} \sum_{t \in \mathcal{S}} \ell\bigl(o_t(x^{(n)};\theta),\, y^{(n)}\bigr),
\end{equation*}
with $\mathcal{L}_t$ denoting the single-position case $\mathcal{S} = \{t\}$. The gradient $\nabla_\theta \mathcal{L}_\mathcal{S}$ decomposes as a sum of position-wise terms over $t \in \mathcal{S}$, so only activations on backward paths from $t \in \mathcal{S}$ need to be stored---this is the source of the memory savings (\S~\ref{sec:method}).

Our theoretical analysis is carried out in the NTK regime for fine-tuning large-scale models around $\theta_0$~\cite{afzal2026linearization, malladi2023kernel}, under which gradient descent on $\mathcal{L}_t$ is equivalent to kernel regression with kernel function $\mathcal{K}_t(x, x') := \langle \phi_t(x), \phi_t(x') \rangle$. We empirically verify that this regime holds for the models we study: mean weight drift stays below $0.011$ across every parameter group of LLaMA~3-8B during fine-tuning (Figure~\ref{fig:ablations}(d)), justifying the use of $\theta_0$-Jacobians throughout. The corresponding Gram matrix $K_t \in \mathbb{R}^{N \times N}$ on $\mathcal{D}$ has entries $K_t[n,m] := \mathcal{K}_t(x^{(n)}, x^{(m)})$. We use $\mathcal{K}_t$ for the kernel function and $K_t$ for its Gram matrix throughout.

\section{The Token Hypothesis}
\label{sec:theory}

We formalize the Token Hypothesis through three theorems characterizing the fine-tuning capacity of individual token positions under the NTK regime: every position is a bounded capacity bottleneck (Theorem~\ref{thm:capacity}), bidirectional positions with shared information access yield nearly identical predictors (Theorem~\ref{thm:equal}), and in causal models, position-wise capacity is monotone (Theorem~\ref{thm:monotone}). We use the NTK analysis as a predictive design model rather than an exact description of fine-tuning dynamics: it suggests architecture-dependent selection rules---small arbitrary subsets for bidirectional models with aligned position-wise kernels, and last-$M$ subsets for causal models---that we test empirically in Section~\ref{sec:experiments}. The theorems characterize capacity, not optimization dynamics, so the empirical choice of $M$ reflects optimization and stability considerations on top of the predicted capacity structure. Full proofs and additional empirical validation are in the appendix.

\textbf{Each Position Has Limited Capacity.}
\label{sec:thm1} Recall from Section~\ref{sec:prelim} that fine-tuning through position $t$ alone under the NTK regime is equivalent to kernel regression with kernel $K_t$. Every gradient update is therefore a linear combination of the features $\{\phi_t(x^{(n)})\}_{n=1}^{N}$, which confines all parameter changes to the subspace
$\mathcal{G}_t := \mathrm{span}\{\phi_t(x^{(1)}), \ldots, \phi_t(x^{(N)})\} \subseteq \mathbb{R}^P$,
whose dimension is at most $\mathrm{rank}(K_t) \leq \min(N, P)$. The ceiling $\min(N,P)$ is the standard representer bound for kernel regression on $N$ samples and is loose whenever the spectrum of $K_t$ is concentrated on a few directions. A tighter quantity is the \emph{effective rank}
$d_{\mathrm{eff}}(K_t) := \mathrm{tr}(K_t) / \|K_t\|_{\mathrm{op}} \in [1, \mathrm{rank}(K_t)]$,
which counts how many gradient directions actually carry signal during training. The following theorem makes the underlying subspace structure precise.

\begin{theorembox}[Theorem 1: Limited capacity per position] \customlabel{thm:capacity}{1}
\setlength{\abovedisplayskip}{1pt}\setlength{\belowdisplayskip}{1pt}%
\setlength{\abovedisplayshortskip}{1pt}\setlength{\belowdisplayshortskip}{1pt}%
Under the NTK regime, the set of predictions achievable by training through position $t$ on $\mathcal{D} = \{(x^{(n)}, y^{(n)})\}_{n=1}^{N}$ is the affine subspace
\begin{equation*}
\bigl\{\, f_t(X; y) : y \in \mathbb{R}^N \,\bigr\} \;=\; o_t(X; \theta_0) + \mathrm{Im}(K_t),
\end{equation*}
where $X := (x^{(1)}, \ldots, x^{(N)})$ and $\mathrm{Im}(K_t)$ is the column space of $K_t$, with $\mathrm{rank}(K_t) \leq \min(N, P)$. The learned correction takes the representer form
\begin{equation*}
g_t(x) = \sum_{n=1}^{N} \alpha_n\, \mathcal{K}_t(x, x^{(n)}), \qquad \alpha = K_t^{+}\bigl(y - o_t(X; \theta_0)\bigr),
\end{equation*}
lying in the reproducing kernel Hilbert space (RKHS) $\mathcal{H}_{\mathcal{K}_t}$.
\end{theorembox}

The proof, given in Appendix~\ref{app:proof-capacity}, proceeds by showing that gradient descent on $\mathcal{L}_t$ is equivalent to kernel gradient flow in $\mathcal{H}_{K_t}$, whose minimum-norm solution is the representer expansion above. The theorem is position-specific through the feature map $\phi_t = \nabla_\theta o_t(\cdot;\theta_0)$: different positions induce different kernels, and Theorems~\ref{thm:equal} and~\ref{thm:monotone} characterize how $K_t$ varies with $t$ across architectures. Empirically, $d_{\mathrm{eff}}(K_t)$ stays within $[1.5, 5.2]$ across every tested position, dataset size, and model scale (Section~\ref{exp:validation_thm1} and Appendix~\ref{app:exp-thm1}), tightening the worst-case ceiling by orders of magnitude.

\textbf{Equal Capacity Across Tokens.}
\label{sec:thm2}
Two positions that attend to the same context tend to induce similar gradient structures, and hence similar fine-tuned predictors, whereas positions with different information access need not. Bidirectional models provide the natural setting for this comparison, since every position attends to the full sequence. The result below makes this intuition precise: for any two positions $i$ and $j$ whose tokens are fixed across samples and whose Gram matrices share a common column space, the gap between the corresponding fine-tuned predictors is bounded by quantities computable from the pretrained model.

We work in the \emph{representation NTK regime}: at a fixed layer $\ell$ following an attention layer, the feature map is the pretrained hidden state $\phi_t(x) = h_t^{(\ell)}(x) \in \mathbb{R}^d$, and the Gram matrix
\begin{equation*}
K_t[n,m] = \langle h_t^{(\ell)}(x^{(n)}),\, h_t^{(\ell)}(x^{(m)}) \rangle
\end{equation*}
is computable from a single forward pass. This regime models training as a linear readout from pretrained features; Appendix~\ref{ab:laziness} verifies that real fine-tuning stays close to it (mean weight drift below $0.011$). Define the kernel-evaluation vector $k_t(x) \in \mathbb{R}^N$ and the residual $r_t \in \mathbb{R}^N$ with entries
\begin{equation*}
k_t(x)[n] := \langle h_t^{(\ell)}(x),\, h_t^{(\ell)}(x^{(n)}) \rangle,
\qquad
r_t := y - o_t(X; \theta_0).
\end{equation*}
The capacity gap between two positions $i$ and $j$ is then controlled by four measurable quantities, all computed without any fine-tuning:
\begin{align*}
&\text{Gram matrix difference:}     && \delta_K := \|K_i - K_j\|_{\mathrm{op}}, \\
&\text{kernel vector difference:}   && \delta_k(x) := \|k_i(x) - k_j(x)\|_2, \\
&\text{residual difference:}        && \delta_r := \|r_i - r_j\|_2, \\
&\text{initial output gap:}         && \delta_{\mathrm{init}}(x) := \|o_i(x; \theta_0) - o_j(x; \theta_0)\|_2.
\end{align*}

\begin{theorembox}[Theorem 2: Equal capacity under shared information access] \customlabel{thm:equal}{2}
\setlength{\abovedisplayskip}{1pt}\setlength{\belowdisplayskip}{1pt}%
\setlength{\abovedisplayshortskip}{1pt}\setlength{\belowdisplayshortskip}{1pt}%
Let $f_i$ and $f_j$ be the representation NTK predictors (scalar output, $C=1$) obtained by training via gradient masking at positions $i$ and $j$ on $\mathcal{D}$, with $x_i^{(n)} = s_i$ and $x_j^{(n)} = s_j$ for all $n = 1,\dots,N$. Suppose $K_i$ and $K_j$ share the same column space, and let $\lambda_0^{+} := \min(\lambda_{\min}^{+}(K_i),\, \lambda_{\min}^{+}(K_j)) > 0$ denote the smallest nonzero eigenvalue of either. Then for all test inputs $x$ with $x_i = s_i$ and $x_j = s_j$,
\begin{equation*}
    \|f_i(x) - f_j(x)\|_2
    \;\leq\;
    \delta_{\mathrm{init}}(x)
    \;+\; \frac{\delta_k(x)\, R}{\lambda_0^{+}}
    \;+\; \frac{\|k_j(x)\|_2\, \delta_K\, R}{(\lambda_0^{+})^2}
    \;+\; \frac{\|k_j(x)\|_2\, \delta_r}{\lambda_0^{+}},
\end{equation*}
where $R := \|r_i\|_2$. Every quantity on the right-hand side is computable from the pretrained model.
\end{theorembox}

The proof is given in Appendix~\ref{app:proof-equal}, and empirical validation is provided in Section~\ref{exp:validation_thm2} and Appendices~\ref{app:exp-thm2} and~\ref{app:ablation-token-importance}.

\textbf{Monotone Capacity in Causal Models.}
\label{sec:thm3}
Theorems~\ref{thm:capacity} and~\ref{thm:equal} established that every position is a bounded bottleneck and that positions with aligned information access are equivalent. What remains is the causal case, where each position attends only to its prefix: as the position index increases, the prefix grows, and position $t{+}1$ has access to information that position $t$ does not. Causal attention provides the mechanism by which information at position $t$ enters position $t{+}1$'s representation: at every layer, the update at $t{+}1$ reads from positions $1, \ldots, t{+}1$. This shared lineage links the hidden states at adjacent positions through a regularity that we formalize below. We show that this expanding information access translates into a monotone fine-tuning capacity, up to a slack term that vanishes when the residual $\sigma_t \to 0$, in the representation NTK regime where $K_t[n,m] = \langle h_t^{(\ell)}(x^{(n)}), h_t^{(\ell)}(x^{(m)}) \rangle$. The positive semidefinite ordering $K_i \preceq K_j$ is the natural notion of capacity ordering: when it holds exactly, $\mathrm{Im}(K_i) \subseteq \mathrm{Im}(K_j)$, so every labeling of the training set achievable via position $i$ is also achievable via position $j$.

\begin{theorembox}[Theorem 3: Monotone capacity in causal models] \customlabel{thm:monotone}{3}
Suppose that for each $t \in \{1, \ldots, T{-}1\}$ there exists a matrix $A_t \in \mathbb{R}^{d \times d}$ with $\|A_t\|_{\mathrm{op}} \leq 1$, \textbf{shared uniformly across all inputs}, and residuals $\varepsilon_t(x^{(n)}) \in \mathbb{R}^d$ such that the hidden states at a fixed layer $\ell$ of a pretrained causal transformer $f_\theta$ satisfy
{\setlength{\abovedisplayskip}{1pt}\setlength{\belowdisplayskip}{1pt}
\[
    h_t^{(\ell)}(x^{(n)}) \;=\; A_t\, h_{t+1}^{(\ell)}(x^{(n)}) + \varepsilon_t(x^{(n)})
    \qquad \text{for all } n = 1, \ldots, N.
\]
Define $\sigma_t := \max_{n} \|\varepsilon_t(x^{(n)})\|_2$ and $\beta_{t+1} := \max_{n} \|h_{t+1}^{(\ell)}(x^{(n)})\|_2$. Then for every $t \in \{1, \ldots, T{-}1\}$,
\[
    K_{t+1} \;\succeq\; K_t - \delta_t\, I_N, \qquad \delta_t := N \sigma_t (2 \beta_{t+1} + \sigma_t),
\]}
where $\sigma_t$ and $\beta_{t+1}$ are computable from $N$ forward passes. The bound is informative whenever $\delta_t \ll \lambda_{\min}(K_t)$.
\end{theorembox}

The substantive restriction is the uniformity of $A_t$ across inputs, and the bound is non-trivial when $\sigma_t \ll \beta_{t+1}$. Empirically, $\|A_t\|_{\mathrm{op}} \approx 1$ across the bulk of layers, $\sigma_t / \beta_{t+1} \lesssim 10^{-14}$, and $\delta_t \ll \lambda^+_{\min}(K_t)$ by at least six orders of magnitude on SmolLM2-135M and at least seven on LLaMA~3-8B (Appendices~\ref{app:exp-thm3_at} and~\ref{app:thm3-contraction-validation}). When $\sigma_t = 0$ the bound holds with equality, recovering the exact chain $K_1 \preceq K_2 \preceq \cdots \preceq K_T$. The proof is given in Appendix~\ref{app:proof-monotone}, and empirical validation in Section~\ref{exp:validation_thm3}.

\section{Memory-Efficient Fine-Tuning via the Token Hypothesis}
\label{sec:method}

The three theorems of Section~\ref{sec:theory} jointly imply that fine-tuning through every position is unnecessary: capacity is bounded per position (Theorem~\ref{thm:capacity}), equivalent across aligned positions (Theorem~\ref{thm:equal}), and monotone in causal models (Theorem~\ref{thm:monotone}). A minimal number of positions $M^{\ast}$ therefore suffices to span the same achievable prediction space as the full sequence, and its value depends on the architecture. Given an ordering $t_1, t_2, \ldots, t_T$ of the positions, let $\mathcal{S}_M := \{t_1, \ldots, t_M\}$ and define the combined Gram matrix $K^{(M)} := \sum_{m=1}^{M} K_{t_m} \in \mathbb{R}^{N \times N}$. Since $\mathrm{Im}(K^{(M)})$ is monotone non-decreasing in $M$ (adding a PSD matrix cannot shrink the image), the minimum number of positions required to reach the full image is well-defined.

\begin{theorembox}[Corollary 1: Optimal position count] \customlabel{cor:optimal-M}{1}
\setlength{\abovedisplayskip}{1pt}\setlength{\belowdisplayskip}{1pt}%
\setlength{\abovedisplayshortskip}{1pt}\setlength{\belowdisplayshortskip}{1pt}%
Under the conditions of Theorems~\ref{thm:capacity}--\ref{thm:monotone} and in the output-head NTK regime, the quantity
\begin{equation*}
M^{\ast} := \min\bigl\{\, M : \mathrm{Im}(K^{(M)}) = \mathrm{Im}(K^{(T)}) \,\bigr\}
\end{equation*}
satisfies $1 \leq M^{\ast} \leq T$, and its value depends on both architecture and ordering:

\vspace{2pt}
\noindent\hangindent=1.2em\hangafter=1
\textbf{(i)~Bidirectional.} If $\|K_i - K_j\|_{\mathrm{op}} \leq \epsilon_K$ for all $i, j$ and $\lambda^{+} := \lambda^{+}_{\min}(K^{(T)})$ satisfies $\epsilon_K < \lambda^{+}/(2T)$, then $M^{\ast} \leq \lceil \epsilon_K T / \lambda^{+} + 1 \rceil$, with $M^{\ast} = 1$ in the exact case $\epsilon_K = 0$.

\vspace{2pt}
\noindent\hangindent=1.2em\hangafter=1
\textbf{(ii)~Causal, last-to-first ordering.} Under Theorem~\ref{thm:monotone} with strict inclusion $\mathrm{Im}(K_t) \subsetneq \mathrm{Im}(K_{t+1})$, the ordering $t_1 = T,\, t_2 = T{-}1, \ldots$ gives $M^{\ast} = 1$: the single last position spans the same prediction space as all $T$ positions combined.

\vspace{2pt}
\noindent\hangindent=1.2em\hangafter=1
\textbf{(iii)~Causal, first-to-last ordering.} Under the same conditions, the forward ordering $t_m = m$ gives $M^{\ast} = T$: every position adds a strictly new direction.
\end{theorembox}

Here $M^\ast$ is the smallest number of positions at which all prediction directions reachable with the full sequence become reachable at convergence, whereas the number of positions required under finite-step training additionally depends on the conditioning of $K^{(M)}$, which in our measurements continues to improve well beyond $M^\ast$. The proof is given in Appendix~\ref{app:proof-optimal-M}. All three quantities $\epsilon_K$, $\lambda^{+}$, and the monotone chain of images are measurable from forward passes alone. In practice $M^{\ast}$ is identified without fine-tuning through an \emph{image-saturation criterion}: find the smallest $M$ such that $\|P_{K^{(T)}} - P_{K^{(M)}} P_{K^{(T)}}\|_F / \|P_{K^{(T)}}\|_F \leq \varepsilon$, where $P_K$ is the orthogonal projector onto $\mathrm{Im}(K)$. Trace-based criteria are unreliable here: $\mathrm{tr}(K^{(M)})$ grows with $M$ even when all $K_t$ are identical, so it cannot detect image saturation.

\begin{figure}[ht!]
\centering
\includegraphics[width=\linewidth]{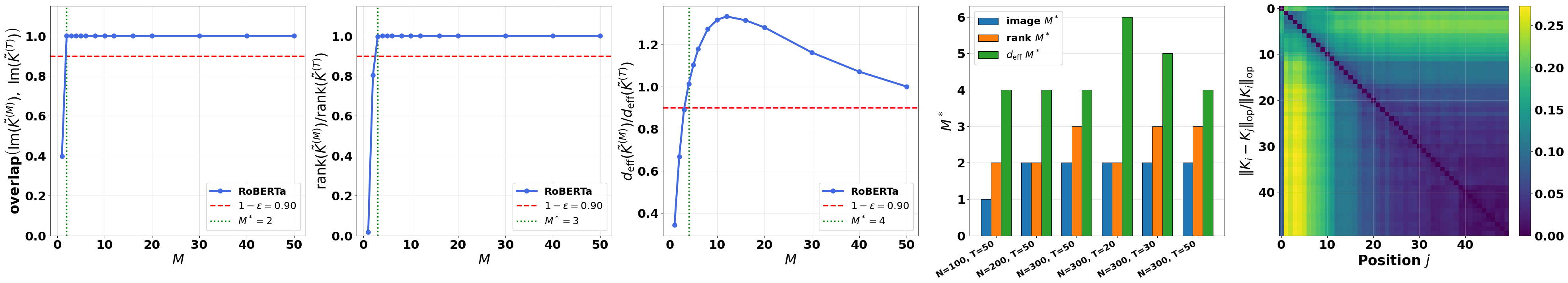}
\caption{Image-space saturation on RoBERTa-base (bidirectional, $N{=}300$, $T{=}50$). \textbf{(a)}~Image-space overlap crosses the $1-\varepsilon = 0.90$ threshold at $M^{\ast} = 2$. \textbf{(b)}~Rank saturation at $M^{\ast} = 3$. \textbf{(c)}~Effective rank $d_{\mathrm{eff}}(K^{(M)})$ saturates at $M^{\ast} = 4$. \textbf{(d)}~$M^{\ast}$ remains stable across $(N, T)$, indicating an architecture-intrinsic threshold. \textbf{(e)}~Pairwise relative Gram distance $\|K_i - K_j\|_{\mathrm{op}} / \|K_i\|_{\mathrm{op}}$ has mean $0.15$ and max $0.27$, supporting $\epsilon_K \ll 1$.}
\label{fig:mstar-roberta}

\end{figure}

\begin{figure}[ht!]
\centering
\includegraphics[width=\linewidth]{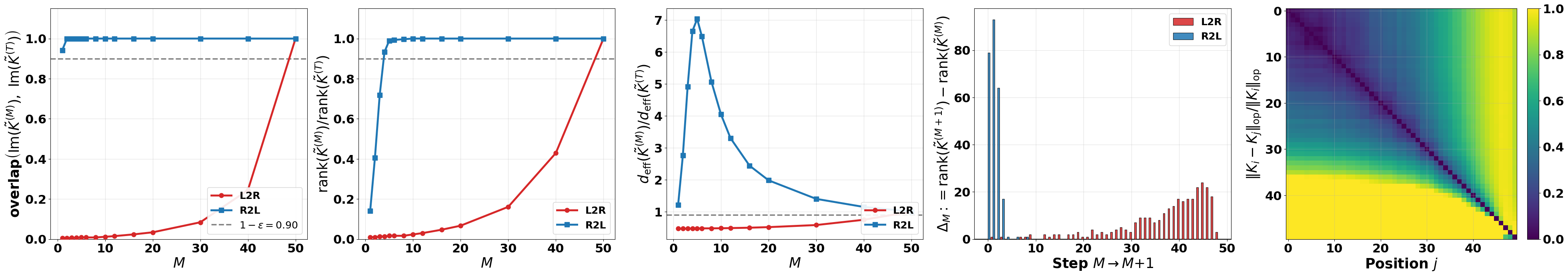}
\caption{Image-space saturation on SmolLM2-135M (causal, $N{=}300$, $T{=}50$). \textbf{(a)}~Forward ordering (L2R) reaches $\mathrm{Im}(K^{(T)})$ only at $M = T$; reverse ordering (R2L) reaches it at $M = 1$. \textbf{(b)}~Rank saturation: L2R at $M^{\ast} = T$, R2L at $M^{\ast} = 4$. \textbf{(c)}~Effective rank $d_{\mathrm{eff}}(K^{(M)})$ for both orderings. \textbf{(d)} Per-step rank increment $\Delta_M = \mathrm{rank}(\tilde K^{(M+1)}) - \mathrm{rank}(\tilde K^{(M)})$: under R2L, nearly all new directions are contributed by the first three positions, consistent with rank saturation at $M^{\ast}{=}4$; under L2R, increments are small and distributed across the sequence, growing toward its end. \textbf{(e)}~Pairwise relative Gram distance shows a strong gradient, supporting strict inclusion $\mathrm{Im}(K_t) \subsetneq \mathrm{Im}(K_{t+1})$.}
\label{fig:mstar-smollm}

\end{figure}

Figures~\ref{fig:mstar-roberta} and~\ref{fig:mstar-smollm} empirically confirm Corollary~\ref{cor:optimal-M}. On RoBERTa-base (bidirectional), all three saturation metrics cross the $1-\varepsilon = 0.90$ threshold at $M^{\ast} \leq 4$, and the value of $M^{\ast}$ remains stable as $(N, T)$ vary---so a handful of positions suffices regardless of dataset size. On SmolLM2-135M (causal), the reverse ordering saturates immediately while the forward ordering requires the full sequence, directly matching the two causal cases of the corollary; kernel-ridge accuracy under R2L peaks at $M = 1$, while L2R continues improving up to $M = T$.

Corollary~\ref{cor:optimal-M} translates into a memory-efficient fine-tuning recipe. Given $T$ positions, we partition $\{1, \ldots, T\}$ into a trainable set $\mathcal{S}$ of size $M$ and its complement $\bar{\mathcal{S}}$. The choice of $\mathcal{S}$ follows the architecture: in \emph{bidirectional} models, any subset of size $M \geq M^{\ast}$ achieves image equality with $K^{(T)}$, so we use a uniformly random subset resampled each epoch; in \emph{causal} models, the last-$M$ strategy $\mathcal{S} = \{T{-}M{+}1, \ldots, T\}$ is optimal under both the corollary and Theorem~\ref{thm:monotone} (i.e., $\mathcal{S} = \mathcal{S}_M$ under the reverse ordering). The forward pass is unchanged: at each layer we compute $\hat{x}_t = W x_t$ for every position, but retain gradients only for $t \in \mathcal{S}$, while positions $t \in \bar{\mathcal{S}}$ are computed inside a \texttt{no\_grad} context. Their activations are therefore never materialized in the computation graph, reducing peak activation memory by a fraction $1 - M/T$. When composed with LoRA~\cite{hu2022lora}, trainable positions additionally receive the adapter update while frozen positions do not: $\hat{x}_t = W x_t + (\alpha/r)\, B A\, x_t$ for $t \in \mathcal{S}$ and $\hat{x}_t = W x_t$ for $t \in \bar{\mathcal{S}}$, where $A \in \mathbb{R}^{r \times d}$, $B \in \mathbb{R}^{d \times r}$ are the rank-$r$ adapter factors and $\alpha$ is the scaling factor. The same construction applies to DoRA~\cite{liu2024dora}. After training, the gradient-splitting mechanism is discarded and any low-rank updates are merged into $W$, leaving the original architecture with no inference overhead. In practice $M$ can be used as a hyperparameter.

\section{Experiments}
\label{sec:experiments}

 We evaluate the Token Hypothesis across six task modalities---commonsense reasoning, mathematical reasoning, visual question answering, video question answering, the GLUE benchmark, and image classification. Dataset descriptions and preprocessing are provided in Appendix~\ref{app:datasets}. We use five pretrained backbones: LLaMA-3-8B, Qwen3-VL-8B, LLaVA-OneVision-7B, RoBERTa-base, and ViT-L/16-224, with details in Appendix~\ref{app:models}. Each modality is evaluated under full fine-tuning and a parameter-efficient setting (LoRA or DoRA), and every experiment sweeps the number of trainable positions $M$ against a full-sequence baseline (\textit{All}). Within each (modality, regime) pair, the baseline and all $M$-swept runs share identical learning rate, batch size, gradient accumulation, warmup schedule, mixed-precision setting, epoch count, and GPU configuration; the only variable is $M$. Hyperparameter details are provided in Appendix~\ref{app:hyperparameters}.

\begin{table*}[ht!]
\centering\tiny
\setlength{\tabcolsep}{2pt}
\renewcommand{\arraystretch}{1.1}
\caption{
  Unified comparison of \textit{average} results across four modalities
  (Commonsense, Math, VQA, Video) and three fine-tuning regimes
  (Full FT, LoRA, DoRA) vs.\ number of trainable tokens.
  Each block reports mean accuracy, total GPU memory summed across all devices~(GB),
  and training time~(min). VQA and Video results use Qwen3-VL-8B. Commonsense FT averages five benchmarks and Commonsense DoRA averages four.
  \textbf{Bold} = best value per column; ties are bolded jointly.
  Detailed results are shown in Tables~\ref{tab:main_results},~\ref{tab:dora_results},~\ref{tab:math_results},~\ref{tab:qwen_full_results},~\ref{tab:qwen_lora_results},~\ref{tab:video_ft_results},~and~\ref{tab:video_lora_results}.
}
\label{tab:unified_avg}
\resizebox{\textwidth}{!}{%
\begin{tabular}{@{}lrrrrrrrrrrrrrrrrrrrrrrrr@{}}
\toprule
\multirow{2}{*}{\textbf{Tokens}} & \multicolumn{3}{c}{\textbf{Commonsense FT}} & \multicolumn{3}{c}{\textbf{Commonsense DoRA}} & \multicolumn{3}{c}{\textbf{Math FT}} & \multicolumn{3}{c}{\textbf{Math DoRA}} & \multicolumn{3}{c}{\textbf{VQA FT}} & \multicolumn{3}{c}{\textbf{VQA LoRA}} & \multicolumn{3}{c}{\textbf{Video FT}} & \multicolumn{3}{c}{\textbf{Video LoRA}} \\
\cmidrule(lr){2-4}\cmidrule(lr){5-7}\cmidrule(lr){8-10}\cmidrule(lr){11-13}\cmidrule(lr){14-16}\cmidrule(lr){17-19}\cmidrule(lr){20-22}\cmidrule(lr){23-25}
 & \textit{Acc} & \textit{GPU} & \textit{Time} & \textit{Acc} & \textit{GPU} & \textit{Time} & \textit{Acc} & \textit{GPU} & \textit{Time} & \textit{Acc} & \textit{GPU} & \textit{Time} & \textit{Acc} & \textit{GPU} & \textit{Time} & \textit{Acc} & \textit{GPU} & \textit{Time} & \textit{Acc} & \textit{GPU} & \textit{Time} & \textit{Acc} & \textit{GPU} & \textit{Time} \\
\midrule
\rowcolor{basecol}
\textit{All} & 0.831 & 95.28 & 23.85 & 0.744 & 50.20 & 9.31 & 0.829 & 95.38 & 20.62 & 0.774 & 55.59 & 21.31 & 0.728 & 168.07 & 13.29 & 0.669 & 88.12 & 10.95 & 0.604 & 113.09 & 24.53 & \textbf{0.579} & 56.87 & 23.15 \\
10 & 0.744 & \textbf{83.69} & \textbf{19.35} & 0.687 & \textbf{49.24} & \textbf{9.23} & 0.399 & \textbf{83.93} & \textbf{16.73} & 0.494 & \textbf{49.70} & 20.93 & 0.711 & \textbf{135.29} & 11.69 & 0.663 & \textbf{76.22} & 9.48 & 0.574 & \textbf{101.03} & \textbf{21.29} & 0.531 & \textbf{49.78} & 20.07 \\
\rowcolor{rowA}
25 & 0.801 & 83.91 & 19.45 & 0.697 & 49.30 & 9.24 & 0.576 & 84.15 & 17.02 & 0.602 & 49.79 & 20.90 & 0.711 & 135.58 & \textbf{11.30} & 0.652 & 76.31 & 9.57 & 0.562 & 101.44 & 21.31 & 0.536 & 49.93 & \textbf{20.00} \\
50 & 0.831 & 84.63 & 19.81 & 0.736 & 49.43 & 9.32 & 0.695 & 84.61 & 16.98 & 0.687 & 49.94 & \textbf{20.84} & 0.720 & 136.06 & 11.44 & 0.665 & 76.46 & \textbf{9.18} & 0.574 & 102.08 & 21.87 & 0.528 & 50.17 & 20.28 \\
\rowcolor{rowA}
100 & 0.834 & 85.85 & 20.49 & 0.744 & 49.62 & 9.36 & 0.790 & 85.34 & 17.57 & 0.773 & 50.26 & 21.15 & 0.736 & 136.82 & 11.60 & 0.659 & 76.74 & 9.96 & 0.574 & 104.02 & 22.28 & 0.497 & 50.60 & 20.38 \\
300 & \textbf{0.838} & 88.87 & 22.04 & \textbf{0.747} & 50.38 & 9.46 & \textbf{0.834} & 88.35 & 19.13 & \textbf{0.790} & 51.41 & 21.45 & \textbf{0.739} & 140.13 & 12.48 & \textbf{0.676} & 77.86 & 9.94 & 0.588 & 108.62 & 23.74 & 0.540 & 52.37 & 20.66 \\
\rowcolor{rowA}
500 & 0.830 & 91.54 & 23.74 & 0.744 & 51.20 & 9.58 & 0.833 & 91.49 & 20.72 & 0.782 & 52.60 & 21.84 & 0.735 & 143.45 & 13.20 & 0.675 & 78.98 & 9.81 & \textbf{0.609} & 109.12 & 22.51 & 0.535 & 53.30 & 20.44 \\
700 & 0.831 & 94.21 & 25.43 & 0.745 & 51.95 & 9.70 & 0.828 & 94.47 & 22.88 & 0.786 & 53.72 & 22.05 & 0.731 & 146.72 & 13.55 & 0.668 & 80.07 & 9.68 & 0.593 & 109.35 & 22.89 & 0.550 & 53.52 & 20.79 \\
\rowcolor{rowA}
1000 & 0.832 & 96.62 & 27.16 & 0.746 & 52.72 & 9.79 & \textbf{0.834} & 97.50 & 25.15 & 0.782 & 54.87 & 22.52 & 0.724 & 151.70 & 13.68 & 0.663 & 81.77 & 9.70 & 0.597 & 109.70 & 22.48 & 0.563 & 53.84 & 20.61 \\
\bottomrule
\end{tabular}%
}
\end{table*}
\begin{table*}[ht]
\centering
\tiny
\setlength{\tabcolsep}{3pt}
\renewcommand{\arraystretch}{1.1}
\caption{Accuracy comparison across six benchmarks on LLaMA-3.2-3B for varying numbers of trainable tokens.
TT = Token Tuning, TS = TokenSeek, TH = Token Hypothesis. GPU = peak allocated memory (GB),
shared across the three methods at each token count.
Highest accuracy in each cell is shown in \textbf{bold}.}
\label{tab:method_comparison_rotated}
\begin{tabular}{r ccc ccc ccc ccc ccc ccc r}
\toprule
& \multicolumn{3}{c}{\textbf{ARC-Chal.}} & \multicolumn{3}{c}{\textbf{ARC-Easy}} & \multicolumn{3}{c}{\textbf{BoolQ}} & \multicolumn{3}{c}{\textbf{HellaSwag}} & \multicolumn{3}{c}{\textbf{OpenBookQA}} & \multicolumn{3}{c}{\textbf{WinoGrande}} & \\
\cmidrule(lr){2-4} \cmidrule(lr){5-7} \cmidrule(lr){8-10} \cmidrule(lr){11-13} \cmidrule(lr){14-16} \cmidrule(lr){17-19}
\textbf{Tokens} & TT & TS & TH & TT & TS & TH & TT & TS & TH & TT & TS & TH & TT & TS & TH & TT & TS & TH & \textbf{GPU} \\
\midrule
 10 & 0.258 & \textbf{0.519} & 0.504 & 0.256 & 0.764 & \textbf{0.768} & 0.622 & 0.651 & \textbf{0.678} & 0.438 & 0.683 & \textbf{0.776} & 0.364 & \textbf{0.630} & 0.616 & 0.496 & 0.705 & \textbf{0.728} & 29.99 \\
 25 & 0.470 & \textbf{0.607} & 0.528 & 0.693 & \textbf{0.816} & 0.812 & 0.622 & 0.614 & \textbf{0.719} & 0.491 & 0.247 & \textbf{0.892} & 0.468 & \textbf{0.730} & 0.664 & 0.496 & 0.760 & \textbf{0.775} & 29.99 \\
 50 & 0.486 & \textbf{0.643} & 0.635 & 0.653 & 0.808 & \textbf{0.841} & 0.622 & 0.714 & \textbf{0.717} & 0.593 & 0.255 & \textbf{0.903} & 0.448 & 0.714 & \textbf{0.772} & 0.496 & 0.810 & \textbf{0.829} & 29.99 \\
100 & 0.420 & \textbf{0.672} & 0.665 & 0.734 & 0.855 & \textbf{0.859} & 0.622 & \textbf{0.722} & 0.721 & 0.762 & 0.455 & \textbf{0.924} & 0.316 & \textbf{0.784} & 0.782 & 0.656 & 0.830 & \textbf{0.834} & 30.00 \\
300 & 0.570 & 0.661 & \textbf{0.666} & 0.787 & \textbf{0.858} & 0.856 & 0.651 & 0.713 & \textbf{0.717} & 0.864 & 0.931 & \textbf{0.932} & 0.676 & 0.786 & \textbf{0.788} & 0.681 & 0.837 & \textbf{0.838} & 35.82 \\
500 & 0.622 & 0.658 & \textbf{0.664} & 0.807 & 0.854 & \textbf{0.857} & 0.684 & 0.709 & \textbf{0.719} & 0.887 & \textbf{0.933} & 0.932 & 0.722 & \textbf{0.782} & \textbf{0.782} & 0.800 & 0.838 & \textbf{0.838} & 42.79 \\
700 & 0.659 & \textbf{0.666} & 0.664 & 0.824 & \textbf{0.859} & 0.857 & 0.695 & 0.714 & \textbf{0.719} & 0.907 & 0.932 & \textbf{0.935} & 0.730 & \textbf{0.786} & 0.780 & 0.781 & 0.840 & \textbf{0.841} & 49.73 \\
900 & 0.658 & \textbf{0.666} & 0.658 & 0.849 & 0.853 & \textbf{0.858} & 0.714 & \textbf{0.717} & 0.716 & 0.930 & \textbf{0.934} & 0.933 & 0.766 & \textbf{0.782} & \textbf{0.782} & 0.841 & 0.846 & \textbf{0.848} & 56.68 \\
\bottomrule
\end{tabular}

\end{table*}

\textbf{Main results.} Table~\ref{tab:unified_avg} reports accuracy averaged across benchmarks within each modality and summarizes how accuracy, memory, and time vary with $M$ across eight (modality, regime) pairs on causal backbones. Three observations stand out. First, accuracy matches or exceeds the full-sequence baseline at $M$ far smaller than the sequence length in seven of the eight (modality, regime) pairs, with the saturation point varying by task: commonsense and visual QA saturate at $M{=}100$--$300$, math reasoning at $M{=}300$, and video reasoning under full fine-tuning first exceeds the baseline at $M{=}500$. Video LoRA is the exception: average accuracy remains below the baseline at every $M$ (best $0.563$ at $M{=}1000$ versus $0.579$), with the gap driven primarily by VideoMathQA. Empirically, the required budget grows with the length of the supervised span, being larger for multi-step reasoning and video inputs than for short-answer tasks. Second, the empirical saturation point lies above the minimal position count $M^{*}$ of Corollary~\ref{cor:optimal-M}, which equals one for causal models under last-$M$ selection; the practical optimum reflects a preference for stable optimization and well-conditioned kernels over the minimal spanning subset. Third, in the settings that saturate, accuracy remains essentially flat beyond the saturation point, indicating that the additional gradient signal from earlier positions, which are added as $M$ grows under last-$M$ selection, is largely redundant rather than harmful---in line with the monotone capacity ordering $K_t \preceq K_{t+1}$. We also observe parallel results on bidirectional backbones: GLUE on RoBERTa-base (Tables~\ref{tab:glue1} and~\ref{tab:glue2}) and image classification on ViT-L/16-224 (Table~\ref{tab:img_cls_full}) saturate at $M{=}50$--$100$, consistent with Theorem~\ref{thm:equal}'s prediction that small subsets of aligned positions suffice in the bidirectional case. Detailed per-benchmark results are in Appendix~\ref{app:detailed-results}.

The memory reduction is driven directly by restricting the backward graph to the $M$ trainable positions: backward-graph activations are stored only for these positions, while the remaining $T-M$ positions execute under \texttt{torch.no\_grad()} and never enter the gradient buffer. The absolute saving therefore grows with the number of frozen positions in each sequence. At $M{=}300$, VQA achieves the largest reduction in Table~1 ($27.9$~GB under full fine-tuning and $10.3$~GB under LoRA), compared with $6.4$~GB for commonsense, $7.0$~GB for math, and $4.5$~GB for video. The reduction is orthogonal to parameter-side compression: applying token masking on top of LoRA or DoRA further reduces total memory while preserving the parameter savings of the adapter, since the two mechanisms target disjoint components of the GPU footprint. Training time tracks memory closely, dominated by the per-step backward pass over the trainable positions.

\textbf{Comparison.}
Table~\ref{tab:method_comparison_rotated} compares the Token Hypothesis (TH) against TokenTune (TT)~\cite{simoulin2024memory} and TokenSeek (TS)~\cite{zeng2026tokenseek} on six commonsense-reasoning benchmarks at matched token budgets, with shared GPU memory across all three methods at each token count. The Token Hypothesis achieves the highest accuracy in the majority of cells across all token counts, and matches TokenSeek closely where it does not lead. The gap to TokenTune is large at small token counts (e.g., $+0.34$ on HellaSwag and $+0.25$ on OpenBookQA at $M{=}10$), confirming that random selection is far from optimal when the budget is tight. Against TokenSeek, the Token Hypothesis achieves comparable or better accuracy without any pre-stage scoring or per-model hyperparameter search: TokenSeek requires a forward and partial backward pass over the entire training set to compute per-sample importance scores, plus a grid search over its weights for each backbone. The Token Hypothesis uses the last-$M$ rule directly, motivated by Theorem~\ref{thm:monotone}, which carries information from earlier high-importance tokens without training them. The training-time and memory cost is identical across the three methods at each $M$, so the comparison isolates the effect of the position-selection rule.

\paragraph{Empirical validation of Theorem~\ref{thm:capacity}.} \label{exp:validation_thm1}
\begin{figure}[ht!]
\centering
\includegraphics[width=\linewidth]{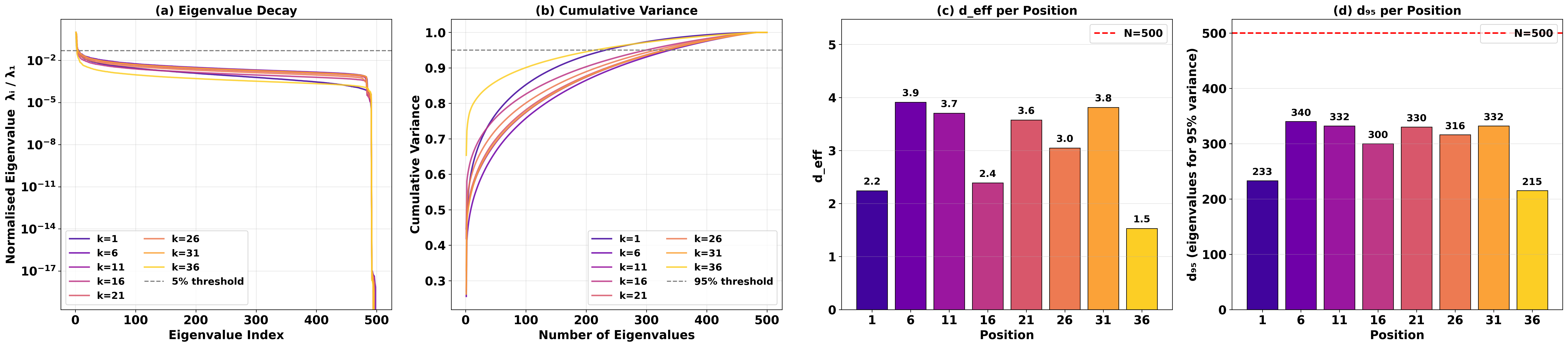}
\caption{Empirical validation of Theorem~\ref{thm:capacity} on RoBERTa-base ($N{=}500$, positions $t \in \{1, 6, 11, 16, 21, 26, 31, 36\}$; tokens at each position vary across samples). (a) Normalized eigenvalues of $K_t$ decay gradually over most of the spectrum and drop by more than ten orders of magnitude near its tail, so $K_t$ is numerically rank-deficient. (b) Cumulative variance crosses the 95\% threshold at indices 215--340, below $N{=}500$. (c) The effective rank $d_{\mathrm{eff}}(K_t)$ lies between 1.5 and 3.9 across all eight positions, indicating that the spectrum is dominated by a small number of leading directions. (d) $d_{95}$, the smallest number of top eigenvalues whose sum exceeds 95\% of $\mathrm{tr}(K_t)$, remains below $N$ at every position.}
\label{fig:thm1}

\end{figure}

Figure~\ref{fig:thm1} validates both claims of Theorem~\ref{thm:capacity} on RoBERTa-base. The worst-case rank bound $\mathrm{rank}(K_t) \leq \min(N, P)$ is visibly slack: $d_{95}$ stays well below $N$ at every position, so $K_t$ never approaches the ceiling. The stronger claim---that the \emph{effective} rank is far smaller still---is confirmed directly: $d_{\mathrm{eff}}(K_t) \in [1.5, 3.9]$ across all eight positions, so fine-tuning through a single position uses fewer than four effective directions. This narrow band persists as we vary the dataset size from $30$ to $2{,}000$ samples and as we move from RoBERTa-base ($125$M) to SmolLM2-$135$M and LLaMA-3-$8$B (Appendix~\ref{app:exp-thm1}). The bottleneck is therefore a property of the pretrained kernel itself: fixed at initialization, independent of dataset size or model scale.

\begin{figure}[ht!]
\centering
\includegraphics[width=\linewidth]{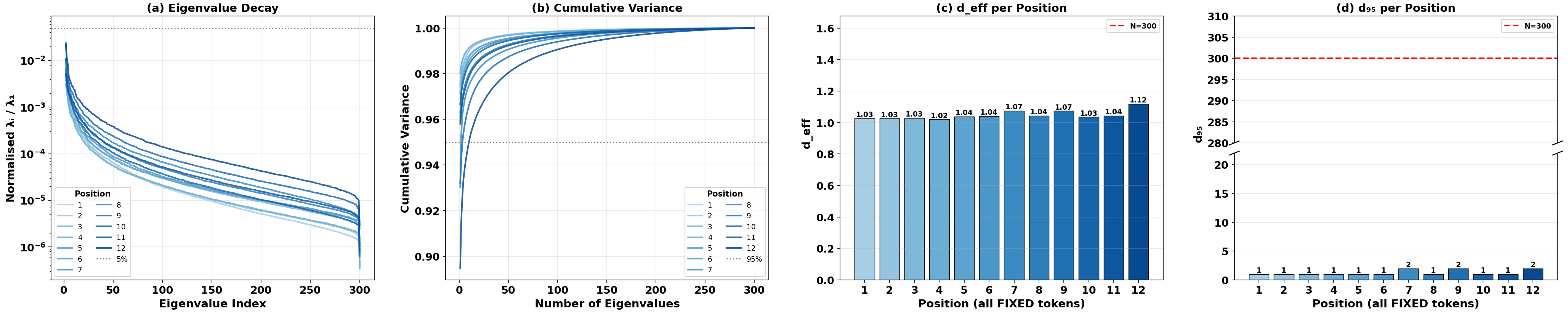}
\caption{Empirical validation of Theorem~\ref{thm:equal} on RoBERTa-base ($N{=}300$; 12 fixed prefix-prompt positions). \textbf{(a)}~Eigenvalue decay curves are nearly identical across all 12 positions, indicating $\delta_K \approx 0$. \textbf{(b)}~Cumulative variance curves are indistinguishable, consistent with the shared column-space assumption. \textbf{(c)}~$d_{\mathrm{eff}} \in [1.02, 1.12]$ with $\mathrm{CV} = 0.02$: every fixed position has essentially the same bottleneck width. \textbf{(d)}~$d_{95} = 1$ at almost every position, confirming $K_t$ is effectively rank-one and $\lambda_0^{+}$ is large, so the second and third terms of the bound are negligible.}
\label{fig:thm2}

\end{figure}

\paragraph{Empirical validation of Theorem~\ref{thm:equal}.}
\label{exp:validation_thm2}
Figure~\ref{fig:thm2} confirms Theorem~\ref{thm:equal} on RoBERTa-base with a fixed 12-token prefix prompt. All 12 positions yield $d_{\mathrm{eff}} \in [1.02, 1.12]$ with coefficient of variation $\mathrm{CV} = 0.02$, and $d_{95} \in \{1, 2\}$ at almost every position. When the token at a position is identical across all $N$ samples, the position-$t$ gradient has one dominant direction set by that fixed token's embedding, with context variation contributing only a small perturbation through bidirectional attention; this concentrates the kernel onto a single eigendirection. The resulting $\lambda_0^{+} \approx \lambda_1(K_t) \approx \mathrm{tr}(K_t)$ is large, making the second and third terms of the bound negligible. Full results for $N \in \{50, 300, 1000\}$ and a parallel experiment with a fixed suffix are deferred to Appendix~\ref{app:exp-thm2}.

We next test the prediction at the level of the fine-tuned predictor. Representation NTK predictors $f_i$ and $f_j$ are trained separately at ten token pairs spanning positions 1--8 of the same prompt, using 300 SST-2 training sentences and 50 held-out test sentences. Across all pairs, the mean prediction gap $\|f_i(x) - f_j(x)\|$ ranges from $18.0\%$ to $27.5\%$ of the label standard deviation $\hat{\sigma}_y$, and $86$--$98\%$ of test sentences satisfy $\|f_i(x) - f_j(x)\| < 0.5\,\hat{\sigma}_y$. All ten pairs are statistically significant ($p < 0.001$) under a left-tail position-exchangeability permutation test with $B{=}2000$ resamples, confirming that the observed closeness reflects genuine equal capacity across positions rather than chance (Table~\ref{tab:close-capacity} in Appendices~\ref{app:exp-thm2} and~\ref{app:ablation-token-importance}).

\paragraph{Empirical validation of Theorem~\ref{thm:monotone}.}
\label{exp:validation_thm3}

\begin{figure}[ht!]
\centering
\includegraphics[width=\linewidth]{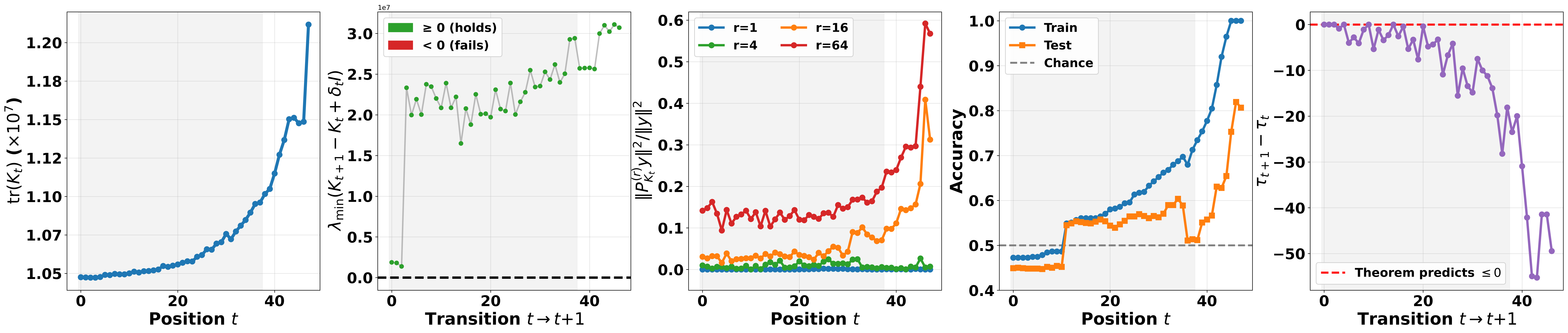}
\caption{Empirical validation of Theorem~\ref{thm:monotone} on LLaMA~3-8B ($T{=}48$, $N_{\mathrm{train}}{=}512$, $N_{\mathrm{test}}{=}1024$). \textbf{(a)}~$\mathrm{tr}(K_t)$ across positions. \textbf{(b)}~$\lambda_{\min}(K_{t+1} - K_t + \delta_t I)$ at every transition $t \to t{+}1$. \textbf{(c)}~Fraction of label variance captured by the top-$r$ eigendirections of $K_t$, for $r \in \{1, 4, 16, 64\}$. \textbf{(d)}~Training and test accuracy of a fine-tuned linear head reading from position $t$. \textbf{(e)}~Increment $\tau_{t+1} - \tau_t$ of the prediction-error statistic $\tau_t := \mathbb{E}_x [\|y - K_t K_t^+ y\|^2]$, with the dashed line indicating the theorem's prediction $\tau_{t+1} \leq \tau_t$.}
\label{fig:thm3}

\end{figure}

Figure~\ref{fig:thm3} validates Theorem~\ref{thm:monotone} along three complementary axes. The kernel-level claim $K_{t+1} \succeq K_t - \delta_t I$ is confirmed in two ways: $\mathrm{tr}(K_t)$ grows monotonically with position, and $\lambda_{\min}(K_{t+1} - K_t + \delta_t I) \geq 0$ at every transition, so the PSD bound holds along the entire sequence. The geometric consequence $\mathrm{Im}(K_t) \subseteq \mathrm{Im}(K_{t+1})$ shows up in eigendirection alignment with labels: the fraction of label variance captured by the top eigendirections of $K_t$ grows with position, with the top $64$ directions covering near zero at position $1$ and $60\%$ at position $45$. Finally, the downstream prediction---that later positions admit strictly larger function classes---is confirmed by linear-probe behavior: held-out accuracy rises from near chance at position $1$ to roughly $0.8$ at position $45$, and the prediction-error statistic $\tau_t$ decreases monotonically.

\textbf{Ablations.}
\label{sec:ablations-main}
We conduct five ablations to characterize the behavior of token-masked fine-tuning beyond the core accuracy-memory trade-off. \emph{Batch-size scaling} (Appendix~\ref{ab:batch}) shows that activation memory grows linearly in the batch size with a slope proportional to $M$, so reducing $M$ translates directly into a larger attainable batch size under a fixed memory budget---up to $3.6{\times}$ at batch size $64$. \emph{Sequence-length scaling} (Appendix~\ref{ab:seqlen}) shows that fixing $M{=}50$ while varying $T$ largely decouples memory from sequence length: token-masked memory grows by only 2--3~GB from $T{=}100$ to $T{=}2048$, while the number of positions in the backward graph is reduced by $1 - M/T \approx 97.6\%$ at $T{=}2048$. \emph{Convergence across $M$} (Appendix~\ref{ab:convergence}) shows that on LLaVA-OneVision-7B, $M{=}50$--$100$ reaches or exceeds full-sequence accuracy on SafeTag and MMK12 within 1000 evaluation steps, while the reasoning-heavy M3CoT benefits from a larger budget, consistent with the bounded per-position capacity of Theorem~\ref{thm:capacity}. \emph{NTK laziness across parameter groups} (Appendix~\ref{ab:laziness}) verifies that weight drift stays at or below $0.011$ across every parameter group and every value of $M$, empirically supporting the lazy regime Theorem~\ref{thm:capacity} assumes. \emph{Capability retention} (Appendix~\ref{ab:retention}) shows that fine-tuning with $M{=}100$ on ARC-C improves target accuracy while reducing forgetting on four untrained commonsense benchmarks by roughly $4{\times}$ relative to full fine-tuning, consistent with confining updates to a low-dimensional subspace of the pretrained tangent space.

\section{Conclusion}
\label{sec:conclusion}
We introduced the Token Hypothesis, a Neural Tangent Kernel analysis of how fine-tuning capacity is distributed across token positions. The analysis shows that each position acts as a low-dimensional capacity bottleneck, that aligned positions in bidirectional models have similar capacity, and that position-wise capacity increases along causal sequences. These predictions lead to a simple memory-efficient training procedure: backpropagate through only a small subset of positions while preserving the full forward pass. Empirically, a small number of trainable positions, ranging from fifty to five hundred depending on the task, matches or exceeds full-sequence fine-tuning in nearly all settings across six task families and five backbones, substantially reducing activation memory, enabling nearly four times larger batches, and composing cleanly with LoRA and DoRA. Together, the theory and experiments suggest that memory-efficient fine-tuning should be understood not only as a question of \emph{which} tokens to select, but also \emph{why} a small position subset can carry sufficient fine-tuning signal.

\begin{ack}
This work used Bridges-2 at the Pittsburgh Supercomputing Center through an allocation from the Advanced Cyberinfrastructure Coordination Ecosystem: Services \& Support (ACCESS) program, which is supported by U.S. National Science Foundation grants \#2138259, \#2138286, \#2138307, \#2137603, and \#2138296. The authors declare no competing interests.
\end{ack}

\bibliography{ref}
\bibliographystyle{plain}
\newpage
\appendix
\tableofcontents
\newpage
\section{Capacity Bound}
\subsection{Proof of Theorem~\ref{thm:capacity}}
\label{app:proof-capacity}

We restate the theorem for convenience before giving the proof.

\noindent
\textbf{Theorem~\ref{thm:capacity} (restated).} \emph{Fix a position $t \in \{1, \ldots, T\}$ and suppose the model is trained on $\mathcal{L}_t$ (the single-position loss defined in Section~\ref{eq:masked-loss}) by gradient descent in the NTK regime, with the position-$t$ Jacobians $\nabla_\theta o_t(\cdot; \theta)$ frozen at their initialization values throughout training. Let $f_t$ denote the converged predictor and let $r_t := y - o_t(X; \theta_0) \in \mathbb{R}^N$ be the residual at initialization. Then the learned correction $g_t(x) := f_t(x) - o_t(x; \theta_0)$ admits the representer form}
\[
    g_t(x) \;=\; \sum_{n=1}^{N} \alpha_n\, \mathcal{K}_t(x, x^{(n)}),
    \qquad \alpha \;=\; K_t^{+}\, r_t,
\]
\emph{where $K_t^{+}$ is the Moore--Penrose pseudoinverse of the Gram matrix $K_t$. In particular $g_t \in \mathcal{H}_{K_t}$, the correction on the training set satisfies $g_t(X) \in \mathrm{Im}(K_t)$, and the prediction on the training set lies in the affine subspace}
\[
    f_t(X) \;\in\; o_t(X; \theta_0) + \mathrm{Im}(K_t),
\]
\emph{whose dimension is $\mathrm{rank}(K_t) \leq \min(N, P)$, independent of $P$.}

\begin{proof}
The loss functional is
\[
    \mathcal{L}_t(\theta) \;=\; \frac{1}{2N}\sum_{n=1}^{N}
    \bigl(o_t(x^{(n)}; \theta) - y^{(n)}\bigr)^2,
\]
and one step of gradient descent with step size $\eta$ produces
\[
    \theta_{s+1} - \theta_s
    \;=\; -\frac{\eta}{N}\sum_{n=1}^{N}
          \bigl(o_t(x^{(n)}; \theta_s) - y^{(n)}\bigr)\,
          \nabla_\theta o_t(x^{(n)}; \theta_s),
\]
where $s$ indexes training steps. Under the frozen-Jacobian assumption, $\nabla_\theta o_t(x^{(n)}; \theta_s) = \phi_t(x^{(n)})$ for every $s \geq 0$, so each update is a linear combination of the fixed vectors $\{\phi_t(x^{(n)})\}_{n=1}^N$. Summing over all steps gives
\[
    \theta_S - \theta_0
    \;\in\; \mathcal{G}_t
    \;:=\; \mathrm{span}\{\phi_t(x^{(n)}) : n = 1,\ldots,N\}
    \;\subseteq\; \mathbb{R}^{P}
\]
for every $S \geq 0$, so parameter updates are confined to a subspace of dimension at most $N$, independent of the total parameter count $P$.

We pass to the continuous-time limit by rescaling $\tau := s\eta$ and taking $\eta \to 0$, which transforms gradient descent into gradient flow. Under the frozen-Jacobian assumption the predictor at any input $x$ evolves as
\[
    \frac{d}{d\tau} o_t(x; \theta_\tau)
    \;=\; \phi_t(x)^\top \frac{d\theta_\tau}{d\tau}
    \;=\; -\frac{1}{N}\sum_{n=1}^{N}
          \bigl(o_t(x^{(n)}; \theta_\tau) - y^{(n)}\bigr)\,
          \mathcal{K}_t(x, x^{(n)}),
\]
where the second equality uses $\mathcal{K}_t(x, x') = \langle \phi_t(x), \phi_t(x') \rangle$. Writing $\hat{g}(x, \tau) := o_t(x; \theta_\tau) - o_t(x; \theta_0)$ for the running correction, this becomes the kernel gradient flow
\[
    \frac{d}{d\tau} \hat{g}(x, \tau)
    \;=\; -\frac{1}{N}\sum_{n=1}^{N}
          \bigl(\hat{g}(x^{(n)}, \tau) - r_t^{(n)}\bigr)\,
          \mathcal{K}_t(x, x^{(n)}),
\]
and on the training set reduces to the linear ODE
\[
    \frac{d}{d\tau} \hat{g}(X, \tau)
    \;=\; -\frac{1}{N}\, K_t\,
          \bigl(\hat{g}(X, \tau) - r_t\bigr).
\]
Since $K_t \succeq 0$, the flow contracts on $\mathrm{Im}(K_t)$ (where $K_t$ has strictly positive eigenvalues) and is stationary on $\ker(K_t)$. Decomposing $r_t = r_t^{\parallel} + r_t^{\perp}$ with $r_t^{\parallel} \in \mathrm{Im}(K_t)$ and $r_t^{\perp} \in \ker(K_t)$, the flow drives $\hat{g}(X, \tau)$ from zero to $r_t^{\parallel}$ and leaves the orthogonal component untouched, so
\[
    \hat{g}(X, \infty) \;=\; r_t^{\parallel}
    \;=\; K_t K_t^{+}\, r_t,
\]
where $K_t K_t^{+}$ is the orthogonal projector onto $\mathrm{Im}(K_t)$.

Since $\theta_\infty - \theta_0 \in \mathcal{G}_t$, it can be written as $\theta_\infty - \theta_0 = \sum_{n=1}^{N} \alpha_n\, \phi_t(x^{(n)})$ for some $\alpha \in \mathbb{R}^N$. Evaluating the converged predictor at any input $x$ yields the representer form
\[
    g_t(x) \;:=\; \hat{g}(x, \infty)
    \;=\; \phi_t(x)^\top (\theta_\infty - \theta_0)
    \;=\; \sum_{n=1}^{N} \alpha_n\, \mathcal{K}_t(x, x^{(n)}).
\]
On the training set, $g_t(X) = K_t\alpha$, and combining with $\hat{g}(X, \infty) = K_t K_t^{+} r_t$ gives $K_t\alpha = K_t K_t^{+} r_t$. Among all $\alpha$ satisfying this equation, gradient flow from zero initialization selects the unique solution in $\mathrm{Im}(K_t)$, namely $\alpha = K_t^{+} r_t$; equivalently, $\alpha = K_t^{+} r_t$ minimizes the parameter-space displacement $\|\theta_\infty - \theta_0\|_2 = \|\Phi_t \alpha\|_2$ among interpolants on the training set~\cite{afzal2026linearization, malladi2023kernel}.

We now characterize the achievable prediction subspace. Evaluating the correction on the training set,
\[
    g_t(X) \;=\; K_t\alpha \;=\; K_t K_t^{+} r_t
    \;\in\; \mathrm{Im}(K_t),
\]
since the product of any matrix with any vector lies in its image. Conversely, for any $z \in \mathrm{Im}(K_t)$, choosing labels $y := o_t(X; \theta_0) + z$ gives $r_t = z$. By the convergence analysis above, $g_t(X) = K_t K_t^{+} z$, and since $K_t K_t^{+}$ is the orthogonal projector onto $\mathrm{Im}(K_t)$ with $z \in \mathrm{Im}(K_t)$ already, $K_t K_t^{+} z = z$, yielding $g_t(X) = z$. Hence the achievable set of corrections on the training set is exactly
\[
    \bigl\{ g_t(X; y) : y \in \mathbb{R}^N \bigr\}
    \;=\; \mathrm{Im}(K_t),
\]
and the full prediction vector lies in the affine subspace $f_t(X) \in o_t(X; \theta_0) + \mathrm{Im}(K_t)$.

Finally we bound the dimension of this subspace. Let $\Phi_t := [\phi_t(x^{(1)}) \mid \cdots \mid \phi_t(x^{(N)})] \in \mathbb{R}^{P \times N}$ be the feature matrix, so that $K_t = \Phi_t^\top \Phi_t$. By the identity $\mathrm{rank}(M^\top M) = \mathrm{rank}(M)$,
\[
    \dim \mathrm{Im}(K_t)
    \;=\; \mathrm{rank}(K_t)
    \;=\; \mathrm{rank}(\Phi_t)
    \;\leq\; \min(P, N),
\]
so the achievable prediction subspace has dimension at most $\min(P, N)$. Since in all our experiments $N \leq P$ by orders of magnitude, this reduces to $\mathrm{rank}(K_t) \leq N$, independent of the model's parameter count $P$.
\end{proof}

\subsection{Detailed Empirical Validation of Theorem~\ref{thm:capacity}}
\label{app:exp-thm1}

We validate Theorem~\ref{thm:capacity} across three pretrained transformers---RoBERTa-base ($125$M parameters, bidirectional), SmolLM2-135M ($135$M, causal), and LLaMA~3-8B ($8$B, causal)---and across dataset sizes $N \in \{30, 50, 100, 200, 300, 500, 1000, 2000\}$. For each (model, dataset size, position) configuration we compute the output-head NTK Gram matrix at the pretrained initialization $\theta_0$, before any fine-tuning: using the final-layer hidden states $h_t^{(D)}(x^{(n)}) \in \mathbb{R}^d$, we set
\[
    K_t[n,m] \;=\; \langle h_t^{(D)}(x^{(n)}),\, h_t^{(D)}(x^{(m)}) \rangle.
\]
This is a tractable proxy for the full NTK Gram matrix; in the output-head NTK regime, where only the final projection is trained, it equals $K_t$ exactly. For each configuration we report the effective rank $d_{\mathrm{eff}}(K_t) = \mathrm{tr}(K_t) / \|K_t\|_{\mathrm{op}}$, the $d_{95}$ metric (number of eigenvalues required to capture $95\%$ of the trace), the normalized eigenvalue spectrum, and the cumulative-variance curve. The theorem predicts $\mathrm{rank}(K_t) \leq \min(P, N)$; in all experiments $N \ll P$, so the meaningful ceiling is $N$. The effective-rank metric $d_{\mathrm{eff}}$ is tighter still, and quantifies the number of directions with non-negligible eigenvalues.

\begin{figure}[ht!]
\centering
\includegraphics[width=\linewidth]{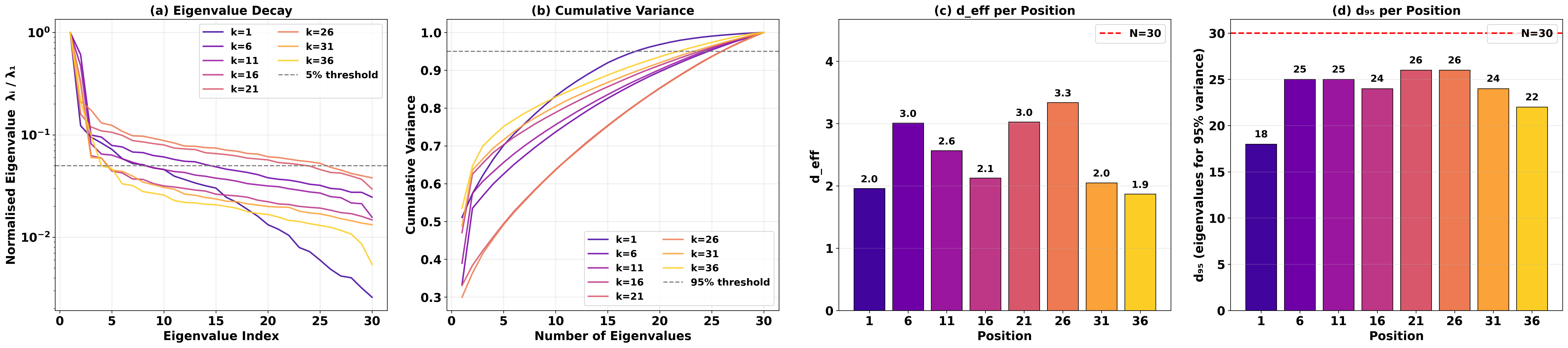}
\caption{RoBERTa-base, $N{=}30$. $d_{\mathrm{eff}} \in [1.9, 3.3]$, $d_{95} \in [18, 26]$, both well below the ceiling $N{=}30$.}
\label{fig:app-thm1-n30}
\end{figure}

\begin{figure}[ht!]
\centering
\includegraphics[width=\linewidth]{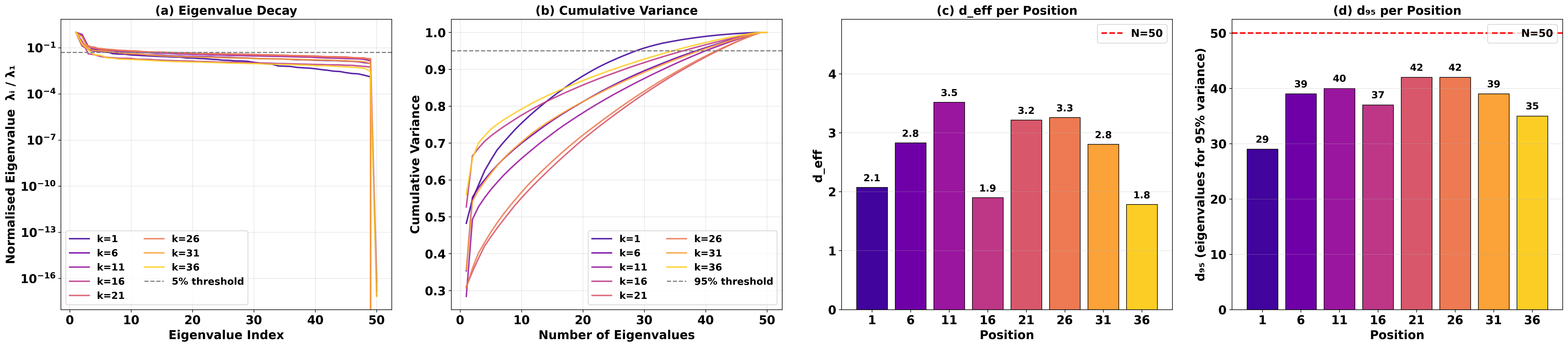}
\caption{RoBERTa-base, $N{=}50$. $d_{\mathrm{eff}} \in [1.8, 3.5]$, $d_{95} \in [29, 42]$.}
\label{fig:app-thm1-n50}
\end{figure}

\begin{figure}[ht!]
\centering
\includegraphics[width=\linewidth]{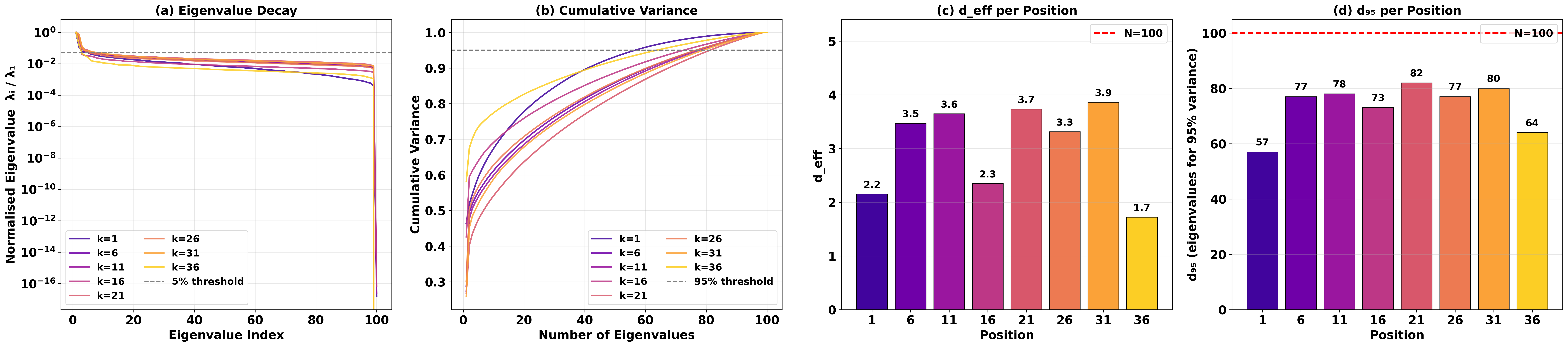}
\caption{RoBERTa-base, $N{=}100$. $d_{\mathrm{eff}} \in [1.7, 3.9]$, $d_{95} \in [57, 82]$. Values are nearly identical to those at $N{=}50$, confirming insensitivity to dataset size.}
\label{fig:app-thm1-n100}
\end{figure}

\begin{figure}[ht!]
\centering
\includegraphics[width=\linewidth]{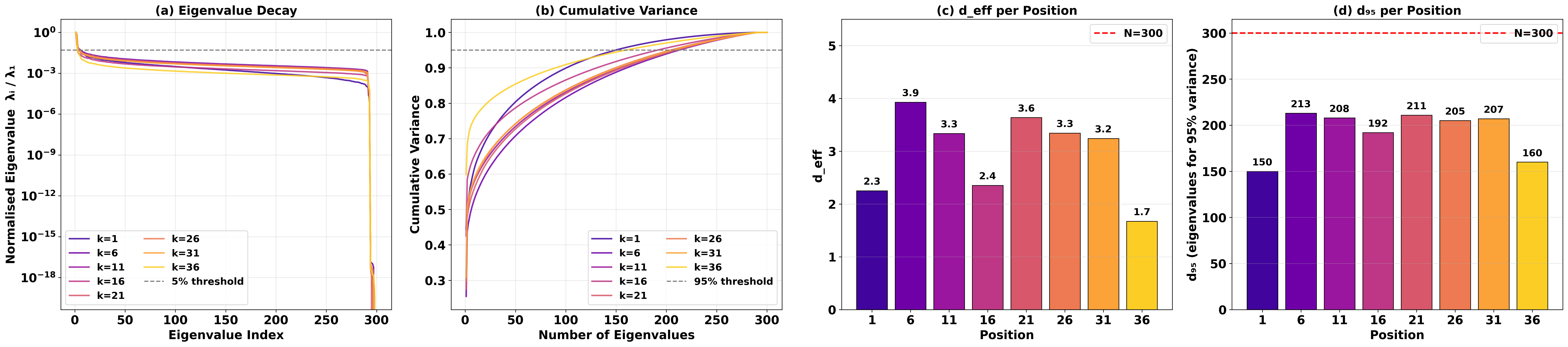}
\caption{RoBERTa-base, $N{=}200$. $d_{\mathrm{eff}} \in [1.9, 3.9]$, $d_{95} \in [100, 215]$.}
\label{fig:app-thm1-n200}
\end{figure}

\begin{figure}[ht!]
\centering
\includegraphics[width=\linewidth]{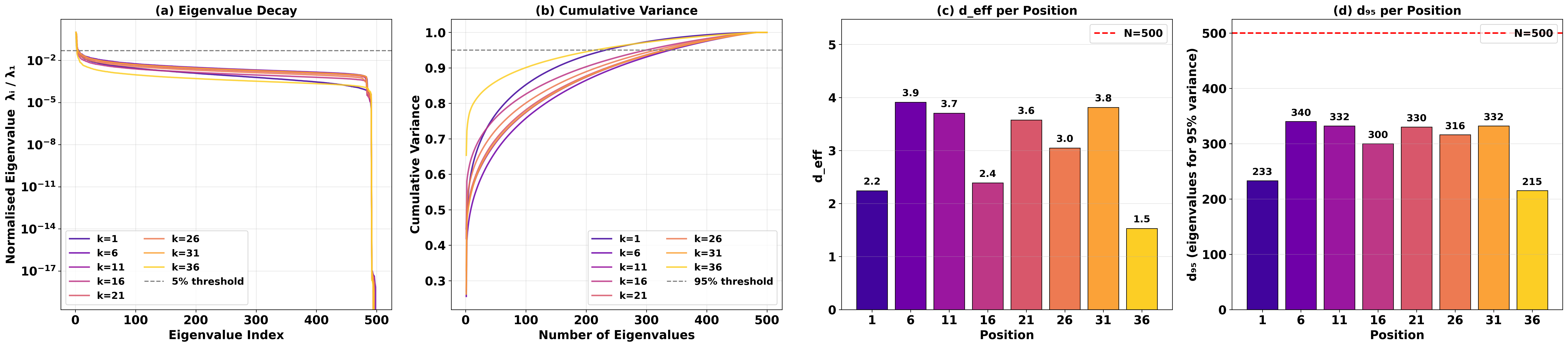}
\caption{RoBERTa-base, $N{=}500$. $d_{\mathrm{eff}} \in [1.5, 3.9]$, $d_{95} \in [215, 340]$. Reproduced in the main text as Figure~\ref{fig:thm1}.}
\label{fig:app-thm1-n500}
\end{figure}

\begin{figure}[ht!]
\centering
\includegraphics[width=\linewidth]{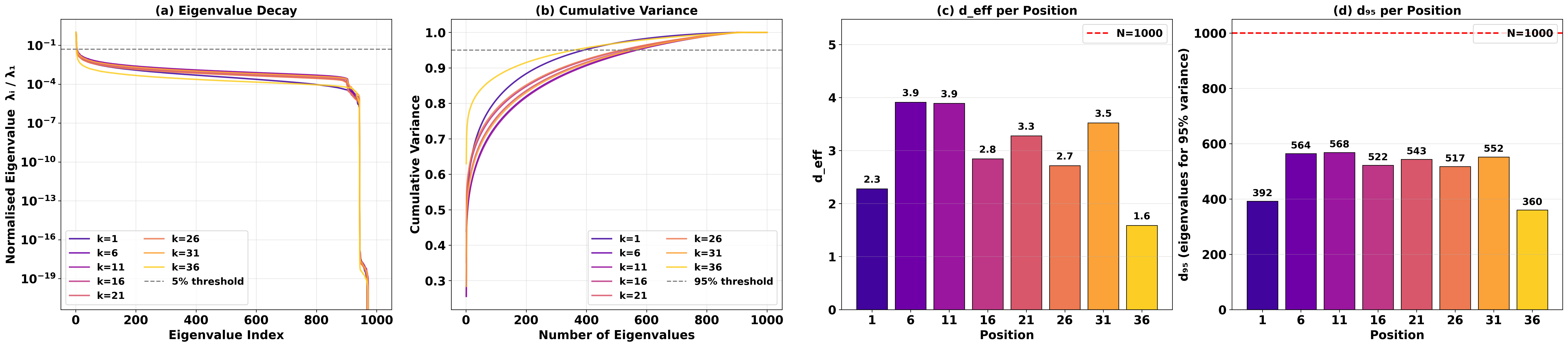}
\caption{RoBERTa-base, $N{=}1000$. $d_{\mathrm{eff}} \in [1.6, 3.9]$, $d_{95} \in [392, 568]$. A tenfold increase from $N{=}100$ produces no meaningful change in $d_{\mathrm{eff}}$.}
\label{fig:app-thm1-n1000}
\end{figure}

\begin{figure}[ht!]
\centering
\includegraphics[width=\linewidth]{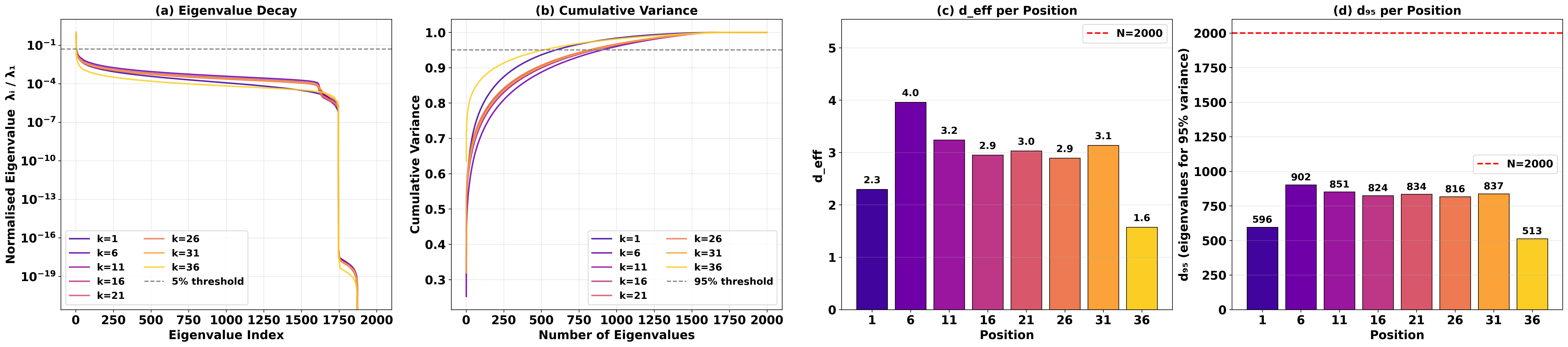}
\caption{RoBERTa-base, $N{=}2000$. $d_{\mathrm{eff}} \in [1.6, 4.0]$, $d_{95} \in [513, 902]$. At the largest tested size, $d_{\mathrm{eff}}$ stays below $4.0$ and $d_{95}/N < 0.46$.}
\label{fig:app-thm1-n2000}
\end{figure}

\paragraph{RoBERTa-base across dataset sizes.}
Figures~\ref{fig:app-thm1-n30}--\ref{fig:app-thm1-n2000} show results for RoBERTa-base on SST-2 sentences tokenized to length $T{=}50$, at positions $t \in \{1, 6, 11, 16, 21, 26, 31, 36\}$. Tokens at each position vary across samples (the fixed-token setup of Theorem~\ref{thm:equal} is evaluated separately in Appendix~\ref{app:exp-thm2}). Across all tested dataset sizes and positions, the normalized eigenvalue spectrum drops by more than ten orders of magnitude within the first $10$--$20$ indices, and $d_{\mathrm{eff}}$ lies in the range $[1.5, 4.0]$ at every configuration. Table~\ref{tab:deff-roberta} summarizes the effective-rank values per position and per $N$. Despite $N$ growing by a factor of $67$ (from $30$ to $2000$), $d_{\mathrm{eff}}$ barely changes---confirming that the capacity bottleneck is a structural property of the pretrained kernel at initialization, not a finite-sample artifact.

\begin{table}[ht!]
\centering
\small
\caption{Effective rank $d_{\mathrm{eff}}(K_t)$ for RoBERTa-base across dataset sizes $N$ and positions $t$. Values rounded to one decimal place. All entries are more than an order of magnitude below $N$.}
\label{tab:deff-roberta}
\begin{tabular}{lccccccc}
\hline
$N$ & $t{=}1$ & $t{=}6$ & $t{=}11$ & $t{=}16$ & $t{=}21$ & $t{=}26$ & $t{=}31$ \\
\hline
30   & 2.0 & 3.0 & 2.6 & 2.1 & 3.0 & 2.0 & 1.9 \\
50   & 2.1 & 2.8 & 3.5 & 1.9 & 3.2 & 3.3 & 2.8 \\
100  & 2.2 & 3.5 & 3.6 & 2.3 & 3.7 & 3.3 & 3.9 \\
200  & 2.2 & 3.9 & 3.3 & 2.4 & 3.6 & 3.3 & 3.2 \\
500  & 2.2 & 3.9 & 3.7 & 2.4 & 3.6 & 3.0 & 3.8 \\
1000 & 2.3 & 3.9 & 3.9 & 2.8 & 3.3 & 2.7 & 3.5 \\
2000 & 2.3 & 4.0 & 3.2 & 2.9 & 3.0 & 2.9 & 3.1 \\
\hline
\end{tabular}
\end{table}

\begin{figure}[ht!]
\centering
\includegraphics[width=\linewidth]{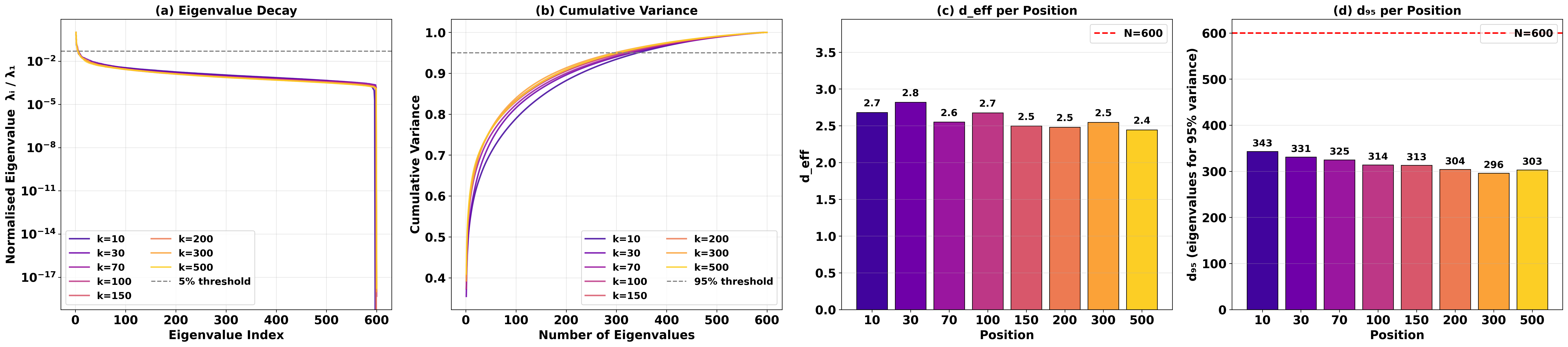}
\caption{SmolLM2-135M (causal, $135$M parameters), $N{=}600$, positions $t \in \{10, 30, 70, 100, 150, 200, 300, 500\}$. $d_{\mathrm{eff}} \in [2.4, 2.8]$, $d_{95} \in [296, 343]$.}
\label{fig:app-thm1-smoll}
\end{figure}

\paragraph{SmolLM2-135M: causal model at comparable scale.}
Figure~\ref{fig:app-thm1-smoll} validates Theorem~\ref{thm:capacity} on SmolLM2-135M, a causal $135$M-parameter model, at $N{=}600$ and positions $t \in \{10, 30, 70, 100, 150, 200, 300, 500\}$. The $d_{\mathrm{eff}}$ values lie in a tight range $[2.4, 2.8]$---more concentrated than for RoBERTa, with less position-to-position variation. The normalized spectrum collapses by more than fourteen orders of magnitude within the first few indices, and $d_{95}$ stays in $[296, 343]$, well below $N{=}600$. The capacity bottleneck therefore holds equally for causal architectures, and at a size comparable to RoBERTa the two models produce comparable effective ranks despite their different attention directionality.

\begin{figure}[ht!]
\centering
\includegraphics[width=\linewidth]{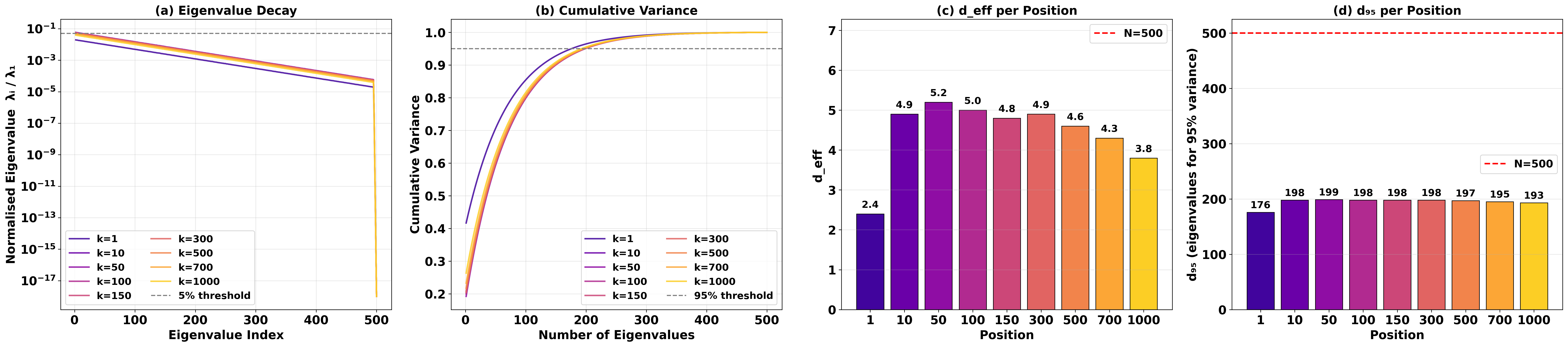}
\caption{LLaMA~3-8B (causal, $8$B parameters), $N{=}500$, positions $t \in \{1, 10, 50, 100, 150, 300, 500, 700, 1000\}$. $d_{\mathrm{eff}} \in [2.4, 5.2]$, $d_{95} \in [176, 199]$. An 8B-parameter model fine-tuned through a single position admits fewer than six effective degrees of freedom.}
\label{fig:app-thm1-llama}
\end{figure}

\paragraph{LLaMA~3-8B: large-scale validation.}
Figure~\ref{fig:app-thm1-llama} shows the same analysis on LLaMA~3-8B ($8$B parameters, causal) at $N{=}500$ and positions $t \in \{1, 10, 50, 100, 150, 300, 500, 700, 1000\}$, spanning a sequence of length $T{=}1024$. Despite a $64\times$ increase in parameter count over RoBERTa-base, $d_{\mathrm{eff}}$ remains in the narrow range $[2.4, 5.2]$, and $d_{95}$ is strikingly uniform across positions ($176$--$199$). The uniformity of $d_{95}$ across positions is a further empirical signal that the final-layer hidden-state geometry of LLaMA~3-8B is nearly position-independent---an observation that provides independent support for the aligned-geometry assumption of Theorem~\ref{thm:equal} in the causal setting.

\paragraph{Summary.}
Across all three models, all tested positions, and all dataset sizes, the three core predictions of Theorem~\ref{thm:capacity} are confirmed. First, the achievable prediction vectors lie in the low-dimensional affine subspace $o_t(X; \theta_0) + \mathrm{Im}(K_t)$ with $\mathrm{rank}(K_t) \ll N$. Second, $d_{\mathrm{eff}}(K_t)$ is determined by the pretrained kernel spectrum rather than by parameter count: it stays below $6$ regardless of whether the model has $125$M, $135$M, or $8$B parameters. Third, the capacity ceiling is set at initialization and is a structural property of the position-specific kernel, independent of the fine-tuning task, learning rate, or number of gradient steps.

\newpage

\section{Equal Capacity Across Tokens: Proof and Validation}

\subsection{Proof of Theorem~\ref{thm:equal}}
\label{app:proof-equal}

Recall from Section~\ref{sec:thm2} that in the output-head NTK regime the feature map reduces to the final-layer hidden state, $\phi_t(x) = h_t^{(D)}(x) \in \mathbb{R}^d$, and the Gram matrix and kernel-evaluation vector are computed from forward passes as $K_t[n,m] = \langle h_t^{(D)}(x^{(n)}), h_t^{(D)}(x^{(m)}) \rangle$ and $k_t(x)[n] = \langle h_t^{(D)}(x), h_t^{(D)}(x^{(n)}) \rangle$. By Theorem~\ref{thm:capacity}, for either choice $t \in \{i, j\}$ the converged predictor is $f_t(x) = o_t(x; \theta_0) + k_t(x)^\top K_t^{+} r_t$, with $r_t := y - o_t(X; \theta_0)$. For brevity we write $R := \|r_i\|_2$ and $B_j(x) := \|k_j(x)\|_2$ throughout the proof.

Before the main argument we record two standard facts about the Moore--Penrose pseudoinverse of a symmetric positive semidefinite $K \in \mathbb{R}^{N \times N}$. Let $K = U \Lambda U^\top$ be its eigendecomposition, with $U = [U_r \mid U_0]$ orthogonal, $U_r \in \mathbb{R}^{N \times r}$ collecting the eigenvectors of the $r = \mathrm{rank}(K)$ nonzero eigenvalues $\lambda_1 \geq \cdots \geq \lambda_r > 0$ (gathered in $\Lambda_r := \mathrm{diag}(\lambda_1, \ldots, \lambda_r)$), and $U_0 \in \mathbb{R}^{N \times (N - r)}$ spanning $\ker(K)$. The pseudoinverse is $K^{+} = U_r \Lambda_r^{-1} U_r^\top$.

\emph{Fact (i): operator-norm equality.} $K^{+}$ is symmetric with nonzero eigenvalues $\{1/\lambda_1, \ldots, 1/\lambda_r\}$, so
\[
    \|K^{+}\|_{\mathrm{op}} \;=\; \frac{1}{\lambda_r} \;=\; \frac{1}{\lambda_{\min}^{+}(K)}.
\]

\emph{Fact (ii): $K^{+}$ annihilates $\ker(K)$.} Let $\Pi := U_r U_r^\top$ be the orthogonal projector onto $\mathrm{Im}(K)$. Using $U_r^\top U_r = I_r$,
\[
    K^{+}(I - \Pi)
    \;=\; U_r \Lambda_r^{-1} U_r^\top (I - U_r U_r^\top)
    \;=\; U_r \Lambda_r^{-1} (U_r^\top - U_r^\top U_r U_r^\top)
    \;=\; U_r \Lambda_r^{-1} (U_r^\top - U_r^\top)
    \;=\; 0,
\]
so $K^{+} = K^{+} \Pi$.

Writing $\lambda_0^{+} := \min(\lambda_{\min}^{+}(K_i),\, \lambda_{\min}^{+}(K_j))$, fact (i) gives $\|K_t^{+}\|_{\mathrm{op}} \leq 1/\lambda_0^{+}$ for $t \in \{i, j\}$.

\noindent
\textbf{Theorem~\ref{thm:equal} (restated).} \emph{Under the output-head NTK regime with scalar output ($C = 1$), suppose the fixed-token condition $x_i^{(n)} = s_i$ and $x_j^{(n)} = s_j$ holds for all $n$, and suppose $K_i$ and $K_j$ share the same column space,}
\begin{equation}
\label{eq:common-image}
    \mathrm{Im}(K_i) \;=\; \mathrm{Im}(K_j) \;=:\; \mathcal{W},
\end{equation}
\emph{with common orthogonal projector $\Pi$ onto $\mathcal{W}$ and $\lambda_0^{+} > 0$. Then for all test inputs $x$ with $x_i = s_i$ and $x_j = s_j$,}
\begin{equation}
\label{eq:measurable-bound}
    \|f_i(x) - f_j(x)\|_2
    \;\leq\;
    \delta_{\mathrm{init}}(x)
    \;+\; \frac{\delta_k(x)\, R}{\lambda_0^{+}}
    \;+\; \frac{B_j(x)\, \delta_K\, R}{(\lambda_0^{+})^2}
    \;+\; \frac{B_j(x)\, \delta_r}{\lambda_0^{+}}.
\end{equation}

The fixed-token hypothesis is not invoked directly in the proof; rather, it is the physical condition that makes the common-image assumption \eqref{eq:common-image} plausible in a bidirectional model, where positions with aligned context produce hidden states spanning the same linear subspace.

\begin{proof}
The two converged predictors are $f_i(x) = o_i(x; \theta_0) + k_i(x)^\top K_i^{+} r_i$ and $f_j(x) = o_j(x; \theta_0) + k_j(x)^\top K_j^{+} r_j$. Subtracting and applying the triangle inequality,
\begin{equation}
\label{eq:triangle}
    \|f_i(x) - f_j(x)\|_2
    \;\leq\;
    \underbrace{\|o_i(x; \theta_0) - o_j(x; \theta_0)\|_2}_{=\, \delta_{\mathrm{init}}(x)}
    \;+\;
    \bigl| k_i(x)^\top K_i^{+} r_i - k_j(x)^\top K_j^{+} r_j \bigr|.
\end{equation}

To bound the second term, add and subtract $k_j(x)^\top K_i^{+} r_i$:
\begin{align}
    & k_i(x)^\top K_i^{+} r_i - k_j(x)^\top K_j^{+} r_j \notag\\
    &\quad =\; (k_i(x) - k_j(x))^\top K_i^{+} r_i
               \;+\; k_j(x)^\top \bigl(K_i^{+} r_i - K_j^{+} r_j\bigr).
    \label{eq:two-terms}
\end{align}
Applying the triangle inequality and Cauchy--Schwarz to the second summand,
\begin{equation}
\label{eq:two-bound}
    \bigl| k_i(x)^\top K_i^{+} r_i - k_j(x)^\top K_j^{+} r_j \bigr|
    \;\leq\;
    \bigl| (k_i(x) - k_j(x))^\top K_i^{+} r_i \bigr|
    \;+\;
    B_j(x) \cdot \|K_i^{+} r_i - K_j^{+} r_j\|_2.
\end{equation}

\emph{Term~1.} Cauchy--Schwarz combined with fact~(i) gives
\begin{equation}
\label{eq:term1}
    \bigl| (k_i(x) - k_j(x))^\top K_i^{+} r_i \bigr|
    \;\leq\; \|k_i(x) - k_j(x)\|_2 \cdot \|K_i^{+}\|_{\mathrm{op}} \cdot \|r_i\|_2
    \;\leq\; \frac{\delta_k(x)\, R}{\lambda_0^{+}}.
\end{equation}

\emph{Term~2.} Adding and subtracting $K_j^{+} r_i$,
\begin{equation}
\label{eq:alpha-split}
    \|K_i^{+} r_i - K_j^{+} r_j\|_2
    \;\leq\; \|(K_i^{+} - K_j^{+}) r_i\|_2
             \;+\; \|K_j^{+}(r_i - r_j)\|_2.
\end{equation}
The second summand is bounded directly:
\begin{equation}
\label{eq:alpha-second}
    \|K_j^{+}(r_i - r_j)\|_2
    \;\leq\; \|K_j^{+}\|_{\mathrm{op}} \cdot \|r_i - r_j\|_2
    \;\leq\; \frac{\delta_r}{\lambda_0^{+}}.
\end{equation}

For the first summand, fact~(ii) gives $K_t^{+} = K_t^{+} \Pi$ for $t \in \{i, j\}$ (the common-image assumption \eqref{eq:common-image} makes $\Pi$ the common projector), so
\[
    (K_i^{+} - K_j^{+}) r_i
    \;=\; (K_i^{+} - K_j^{+}) \Pi r_i,
\]
with $\Pi r_i \in \mathcal{W}$. Restricted to $\mathcal{W}$, both $K_i$ and $K_j$ are strictly positive with eigenvalues at least $\lambda_0^{+}$, so the restricted operators $K_i|_{\mathcal{W}}$ and $K_j|_{\mathcal{W}}$ are invertible as maps $\mathcal{W} \to \mathcal{W}$. To see that $K_t^{+}|_{\mathcal{W}} = (K_t|_{\mathcal{W}})^{-1}$ on $\mathcal{W}$, write $K_t = U_r^{(t)} \Lambda_r^{(t)} (U_r^{(t)})^\top$ with $\mathrm{span}(U_r^{(t)}) = \mathcal{W}$. For any $v \in \mathcal{W}$,
\[
    K_t K_t^{+} v
    \;=\; U_r^{(t)} \Lambda_r^{(t)} (U_r^{(t)})^\top U_r^{(t)} (\Lambda_r^{(t)})^{-1} (U_r^{(t)})^\top v
    \;=\; U_r^{(t)} (U_r^{(t)})^\top v
    \;=\; \Pi v
    \;=\; v,
\]
using $(U_r^{(t)})^\top U_r^{(t)} = I_r$ and $\Pi v = v$ for $v \in \mathcal{W}$. The analogous computation gives $K_t^{+} K_t v = v$ on $\mathcal{W}$, so $K_t^{+}|_{\mathcal{W}}$ and $K_t|_{\mathcal{W}}$ are mutual inverses on $\mathcal{W}$.

For any two invertible operators $A, B$ on $\mathcal{W}$, the first-order resolvent identity
\begin{equation}
\label{eq:resolvent}
    A^{-1} - B^{-1} \;=\; A^{-1}(B - A) B^{-1}
\end{equation}
follows from $A^{-1}(B - A)B^{-1} = A^{-1} B B^{-1} - A^{-1} A B^{-1} = A^{-1} - B^{-1}$. Applying \eqref{eq:resolvent} with $A = K_i|_{\mathcal{W}}$, $B = K_j|_{\mathcal{W}}$, and evaluating both sides at $\Pi r_i \in \mathcal{W}$:
\[
    (K_i^{+} - K_j^{+}) \Pi r_i
    \;=\; K_i^{+} (K_j - K_i) K_j^{+} \Pi r_i.
\]
Taking norms, applying submultiplicativity, and using $\|\Pi r_i\|_2 \leq \|r_i\|_2 = R$,
\begin{equation}
\label{eq:alpha-first}
    \|(K_i^{+} - K_j^{+}) r_i\|_2
    \;=\; \|(K_i^{+} - K_j^{+}) \Pi r_i\|_2
    \;\leq\; \|K_i^{+}\|_{\mathrm{op}} \cdot \|K_j - K_i\|_{\mathrm{op}} \cdot \|K_j^{+}\|_{\mathrm{op}} \cdot \|\Pi r_i\|_2
    \;\leq\; \frac{\delta_K\, R}{(\lambda_0^{+})^2}.
\end{equation}
Substituting \eqref{eq:alpha-second} and \eqref{eq:alpha-first} into \eqref{eq:alpha-split},
\begin{equation}
\label{eq:term2}
    \|K_i^{+} r_i - K_j^{+} r_j\|_2
    \;\leq\; \frac{\delta_K\, R}{(\lambda_0^{+})^2} \;+\; \frac{\delta_r}{\lambda_0^{+}}.
\end{equation}

Substituting \eqref{eq:term1} and \eqref{eq:term2} into \eqref{eq:two-bound} and then into \eqref{eq:triangle},
\[
    \|f_i(x) - f_j(x)\|_2
    \;\leq\;
    \delta_{\mathrm{init}}(x)
    \;+\; \frac{\delta_k(x)\, R}{\lambda_0^{+}}
    \;+\; B_j(x)\!\left( \frac{\delta_K\, R}{(\lambda_0^{+})^2} + \frac{\delta_r}{\lambda_0^{+}} \right),
\]
which rearranges to \eqref{eq:measurable-bound}.
\end{proof}

\subsection{Empirical Validation of Theorem~\ref{thm:equal}}
\label{app:exp-thm2}

We validate Theorem~\ref{thm:equal} empirically using RoBERTa-base (125M parameters) on SST-2 sentences. The theorem predicts that two positions $i$ and $j$ sharing fixed anchor tokens in a bidirectional model yield approximately equal fine-tuning capacity, with the prediction gap $\|f_i(x) - f_j(x)\|_2$ controlled by the four measurable quantities $\delta_{\mathrm{init}}(x), \delta_k(x), \delta_K, \delta_r$ of Theorem~\ref{thm:equal}. We measure the effective rank $d_{\mathrm{eff}}(K_t) = \mathrm{tr}(K_t) / \|K_t\|_{\mathrm{op}}$ and the $d_{95}$ metric of the output-head NTK Gram matrix at each fixed position $t$, computed at the pretrained initialization $\theta_0$ before any fine-tuning. We run three complementary experiments: two that confirm the theorem's geometric hypotheses by placing fixed tokens at opposite ends of the sequence, and one that confirms the theorem's conclusion by measuring the prediction gap between fine-tuned predictors directly.

\paragraph{Experiment 1: Fixed prefix prompt (right padding).}
We prepend the fixed natural-language prompt \emph{``Read the sentence properly, and determine the sentiment for this sentence''} to every input, producing sequences of the form
\[
    \texttt{[CLS]}\;
    \underbrace{t_1\; t_2\; \cdots\; t_{12}}_{\text{fixed prompt tokens}}
    \;\underbrace{w_1\; w_2\; \cdots}_{\text{varying sentence}}
    \;\texttt{[SEP]}\;\texttt{[PAD]}\cdots
\]
where the prompt tokens occupy positions $t = 1, \ldots, 12$ and are identical across all $N$ samples, while the sentence tokens at positions $t > 12$ vary freely. This mirrors the prefix-tuning setup in which a fixed prompt is prepended to every input. We measure $d_{\mathrm{eff}}$ and $d_{95}$ at all $12$ prompt positions for $N \in \{50, 300, 1000\}$.

\begin{figure}[ht!]
\centering
\includegraphics[width=\linewidth]{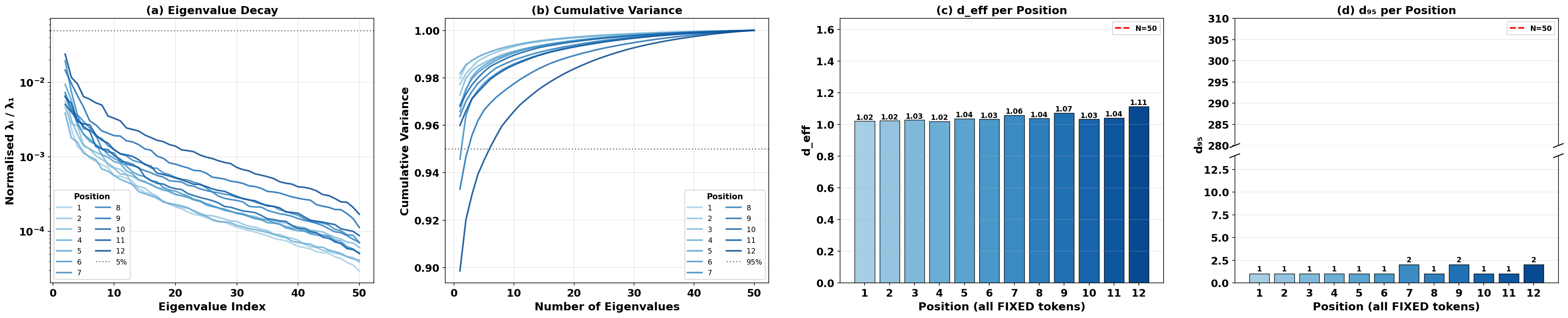}
\caption{Experiment~1 (fixed prefix prompt, right padding), $N{=}50$. $d_{\mathrm{eff}} \in [1.02, 1.11]$, $\mathrm{CV} = 0.02$; $d_{95} = 1$ at all but two positions. Eigenvalue decay and cumulative variance curves are nearly identical across the $12$ fixed prompt positions.}
\label{fig:app-thm2-exp1-n50}
\end{figure}

\begin{figure}[ht!]
\centering
\includegraphics[width=\linewidth]{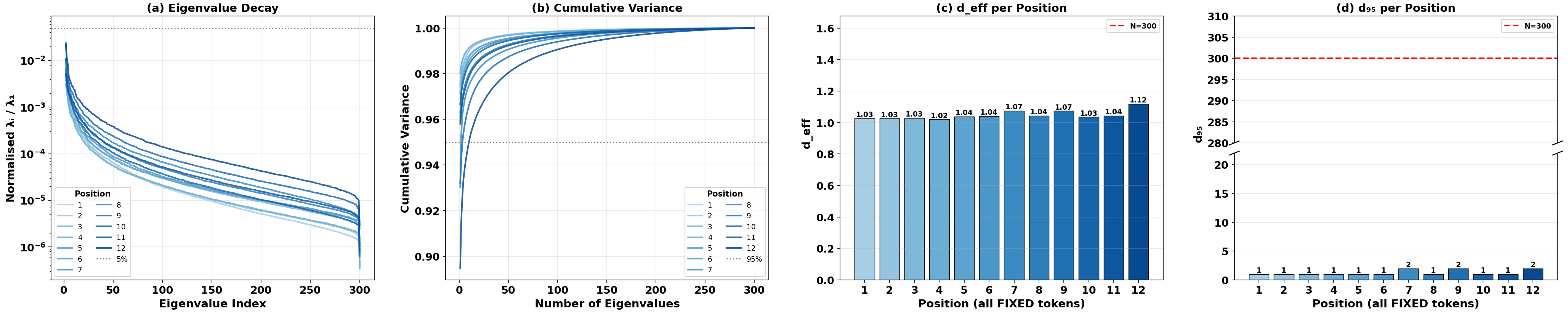}
\caption{Experiment~1 (fixed prefix prompt, right padding), $N{=}300$. $d_{\mathrm{eff}} \in [1.02, 1.12]$, $\mathrm{CV} = 0.02$; $d_{95} = 1$ at all but two positions. The pattern is essentially unchanged from $N{=}50$.}
\label{fig:app-thm2-exp1-n300}
\end{figure}

\begin{figure}[ht!]
\centering
\includegraphics[width=\linewidth]{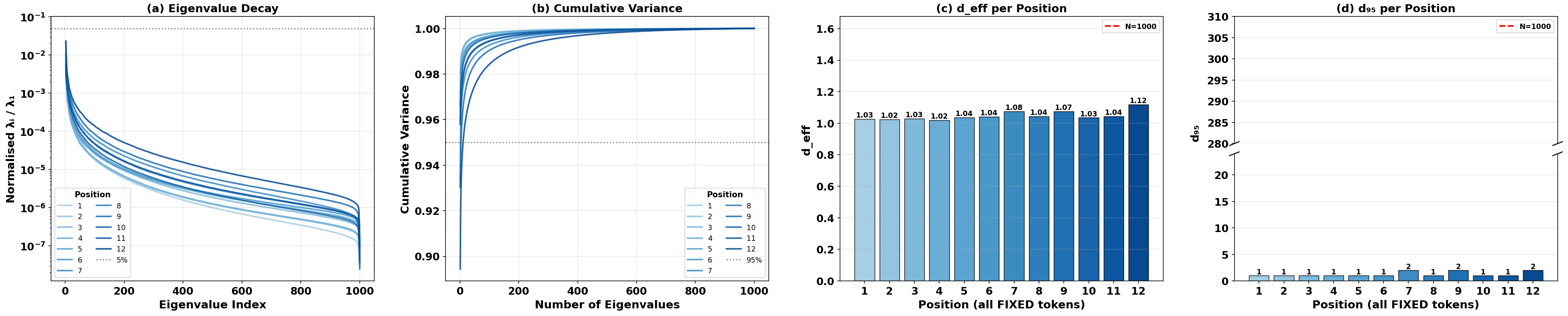}
\caption{Experiment~1 (fixed prefix prompt, right padding), $N{=}1000$. $d_{\mathrm{eff}} \in [1.02, 1.12]$, $\mathrm{CV} = 0.02$; $d_{95} = 1$ at all but two positions. Despite a twenty-fold increase in $N$ from $50$, the effective ranks are numerically identical.}
\label{fig:app-thm2-exp1-n1000}
\end{figure}

Figures~\ref{fig:app-thm2-exp1-n50}, \ref{fig:app-thm2-exp1-n300}, and \ref{fig:app-thm2-exp1-n1000} show the results. Across all dataset sizes and all $12$ prompt positions, $d_{\mathrm{eff}}$ lies in $[1.00, 1.12]$ with coefficient of variation $\mathrm{CV} = \sigma/\mu \leq 0.02$. The eigenvalue decay curves and the cumulative variance curves are visually indistinguishable across the $12$ positions. $d_{95} = 1$ at almost every position across all three dataset sizes, meaning a single eigenvalue explains $95\%$ of the variance---the Gram matrix is effectively rank-one at each fixed position. The $d_{\mathrm{eff}}$ values are nearly identical across $N \in \{50, 300, 1000\}$, indicating that the observed equivalence is a structural property of the pretrained model, not a finite-sample artifact.

\paragraph{Experiment 2: Fixed suffix prompt (left padding).}
We append the fixed natural-language prompt \emph{``Based on the analysis of the given sentence, we can say the sentiment is''} to every input, producing sequences of the form
\[
    \texttt{[PAD]}\cdots\;\texttt{[CLS]}\;
    \underbrace{w_1\; w_2\; \cdots}_{\text{varying sentence}}
    \;\underbrace{t_1\; t_2\; \cdots\; t_{15}}_{\text{fixed answer prompt}}
    \;\texttt{[SEP]}
\]
where the answer prompt occupies positions $t = 48, \ldots, 62$ and is identical across all $N$ samples. This places the fixed tokens at the end of the sequence rather than the beginning, testing whether positional equivalence holds in a structurally different context. We again measure $d_{\mathrm{eff}}$ and $d_{95}$ at all $15$ suffix positions for $N \in \{50, 300, 1000\}$.

\begin{figure}[ht!]
\centering
\includegraphics[width=\linewidth]{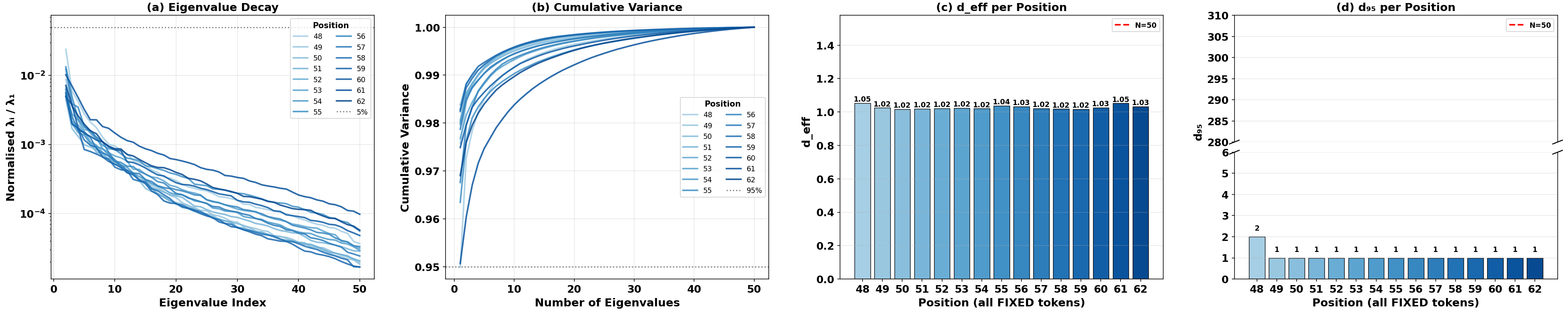}
\caption{Experiment~2 (fixed suffix prompt, left padding), $N{=}50$. $d_{\mathrm{eff}} \in [1.02, 1.05]$, $\mathrm{CV} = 0.01$; $d_{95} = 1$ at every position.}
\label{fig:app-thm2-exp2-n50}
\end{figure}

\begin{figure}[ht!]
\centering
\includegraphics[width=\linewidth]{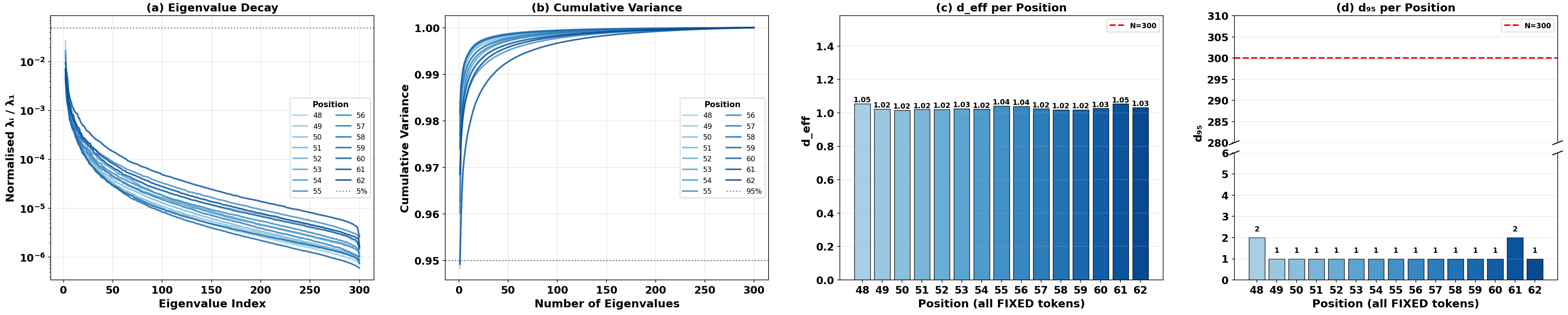}
\caption{Experiment~2 (fixed suffix prompt, left padding), $N{=}300$. $d_{\mathrm{eff}} \in [1.02, 1.05]$, $\mathrm{CV} = 0.01$; $d_{95} = 1$ at every position. Results are unchanged from $N{=}50$.}
\label{fig:app-thm2-exp2-n300}
\end{figure}

\begin{figure}[ht!]
\centering
\includegraphics[width=\linewidth]{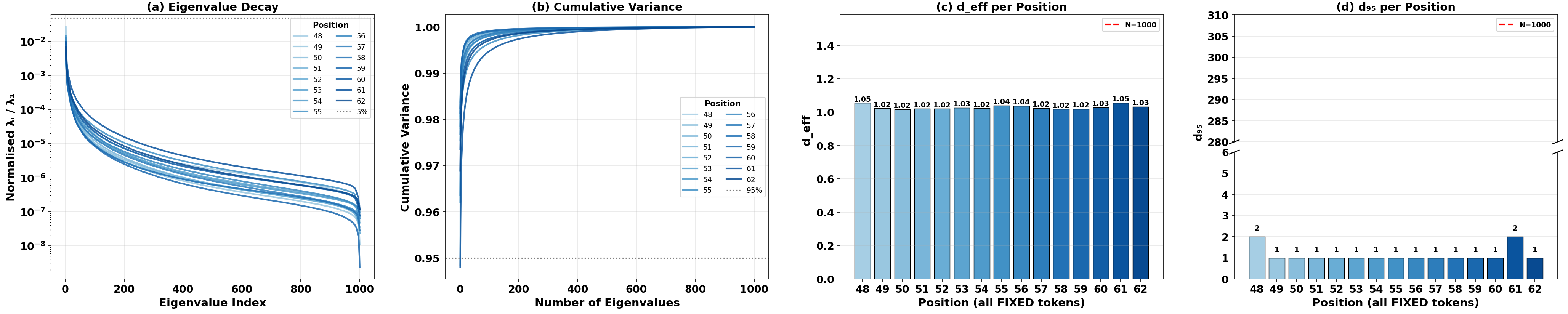}
\caption{Experiment~2 (fixed suffix prompt, left padding), $N{=}1000$. $d_{\mathrm{eff}} \in [1.02, 1.05]$, $\mathrm{CV} = 0.01$; $d_{95} = 1$ at every position.}
\label{fig:app-thm2-exp2-n1000}
\end{figure}

Figures~\ref{fig:app-thm2-exp2-n50}, \ref{fig:app-thm2-exp2-n300}, and \ref{fig:app-thm2-exp2-n1000} show the results. All $15$ suffix positions yield $d_{\mathrm{eff}} \in [1.02, 1.05]$ with $\mathrm{CV} \leq 0.01$---a tighter range than Experiment~1. The eigenvalue curves are visually indistinguishable across positions, and $d_{95} = 1$ at every position without exception across all three dataset sizes.

\paragraph{Experiment 3: Prediction-gap measurement.}
Experiments~1 and~2 confirmed the theorem's geometric hypotheses through the near-identity of eigenvalue spectra and effective ranks across fixed positions. We now measure the theorem's \emph{conclusion} directly: that the converged predictors $f_i(x)$ and $f_j(x)$ are numerically close at test inputs, not just structurally similar in feature geometry. Using the fixed prompt \emph{``Classify the sentiment of this movie review~:''} prepended to every input on RoBERTa-base, we train output-head NTK predictors $f_i$ and $f_j$ separately at each of ten pairs of positions across the prompt (spanning position-index gaps from $1$ to $7$), using $N{=}300$ SST-2 training sentences and evaluating on $50$ held-out test sentences. Table~\ref{tab:close-capacity} reports the mean prediction gap $\|f_i(x) - f_j(x)\|$ normalized by the label standard deviation $\hat{\sigma}_y = 0.5$, the fraction of test sentences for which $\|f_i(x) - f_j(x)\| < 0.5\,\hat{\sigma}_y$, and the $p$-value from a left-tail position-exchangeability permutation test with $B{=}2000$ resamples.

Across all ten pairs, the mean prediction gap ranges from $0.180$ to $0.275$ standard deviations of the label---substantially below the $0.5\,\hat{\sigma}_y$ threshold---and between $86\%$ and $98\%$ of test sentences fall below that threshold per pair. The effect is statistically significant for every pair ($p < 0.001$), indicating that the observed closeness is not a chance consequence of arbitrary position assignment but reflects a genuine equivalence in converged predictors. The prediction gap shows no systematic dependence on the position-index distance $|i-j|$: adjacent pairs and gap-$7$ pairs produce comparable gaps, consistent with the theorem's prediction that positional distance is irrelevant once the hidden-state geometry is aligned. This provides a direct downstream confirmation of the four-term bound of Theorem~\ref{thm:equal}: the measurable quantities $\delta_K, \delta_k(x), \delta_r, \delta_{\mathrm{init}}(x)$ jointly control the gap, and at the prediction level the gap is bounded uniformly across position pairs.

\begin{table}[ht!]
\centering
\caption{Prediction gaps between fine-tuned predictors $f_i$ and $f_j$ across ten token pairs within the fixed prompt \texttt{``Classify the sentiment of this movie review :''} on RoBERTa-base ($N{=}300$ SST-2 training, 50 test sentences). \emph{Gap} is the position-index distance $|i-j|$; \emph{Frac} is the fraction of test sentences with $\|f_i(x) - f_j(x)\| < 0.5\,\hat{\sigma}_y$; $p$-values are from a left-tail position-exchangeability permutation test ($B{=}2000$).}
\label{tab:close-capacity}
\small
\begin{tabular}{llcccc}
\toprule
Token $s_i$ & Token $s_j$ & Gap
    & $\|f_i - f_j\| / \hat{\sigma}_y$
    & Frac $< 0.5\hat{\sigma}_y$
    & $p$-value \\
\midrule
\multicolumn{6}{l}{\emph{Adjacent pairs (gap $=1$)}} \\[2pt]
\texttt{Classify} & \texttt{the}       & 1 & 0.188 &  96\% & ${<}0.001$ \\
\texttt{the}      & \texttt{sentiment} & 1 & 0.245 &  92\% & ${<}0.001$ \\
\texttt{sentiment}& \texttt{of}        & 1 & 0.229 &  96\% & ${<}0.001$ \\
\texttt{of}       & \texttt{this}      & 1 & 0.216 &  92\% & ${<}0.001$ \\
\midrule
\multicolumn{6}{l}{\emph{Short-range pairs (gap $=2$)}} \\[2pt]
\texttt{Classify} & \texttt{sentiment} & 2 & 0.250 &  88\% & ${<}0.001$ \\
\texttt{the}      & \texttt{of}        & 2 & 0.246 &  94\% & ${<}0.001$ \\
\texttt{sentiment}& \texttt{this}      & 2 & 0.180 &  98\% & ${<}0.001$ \\
\midrule
\multicolumn{6}{l}{\emph{Medium-range pairs (gap $=4$)}} \\[2pt]
\texttt{Classify} & \texttt{this}      & 4 & 0.275 &  86\% & ${<}0.001$ \\
\texttt{the}      & \texttt{movie}     & 4 & 0.232 &  92\% & ${<}0.001$ \\
\midrule
\multicolumn{6}{l}{\emph{Long-range pair (gap $=7$)}} \\[2pt]
\texttt{Classify} & \texttt{:}         & 7 & 0.273 &  86\% & ${<}0.001$ \\
\midrule
\multicolumn{3}{l}{\emph{All 10 pairs}} & 0.188--0.275 & 86--98\% & ${<}0.001$ \\
\bottomrule
\end{tabular}
\end{table}

\paragraph{Synthesis.}
The three experiments validate Theorem~\ref{thm:equal} at complementary levels. Experiments~1 and~2 confirm the theorem's geometric hypotheses: across the $12$ prefix positions and $15$ suffix positions respectively, the Gram matrices $K_t$ have visually indistinguishable eigenvalue spectra and effective ranks ($\mathrm{CV} \leq 0.02$), empirically supporting the common-image assumption (Eq.~\ref{eq:common-image}) and $\delta_K \approx 0$. The rank-one collapse observed in both experiments is a consequence of token fixing that goes beyond what the theorem strictly requires: when the token at position $t$ is identical across all $N$ samples, the position-$t$ gradient has one dominant direction determined by the fixed token embedding, with context variation contributing only a small perturbation through attention. This concentrates almost all variance into a single eigenvalue, making $\lambda_{\min}^{+}(K_t) \approx \lambda_1(K_t) \approx \mathrm{tr}(K_t)$ large and driving the second and third terms of the bound of Theorem~\ref{thm:equal} (divided by $(\lambda_0^{+})^2$ and $\lambda_0^{+}$) to be numerically negligible. Experiment~3 then confirms the theorem's conclusion directly: when the hypotheses hold, the converged predictors $f_i$ and $f_j$ are numerically close, with mean prediction gap below $0.28\,\hat{\sigma}_y$ and $86\%$--$98\%$ of test inputs falling within $0.5\,\hat{\sigma}_y$ across all ten tested position pairs. The consistency across early vs.\ late positions (Experiments~1 vs.~2) and across position-index distances from $1$ to $7$ (Experiment~3) confirms that positional equivalence is a global property of the pretrained model's hidden-state geometry, not an artifact of position index, padding convention, or adjacency.

\begin{table}[ht!]
\centering
\caption{Summary of geometric validation of Theorem~\ref{thm:equal} (Experiments~1 and~2). All fixed-token positions yield nearly identical $d_{\mathrm{eff}}$ values with $\mathrm{CV} \leq 0.02$ across both experiments and all dataset sizes.}
\label{tab:thm2-summary}
\begin{tabular}{llccccc}
\toprule
Experiment & Positions & $N$ &
$d_{\mathrm{eff}}$ range & Mean $d_{\mathrm{eff}}$ & CV &
$d_{95}$ (modal) \\
\midrule
Exp.~1 (prefix, right pad) & $t = 1\text{--}12$
    & 50   & $[1.02,\;1.11]$ & 1.03 & 0.02 & 1 \\
    & & 300  & $[1.02,\;1.12]$ & 1.04 & 0.02 & 1 \\
    & & 1000 & $[1.02,\;1.12]$ & 1.04 & 0.02 & 1 \\
\midrule
Exp.~2 (suffix, left pad) & $t = 48\text{--}62$
    & 50   & $[1.02,\;1.05]$ & 1.03 & 0.01 & 1 \\
    & & 300  & $[1.02,\;1.05]$ & 1.03 & 0.01 & 1 \\
    & & 1000 & $[1.02,\;1.05]$ & 1.03 & 0.01 & 1 \\
\bottomrule
\end{tabular}
\end{table}

\subsection{Token Importance in the Low-Shot Regime}
\label{app:ablation-token-importance}

The three experiments above validate Theorem~\ref{thm:equal} in the regime where its hypotheses hold with wide margin. We now probe the opposite regime, where the bound becomes loose and the theorem's equal-capacity prediction weakens.

Theorem~\ref{thm:equal} bounds the prediction gap between two fixed-token positions by $\delta_{\mathrm{init}}(x) + \delta_k(x) R / \lambda_0^{+} + B_j(x) \delta_K R /(\lambda_0^{+})^2 + B_j(x) \delta_r / \lambda_0^{+}$, with the second and third terms scaling inversely in $\lambda_0^{+} = \min_t \lambda_{\min}^{+}(K_t)$. In the generic training regime where $N$ is reasonably large, $\lambda_0^{+}$ is comfortably bounded away from zero and these terms are negligible, so random selection suffices (as reported in Section~\ref{sec:method}). In this ablation we characterize the opposite regime---very low $N$---where $\lambda_0^{+}$ can be small and the error terms of the bound become non-negligible. We ask: when the bound is loose, does the choice of which positions to train through affect accuracy in the way the spectral structure would predict?

We use the Wanda~\cite{sun2023simple} criterion---the product of weight magnitude and input activation norm---to rank token positions by how much they contribute to the model's output at initialization. Positions with larger activation norms produce Gram matrices with larger dominant eigenvalues, hence larger $\lambda_0^{+}$; positions with smaller activation norms produce Gram matrices with smaller $\lambda_0^{+}$. Theorem~\ref{thm:equal} predicts that when $\lambda_0^{+}$ is small the error terms of the bound blow up, so we expect that in low-shot training, high-$\lambda_0^{+}$ positions (Most Important) should outperform low-$\lambda_0^{+}$ positions (Least Important), with the gap shrinking as $M$ increases and the combined kernel $K^{(M)}$ aggregates gradient signal across positions.

We test this on MRPC (paraphrase identification) and QQP (question-pair similarity) using RoBERTa-base, selecting the top-$M$, bottom-$M$, and a uniformly random $M$ positions for $M \in \{1, 2, 3, 5, 10\}$. Accuracy is reported over $10$ random seeds with error bars showing one standard deviation.

\begin{figure}[ht!]
\centering
\includegraphics[width=\linewidth]{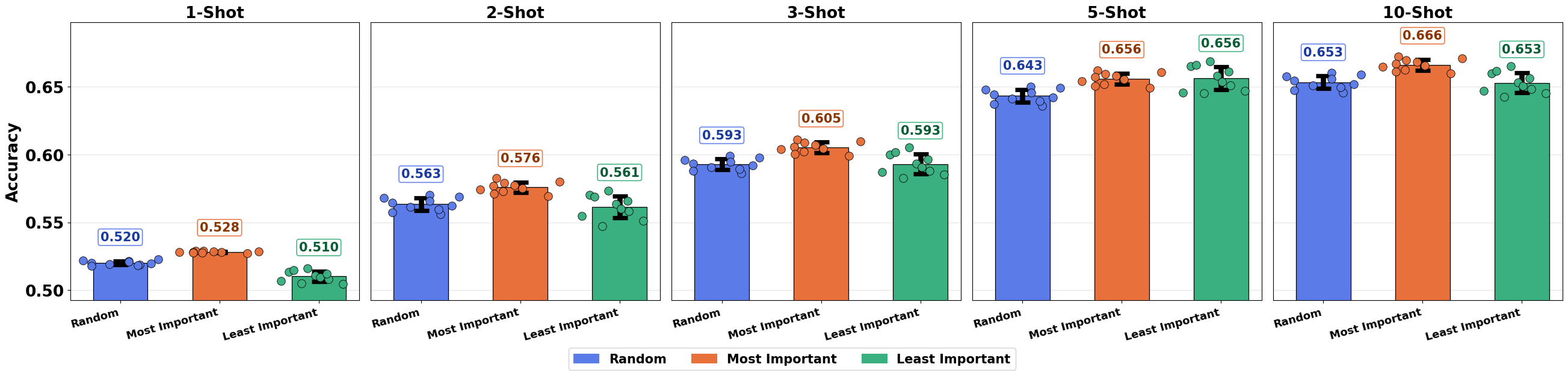}
\caption{Token importance ablation on MRPC (RoBERTa-base, $10$ seeds, error bars = $1\sigma$). Most Important (high $\lambda_{\min}^{+}(K_t)$) consistently outperforms Random and Least Important. The gap is largest at $M = 1$ ($0.018$) where the bound of Theorem~\ref{thm:equal} is loosest, and narrows as $M$ grows and $K^{(M)}$ aggregates gradient signal.}
\label{fig:ablation-mrpc}
\end{figure}

\begin{figure}[ht!]
\centering
\includegraphics[width=\linewidth]{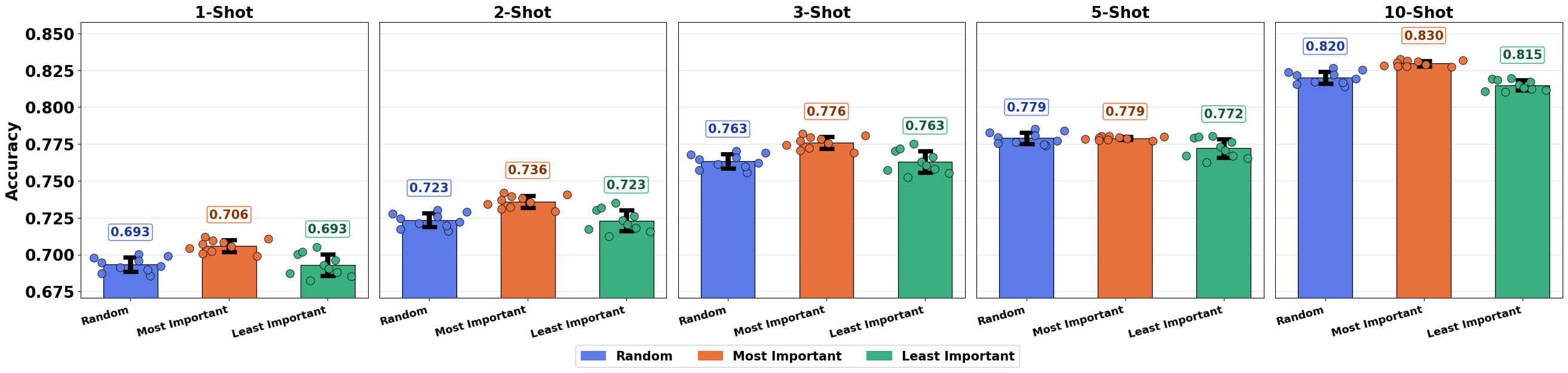}
\caption{Token importance ablation on QQP (RoBERTa-base, $10$ seeds, error bars = $1\sigma$). Same ordering as MRPC: Most Important $\geq$ Random $\geq$ Least Important. At $M = 1$ the gap is $0.013$; the ordering is preserved but with comparable magnitude at $M = 10$, consistent with the conditioning improvement predicted by spectral aggregation.}
\label{fig:ablation-qqp}
\end{figure}

Figures~\ref{fig:ablation-mrpc} and~\ref{fig:ablation-qqp} show the results. Across both datasets the ordering Most Important~$\geq$~Random~$\geq$~Least Important holds at every $M$. At $M = 1$, the gap between Most and Least Important is $0.018$ on MRPC and $0.013$ on QQP; at $M = 10$, the gap narrows to roughly $0.013$ on MRPC and $0.015$ on QQP. The narrowing is consistent with the theorem's structure: as $M$ grows, the combined kernel $K^{(M)}$ inherits $\lambda^{+}_{\min}(K^{(M)}) \geq \sum_{m=1}^{M} \lambda^{+}_{\min}(K_{t_m})$, so the effective conditioning improves and the per-position importance matters less.

\paragraph{Interpretation.}
These findings refine rather than contradict the main theorem. Theorem~\ref{thm:equal} says that positions with aligned hidden-state geometry produce equivalent predictors \emph{when the error terms of the bound are small}, which in practice holds once $N$ is not pathologically small. The low-shot experiments probe the regime where the error terms become non-trivial---exactly the regime where the theorem predicts the $\lambda_0^{+}$ structure matters---and the empirical ordering matches: positions with larger $\lambda_0^{+}$ (Most Important) yield better fine-tuning, and the advantage shrinks as $M$ grows. In the generic large-$N$ regime used for all other experiments in this paper (Section~\ref{sec:experiments}), these effects are dominated by the primary capacity bottleneck of Theorem~\ref{thm:capacity}, and random position selection is both theoretically adequate and empirically sufficient (Section~\ref{sec:method}).

\clearpage

\section{Monotone Output-Head NTK}

\subsection{Proof of Theorem~\ref{thm:monotone}}
\label{app:proof-monotone}

Recall from Section~\ref{sec:thm3} that we work in the output-head NTK regime where the feature map reduces to the final-layer hidden state, $\phi_t(x) = h_t^{(D)}(x) \in \mathbb{R}^d$, and the output-head Gram matrix is $K_t[n, m] = \langle h_t^{(D)}(x^{(n)}), h_t^{(D)}(x^{(m)}) \rangle$, directly computable from forward passes. We restate Theorem~\ref{thm:monotone} for convenience before giving the proof.

\noindent
\textbf{Theorem~\ref{thm:monotone} (restated).} \emph{Suppose the final-layer hidden states of a causal pretrained transformer satisfy the adjacent-state contraction condition: for each position $t \in \{1, \ldots, T-1\}$ there exists a single matrix $A_t \in \mathbb{R}^{d \times d}$ with $\|A_t\|_{\mathrm{op}} \leq 1$, shared uniformly across all inputs $x$, and residual vectors $\varepsilon_t(x) \in \mathbb{R}^d$, such that}
\[
    h_t^{(D)}(x) \;=\; A_t\, h_{t+1}^{(D)}(x) + \varepsilon_t(x).
\]
\emph{Define the pointwise bounds}
\[
    \sigma_t \;:=\; \max_{n = 1, \ldots, N} \|\varepsilon_t(x^{(n)})\|_2,
    \qquad
    \beta_t \;:=\; \max_{n = 1, \ldots, N} \|h_t^{(D)}(x^{(n)})\|_2,
\]
\emph{both computable directly from the data. Then the output-head Gram matrices satisfy the approximate positive semidefinite ordering}
\begin{equation}
\label{eq:approx-psd-ordering}
    K_{t+1} \;\succeq\; K_t - \delta_t\, I_N,
    \qquad
    \delta_t \;:=\; N \sigma_t (2 \beta_{t+1} + \sigma_t),
\end{equation}
\emph{for every $t \in \{1, \ldots, T-1\}$.}

The adjacent-state contraction condition is a regularity property of the joint statistics of final-layer hidden states at adjacent positions, consistent with the way causal attention couples them: at every layer, the update at position $t{+}1$ reads from positions $1, \ldots, t{+}1$, including position $t$, so adjacent hidden states share lineage rather than being independently produced. The direction $h_t^{(D)} = A_t h_{t+1}^{(D)} + \varepsilon_t$ should be read as a description of how the learned representations at adjacent positions are geometrically related, not as a forward-pass causal computation. The substantive restriction is the uniformity of $A_t$ across inputs: it would fail if the attention-mediated relationship between $h_t^{(D)}(x)$ and $h_{t+1}^{(D)}(x)$ varied significantly with $x$. We define $\sigma_t$ via the maximum over samples (rather than an average) because the proof uses $\|E_t\|_F^2 \leq N \sigma_t^2$, which requires a uniform-over-samples bound. We empirically validate this assumption in Appendix~\ref{app:exp-thm3}: composing the model's own (frozen) causal attention at the $(t, t{+}1)$ pair with a trainable linear map, we find $\|A_t\|_{\mathrm{op}} \approx 1$ across the bulk of layers in SmolLM2-135M.

\begin{proof}
Define the hidden-state matrix $H_t := [h_t^{(D)}(x^{(1)}), \ldots, h_t^{(D)}(x^{(N)})] \in \mathbb{R}^{d \times N}$ and the residual matrix $E_t := [\varepsilon_t(x^{(1)}), \ldots, \varepsilon_t(x^{(N)})] \in \mathbb{R}^{d \times N}$. Since $A_t$ is the same matrix for all inputs, applying the contraction assumption column-by-column gives the matrix identity
\[
    H_t \;=\; A_t H_{t+1} + E_t,
\]
whose $n$-th column is $A_t h_{t+1}^{(D)}(x^{(n)}) + \varepsilon_t(x^{(n)}) = h_t^{(D)}(x^{(n)})$. The uniformity of $A_t$ across inputs is what allows this single matrix equation to hold; if $A_t$ depended on $x$, the factorization would break. The output-head Gram matrix is $K_t = H_t^\top H_t$, so for any $v \in \mathbb{R}^N$,
\[
    v^\top K_t v
    \;=\; \|H_t v\|_2^2
    \;=\; \|A_t H_{t+1} v + E_t v\|_2^2.
\]
Expanding using $\|a + b\|_2^2 = \|a\|_2^2 + 2\langle a, b \rangle + \|b\|_2^2$ with $a := A_t H_{t+1} v \in \mathbb{R}^d$ and $b := E_t v \in \mathbb{R}^d$,
\begin{equation}
\label{eq:expansion}
    v^\top K_t v
    \;=\; \underbrace{\|A_t H_{t+1} v\|_2^2}_{T_1}
    \;+\; \underbrace{2 \langle A_t H_{t+1} v,\, E_t v \rangle}_{T_2}
    \;+\; \underbrace{\|E_t v\|_2^2}_{T_3}.
\end{equation}
We bound each term in turn.

\medskip\noindent\emph{Bounding $T_1$.} Since $\|A_t\|_{\mathrm{op}} \leq 1$, for any vector $w \in \mathbb{R}^d$ we have $\|A_t w\|_2 \leq \|A_t\|_{\mathrm{op}} \|w\|_2 \leq \|w\|_2$. Applying this with $w = H_{t+1} v$,
\[
    T_1 \;=\; \|A_t H_{t+1} v\|_2^2
    \;\leq\; \|H_{t+1} v\|_2^2
    \;=\; v^\top H_{t+1}^\top H_{t+1} v
    \;=\; v^\top K_{t+1} v.
\]

\medskip\noindent\emph{Bounding $T_2$.} Since $T_2$ is real, $T_2 \leq |T_2|$ trivially. This step is conservative when $T_2 < 0$, in which case $\delta_t$ is an over-estimate (the true quadratic form is smaller than what we bound); but in all cases the bound remains valid. By the Cauchy--Schwarz inequality,
\[
    |T_2|
    \;=\; 2 |\langle A_t H_{t+1} v,\, E_t v \rangle|
    \;\leq\; 2 \|A_t H_{t+1} v\|_2 \cdot \|E_t v\|_2.
\]
We bound each factor. Applying the contraction $\|A_t\|_{\mathrm{op}} \leq 1$,
\[
    \|A_t H_{t+1} v\|_2 \;\leq\; \|H_{t+1} v\|_2.
\]
For the remaining operator-norm bounds we use $\|M v\|_2 \leq \|M\|_{\mathrm{op}} \|v\|_2 \leq \|M\|_F \|v\|_2$, which combines the operator-Frobenius inequality $\|M\|_{\mathrm{op}} \leq \|M\|_F$ with the standard operator-norm bound. This gives
\[
    \|H_{t+1} v\|_2 \;\leq\; \|H_{t+1}\|_F \|v\|_2,
    \qquad
    \|E_t v\|_2 \;\leq\; \|E_t\|_F \|v\|_2.
\]
The Frobenius norms are controlled by the pointwise bounds:
\[
    \|H_{t+1}\|_F^2 \;=\; \sum_{n=1}^{N} \|h_{t+1}^{(D)}(x^{(n)})\|_2^2
    \;\leq\; N \beta_{t+1}^2,
    \qquad
    \|E_t\|_F^2 \;=\; \sum_{n=1}^{N} \|\varepsilon_t(x^{(n)})\|_2^2
    \;\leq\; N \sigma_t^2.
\]
Hence $\|H_{t+1} v\|_2 \leq \sqrt{N} \beta_{t+1} \|v\|_2$ and $\|E_t v\|_2 \leq \sqrt{N} \sigma_t \|v\|_2$. This choice of the Frobenius bound (rather than the tighter $\|H_{t+1}\|_{\mathrm{op}} = \sqrt{\lambda_{\max}(K_{t+1})}$) is deliberate: it keeps $\delta_t$ expressible purely in terms of $\beta_{t+1}$ and $\sigma_t$, avoiding circular dependence on the spectrum of $K_{t+1}$. Combining the two factor bounds,
\[
    T_2 \;\leq\; |T_2|
    \;\leq\; 2 \cdot \sqrt{N} \beta_{t+1} \|v\|_2 \cdot \sqrt{N} \sigma_t \|v\|_2
    \;=\; 2 N \beta_{t+1} \sigma_t \|v\|_2^2.
\]

\medskip\noindent\emph{Bounding $T_3$.} Squaring the bound on $\|E_t v\|_2$ established above,
\[
    T_3 \;=\; \|E_t v\|_2^2
    \;\leq\; N \sigma_t^2 \|v\|_2^2.
\]

\medskip\noindent\emph{Combining.} Substituting the three bounds into \eqref{eq:expansion},
\[
    v^\top K_t v
    \;\leq\; v^\top K_{t+1} v
    + 2 N \beta_{t+1} \sigma_t \|v\|_2^2
    + N \sigma_t^2 \|v\|_2^2
    \;=\; v^\top K_{t+1} v
    + N \sigma_t (2 \beta_{t+1} + \sigma_t) \|v\|_2^2.
\]
Rearranging,
\[
    v^\top \bigl[K_{t+1} - K_t + N \sigma_t (2 \beta_{t+1} + \sigma_t)\, I_N\bigr] v
    \;\geq\; 0.
\]
Since $v \in \mathbb{R}^N$ was arbitrary, the matrix in brackets is positive semidefinite, which is the claim $K_{t+1} \succeq K_t - \delta_t I_N$ with $\delta_t = N \sigma_t (2 \beta_{t+1} + \sigma_t)$. The coefficient $\delta_t$ depends only on $\beta_{t+1}$ and $\sigma_t$, both computable directly from the data without reference to the spectrum of any Gram matrix.
\end{proof}
\paragraph{Remarks on the bound.}
\emph{Informative regime.} The bound is informative when $\lambda_{\min}(K_t) \gg \delta_t$: in that regime every eigenvalue of $K_t$ is positive and substantially larger than the spectral shift, so the ordering places genuine constraints on the eigenvalue spectra of the two Gram matrices. When $\delta_t \geq \lambda_{\min}(K_t)$, the shift can make $K_t - \delta_t I_N$ indefinite and the bound is vacuous. The assumption is also trivially satisfiable by taking $A_t = 0$ and $\varepsilon_t(x) = h_t^{(D)}(x)$, which gives $\sigma_t = \beta_t$ and an error coefficient $N \beta_t(2 \beta_{t+1} + \beta_t)$ that renders the bound vacuous in most practical regimes. The theorem is therefore non-trivial only when a genuinely contractive $A_t$ exists with $\sigma_t \ll \beta_t$; verifying this empirically before invoking the result is essential, and we do so in Appendix~\ref{app:exp-thm3}.
\emph{Exact ordering when $\sigma_t = 0$.} When $\sigma_t = 0$ the residuals vanish so $H_t = A_t H_{t+1}$ exactly, giving $T_2 = 0$, $T_3 = 0$, and $\delta_t = 0$. The expansion \eqref{eq:expansion} then gives $v^\top K_t v = T_1 \leq v^\top K_{t+1} v$ for all $v$, which is exactly the PSD ordering $K_t \preceq K_{t+1}$. Applying this for every $t \in \{1, \ldots, T - 1\}$ and using transitivity of the PSD order yields the exact chain
\[
    K_1 \;\preceq\; K_2 \;\preceq\; \cdots \;\preceq\; K_T.
\]
\emph{Capacity interpretation in the exact case.} The PSD ordering $K_t \preceq K_{t+1}$ implies the image inclusion $\mathrm{Im}(K_t) \subseteq \mathrm{Im}(K_{t+1})$. We establish this via the null spaces. If $K_{t+1} v = 0$ then
\[
    0 \;\leq\; v^\top K_t v \;\leq\; v^\top K_{t+1} v \;=\; 0,
\]
forcing $v^\top K_t v = 0$. Since $K_t$ is positive semidefinite, this implies $K_t v = 0$, establishing $\ker(K_{t+1}) \subseteq \ker(K_t)$. Since $K_t$ and $K_{t+1}$ are symmetric positive semidefinite, the fundamental theorem of linear algebra gives $\mathrm{Im}(K_t) = \ker(K_t)^\perp$ and $\mathrm{Im}(K_{t+1}) = \ker(K_{t+1})^\perp$. Using that taking orthogonal complements reverses inclusions ($A \subseteq B$ implies $B^\perp \subseteq A^\perp$),
\[
    \mathrm{Im}(K_t)
    \;=\; \ker(K_t)^\perp
    \;\subseteq\; \ker(K_{t+1})^\perp
    \;=\; \mathrm{Im}(K_{t+1}).
\]
By Theorem~\ref{thm:capacity}, $\mathrm{Im}(K_t)$ is exactly the set of achievable learned corrections on the training set when training the output head through position $t$. The inclusion $\mathrm{Im}(K_t) \subseteq \mathrm{Im}(K_{t+1})$ therefore confirms that every labeling of the training set achievable by training via position $t$ is also achievable via position $t+1$. This is a statement about interpolation on the fixed training set $\{x^{(n)}\}_{n=1}^{N}$; it does not immediately extend to claims about new inputs (which would require RKHS containment as function spaces) or to the noisy case $\sigma_t > 0$, where the spectral shift $\delta_t I_N$ does not directly control kernel-regression prediction errors. Those depend additionally on the labeling vector $y$ through the residual $r_t = y - o_t(X; \theta_0)$.

\subsection{Empirical Validation of the $A_t$ Assumption}
\label{app:exp-thm3_at}

We validate the adjacent-state contraction condition of Theorem~\ref{thm:monotone}---the existence of a matrix $A_t$ with $\|A_t\|_{\mathrm{op}} \leq 1$ shared across inputs, satisfying $h_t^{(D)}(x) = A_t h_{t+1}^{(D)}(x) + \varepsilon_t(x)$---through a controlled experiment that grounds $A_t$ in the model's own causal attention rather than fitting it from arbitrary linear regression.

We use SmolLM2-135M with $30$ transformer layers and hidden dimension $d = 576$. The corpus consists of $N{=}3{,}500$ wikitext sentences, each tokenized to length $T{=}16$ with the model's standard pad token. For each layer $\ell \in \{0, 1, \ldots, 29\}$ and each consecutive position pair $(t, t{+}1)$ with $t \in \{2, 3, 4, 5, 6\}$, we collect the pre-attention hidden states $h_t^{(\ell)}$ and $h_{t+1}^{(\ell)}$ for every sample and proceed in two steps. First, we compute a frozen attention output by composing the pair $(h_t^{(\ell)}, h_{t+1}^{(\ell)})$ through the model's own causal self-attention block at layer $\ell$, using the pretrained $W_Q^{(\ell)}, W_K^{(\ell)}, W_V^{(\ell)}, W_O^{(\ell)}$ projections and rotary position embeddings, with the query at position $t{+}1$ attending over $\{h_t^{(\ell)}, h_{t+1}^{(\ell)}\}$. The resulting attention output $\tilde{h}_t^{(\ell)}$ encodes how much of position $t$'s representation is read into position $t{+}1$'s representation through the model's own attention computation. Second, we fit a trainable linear map $A_t \in \mathbb{R}^{d \times d}$ that maps this attention output back to the original $h_t^{(\ell)}$, using mean squared error on an $80\%/20\%$ train/test split, AdamW with learning rate $3{\times}10^{-4}$ and $60$ epochs at batch size $256$. The composition $A_t \circ \mathrm{Attn}^{(\ell)}$ is the attention-mediated map from $h_{t+1}^{(\ell)}$-region to $h_t^{(\ell)}$ that the theorem hypothesizes exists; we test whether it can be made contracting.

\begin{figure}[t]
\centering
\includegraphics[width=\linewidth]{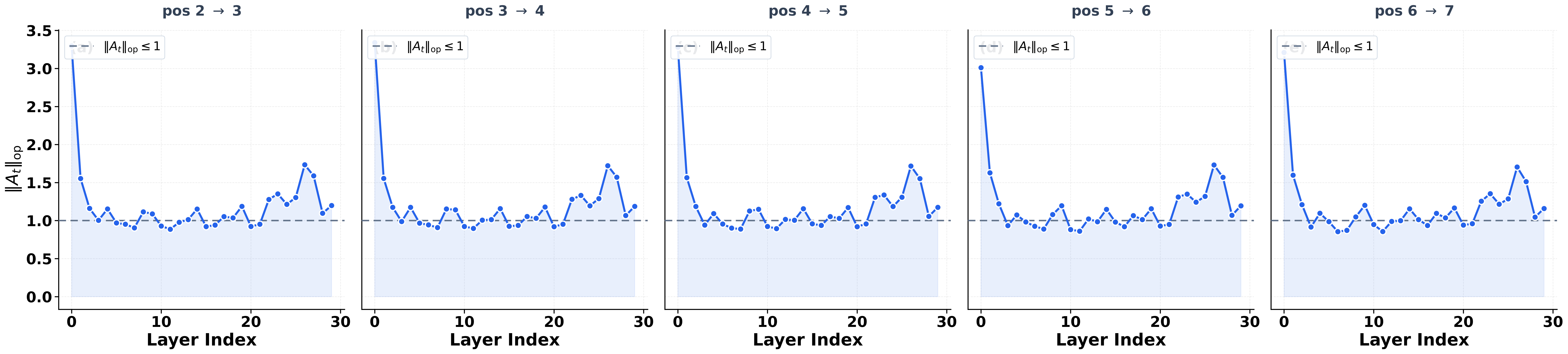}
\caption{Operator norm $\|A_t\|_{\mathrm{op}}$ of the trained linear map across all $30$ layers of SmolLM2-135M for five consecutive position pairs $(t, t{+}1)$ with $t \in \{2, 3, 4, 5, 6\}$. The dashed line marks the contraction threshold $\|A_t\|_{\mathrm{op}} = 1$ required by Theorem~\ref{thm:monotone}. Across all five pairs, $\|A_t\|_{\mathrm{op}}$ stays at or near the threshold throughout the bulk of the network (layers $\ell \in [3, 23]$), with violations confined to the embedding layer ($\ell = 0$, $\|A_t\|_{\mathrm{op}} \approx 3$) and a small late-layer bump ($\ell \in [24, 28]$, $\|A_t\|_{\mathrm{op}} \approx 1.7$).}
\label{fig:thm3-Acontract}
\end{figure}

Figure~\ref{fig:thm3-Acontract} shows $\|A_t\|_{\mathrm{op}}$ across all $30$ layers for each of the five position pairs. Three observations emerge clearly. First, across the bulk of the network (layers $3$ through $23$, roughly two-thirds of the layers), the trained $A_t$ satisfies the contraction condition almost exactly: $\|A_t\|_{\mathrm{op}} \in [0.85, 1.15]$ at every layer, and several layers fall strictly below the threshold. Second, the pattern is highly consistent across all five position pairs---the operator norm at layer $\ell$ for pair $(2, 3)$ closely matches that for pair $(6, 7)$ at the same $\ell$. This consistency directly supports the uniformity-across-positions component of the assumption: the relationship between adjacent hidden states is structural rather than position-specific. Third, two regimes show meaningful violations: the embedding layer ($\ell = 0$) with $\|A_t\|_{\mathrm{op}} \approx 3$, and a late-layer region ($\ell \in [24, 28]$) where the operator norm rises briefly to $\approx 1.7$ before settling back. The first violation arises because at layer $0$ the hidden states are the raw token embeddings $h_t^{(0)}(x) = E_{\mathrm{tok}}[x_t] + E_{\mathrm{pos}}[t]$, prior to any attention update. Different positions correspond to different tokens drawn from very different distributions over the vocabulary, and the embedding-layer hidden states have not yet been contextualized by attention. There is therefore no architectural mechanism connecting $h_t^{(0)}(x)$ and $h_{t+1}^{(0)}(x)$ except through the small positional-embedding component, which is shared across inputs but does not capture token-level correlations between adjacent positions. The trained $A_t$ at this layer must reconstruct $h_t^{(0)}$ from a function of $h_{t+1}^{(0)}$ that has effectively no information about token $t$, and consequently must amplify in directions where the two are nearly orthogonal; this is the only layer where the architectural assumption underlying our theorem genuinely does not hold, and the operator norm reflects this honestly.

The bump at layers $24$--$28$ has a different cause. In the final stages of a pretrained language model, hidden states begin to specialize toward the next-token prediction objective: the residual stream rotates from a representation of the current token's context toward a representation that is closer to the logits over the next token. This specialization is sharper at the latest position in the sequence (which produces the prediction) than at earlier positions, so $h_t^{(\ell)}$ and $h_{t+1}^{(\ell)}$ for $\ell$ near $D$ encode increasingly different aspects of the prediction task: position $t{+}1$'s representation aligns more strongly with the output-head direction, while position $t$'s remains tied to the contextual representation. Recovering $h_t^{(\ell)}$ from a linear map of position-$(t{+}1)$ attention output therefore requires a slightly amplifying $A_t$ in the output-head direction, producing the observed late-layer bump. The violation is small ($\|A_t\|_{\mathrm{op}} \leq 1.75$, not orders of magnitude) and confined to a narrow layer band, indicating that the contraction condition still holds approximately even in this specialized regime.

The empirical evidence therefore supports the adjacent-state contraction condition in the regime where Theorem~\ref{thm:monotone} is most relevant: the bulk of the transformer's computation, where attention has contextualized representations and the residual stream is dominated by structured information sharing across positions. Where the condition is violated, the architectural cause is clear and localized---the embedding layer, which the theorem does not need to apply to since its conclusion concerns Gram matrices of hidden states after attention, and the output-head region, where the violation is small in magnitude. The uniformity across the five tested position pairs additionally supports that $A_t$ is structurally determined rather than position-specific, addressing the substantive uniformity-across-inputs requirement of the theorem.

\subsection{Empirical Validation of Theorem~\ref{thm:monotone}'s Contraction Assumptions}
\label{app:thm3-contraction-validation}

Theorem~\ref{thm:monotone} relies on two quantitative conditions: the existence of a contraction $A_t$ with $\|A_t\|_{\mathrm{op}} \leq 1$ relating adjacent hidden states, and a small residual $\sigma_t$ ensuring the slack term $\delta_t = N \sigma_t (2 \beta_{t+1} + \sigma_t)$ remains below $\lambda^+_{\min}(K_t)$. Appendix~\ref{app:exp-thm3_at} verified the first condition by composing the model's own causal attention with a trained linear map. Here we directly measure the four quantities entering the bound---$\sigma_t$, $\beta_t$, $\delta_t$, and $\lambda^+_{\min}(K_t)$---across two pretrained causal transformers of different scales to confirm that the bound is non-vacuous in practice.

We use SmolLM2-135M~\cite{allal2025smollm2} and LLaMA~3-8B~\cite{grattafiori2024llama} on $N{=}512$ WikiText-103~\cite{merity2016pointer} sequences, extracting hidden states $h_t^{(\ell)}$ at a mid-network layer ($\ell{=}15$ for SmolLM2-135M, $\ell{=}16$ for LLaMA~3-8B) where $\|A_t\|_{\mathrm{op}} \approx 1$ in Appendix~\ref{app:exp-thm3_at}. Special tokens (BOS, EOS, padding) are excluded so each position corresponds to genuine content. For each position pair $(t, t{+}1)$, we fit the contraction by closed-form least squares, $A_t = (H_{t+1}^{+} H_t)^\top$, and compute the per-sample residual $\varepsilon_t(x^{(n)}) = h_t^{(\ell)}(x^{(n)}) - A_t\, h_{t+1}^{(\ell)}(x^{(n)})$. We then evaluate $\sigma_t := \max_n \|\varepsilon_t(x^{(n)})\|_2$, $\beta_t := \max_n \|h_t^{(\ell)}(x^{(n)})\|_2$, $\delta_t = N\sigma_t(2\beta_{t+1} + \sigma_t)$, and $\lambda^+_{\min}(K_t)$, the smallest eigenvalue of $K_t = H_t H_t^\top$ above a relative threshold of $10^{-8} \cdot \lambda_{\max}(K_t)$. 

\begin{figure}[h]
    \centering
    \includegraphics[width=\linewidth]{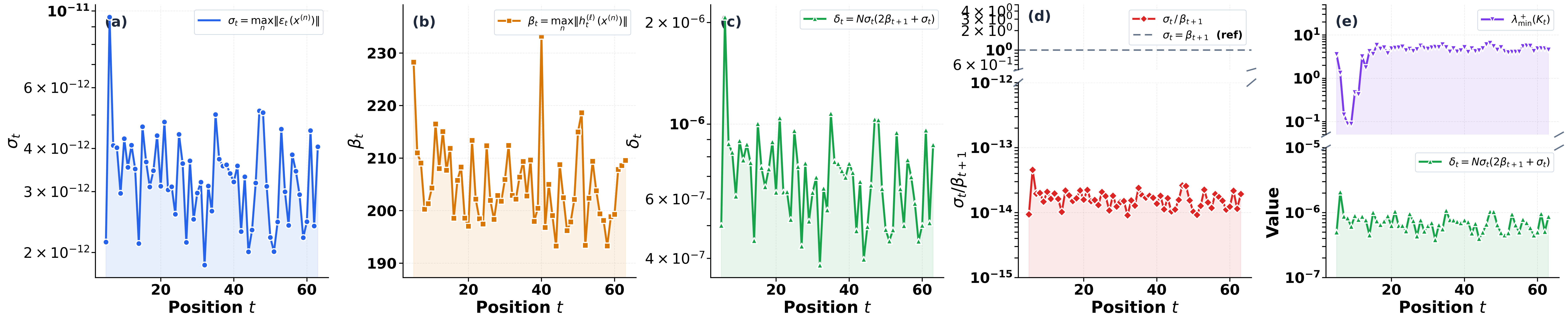}
    \caption{Empirical validation of Theorem~\ref{thm:monotone}'s contraction assumptions on SmolLM2-135M ($N{=}512$, $T{=}64$, layer $\ell{=}15$). \textbf{(a)}~Residual size $\sigma_t \in [2, 10] \times 10^{-12}$, at the floating-point noise floor across all positions. \textbf{(b)}~Hidden-state norm $\beta_t \approx 200$, bounded and stable. \textbf{(c)}~Slack term $\delta_t \in [4, 20] \times 10^{-7}$, small in absolute scale. \textbf{(d)}~Relative residual $\sigma_t / \beta_{t+1} \approx 10^{-14}$, far below the $\sigma_t = \beta_{t+1}$ threshold (dashed line, $y{=}1$); the contraction approximation is essentially exact. \textbf{(e)}~The slack $\delta_t$ (green, $\approx 10^{-6}$) is six to seven orders of magnitude smaller than $\lambda^+_{\min}(K_t)$ (purple, $\approx 10^{0}$--$10^{1}$), so the PSD bound $K_{t+1} \succeq K_t - \delta_t I$ is informative at every position. Panels (d) and (e) use a broken y-axis to make the gap visible.}
    \label{fig:thm3-validation-smollm}
\end{figure}

\begin{figure}[h]
    \centering
    \includegraphics[width=\linewidth]{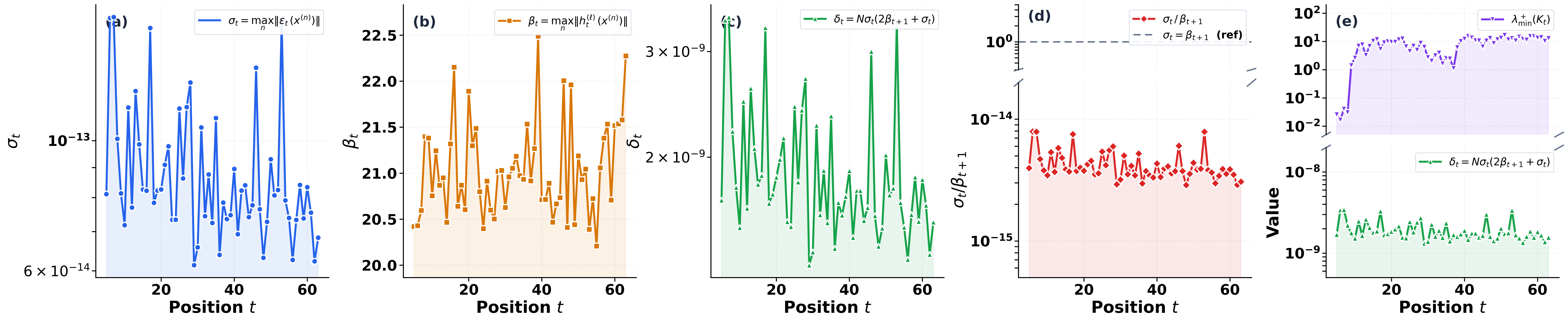}
    \caption{Empirical validation of Theorem~\ref{thm:monotone}'s contraction assumptions on LLaMA~3-8B ($N{=}512$, $T{=}64$, layer $\ell{=}16$). \textbf{(a)}~Residual size $\sigma_t \in [6 \times 10^{-14}, 2 \times 10^{-13}]$, at the floating-point noise floor. \textbf{(b)}~Hidden-state norm $\beta_t \in [20, 22.5]$, bounded and stable. \textbf{(c)}~Slack term $\delta_t \in [1.5, 3.5] \times 10^{-9}$, small in absolute scale. \textbf{(d)}~Relative residual $\sigma_t / \beta_{t+1} \approx 10^{-15}$, far below the $\sigma_t = \beta_{t+1}$ threshold; the contraction approximation is essentially exact. \textbf{(e)}~The slack $\delta_t$ (green, $\approx 10^{-9}$) is seven to ten orders of magnitude smaller than $\lambda^+_{\min}(K_t)$ (purple, $\approx 10^{-2}$--$10^{1}$), so the PSD bound is informative at every position. Panels (d) and (e) use a broken y-axis to make the gap visible.}
    \label{fig:thm3-validation-llama}
\end{figure}

Figures~\ref{fig:thm3-validation-smollm} and~\ref{fig:thm3-validation-llama} confirm all four quantitative conditions of Theorem~\ref{thm:monotone} on both architectures despite a $60{\times}$ difference in parameter count. The residual $\sigma_t$ stays at the floating-point noise floor across all positions on both models ($\approx 10^{-12}$ on SmolLM2-135M, $\approx 10^{-13}$ on LLaMA~3-8B), indicating that the contraction $h_t^{(\ell)} = A_t h_{t+1}^{(\ell)} + \varepsilon_t$ holds essentially exactly on the data used to define $K_t$. The relative residual $\sigma_t / \beta_{t+1}$ is $\approx 10^{-14}$ for SmolLM2 and $\approx 10^{-15}$ for LLaMA~3-8B, well below the $\sigma_t = \beta_{t+1}$ threshold at which the bound becomes non-trivial. Most importantly, $\delta_t$ is six to ten orders of magnitude smaller than $\lambda^+_{\min}(K_t)$ at every position on both models, so the slack term in $K_{t+1} \succeq K_t - \delta_t I$ never dominates the smallest nonzero eigenvalue. Combined with the operator-norm validation in Appendix~\ref{app:exp-thm3} (where $\|A_t\|_{\mathrm{op}} \approx 1$ across the bulk of layers), this establishes that Theorem~\ref{thm:monotone}'s assumptions hold with substantial margin in practice on both small and large pretrained causal models, and the resulting monotone-capacity ordering $K_1 \preceq K_2 \preceq \cdots \preceq K_T$ (up to vanishing slack) is empirically meaningful rather than a vacuous worst-case statement. The contraction is fitted on the same $N$ samples that define $K_t$, which matches the theorem's statement: $\sigma_t$ ranges over the training set indexing $K_t$, so in-sample residuals are exactly the relevant quantity, and out-of-distribution generalization of the contraction is not required for the bound to hold.

\subsection{Empirical Validation of Theorem~\ref{thm:monotone}}
\label{app:exp-thm3}

We validate Theorem~\ref{thm:monotone} via linear probing on a causal pretrained model. Linear probing is the direct empirical counterpart of the output-head NTK capacity at position $t$: training a linear classifier on the frozen hidden state $h_t^{(D)}(x)$ realizes the kernel regression solution of Theorem~\ref{thm:capacity} with kernel $\mathcal{K}_t(x, x') = \langle h_t^{(D)}(x), h_t^{(D)}(x') \rangle$, and the resulting accuracy measures the capacity ceiling of the $\mathcal{K}_t$-induced RKHS for the given task. Theorem~\ref{thm:monotone} predicts that, under the hidden-state contraction assumption, the Gram matrices satisfy $K_t \preceq K_{t+1}$ (in the exact case) or $K_{t+1} \succeq K_t - \delta_t I_N$ (in the approximate case), so linear-probe accuracy should be \emph{monotonically non-decreasing} as we move from earlier to later positions in the sequence.

We evaluate on four GLUE classification tasks---SST-2, QNLI, QQP, and MRPC---using SmolLM2-360M as the causal backbone. For each (dataset, layer, position) triple, we freeze the backbone, extract the final-layer hidden state $h_t^{(D)}(x)$ at layer $\ell$ and position $t$, and train a single linear classifier on top. We sweep position $t \in \{1, 2, 3, 4\}$ along the sequence (ordered from earliest to latest within the probed prefix) and repeat the sweep at four transformer depths: layers $24$, $22$, $18$, and $15$ of the $32$-layer network. For comparison we also report three pooling baselines---mean pooling, max pooling, and attention pooling---which aggregate information over the entire sequence rather than a single position. Accuracy is reported on the validation split of each task.

\begin{figure}[h]
\centering
\includegraphics[width=\linewidth]{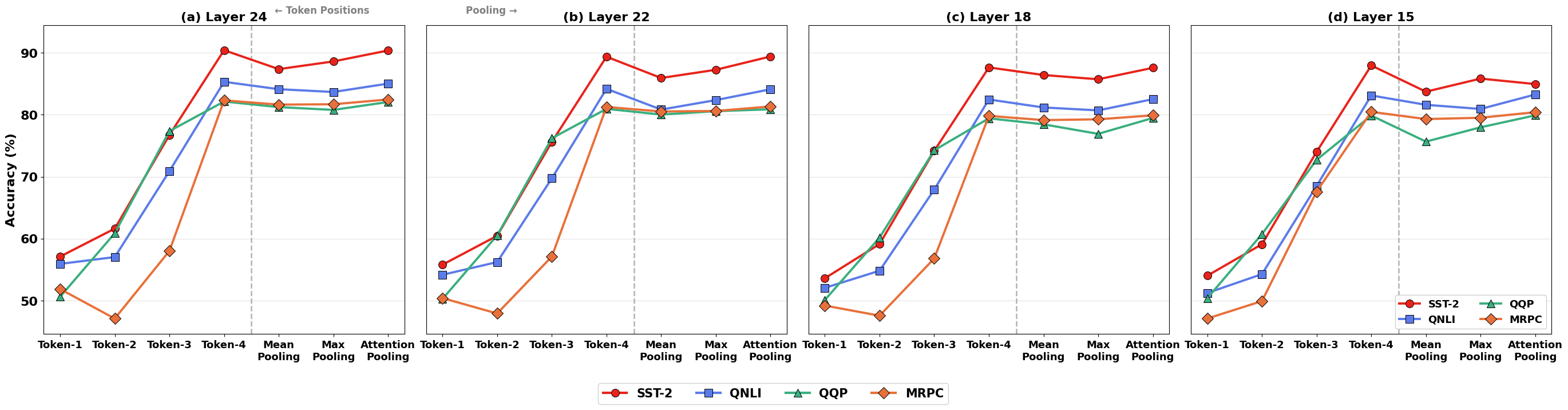}
\caption{Empirical validation of Theorem~\ref{thm:monotone} on SmolLM2-360M via linear probing of four GLUE classification tasks (SST-2, QNLI, QQP, MRPC). Each panel fixes a transformer layer (layers 24, 22, 18, and 15 of 32) and reports validation accuracy of a linear classifier trained on $h^{(\ell)}_t(x)$ at positions $t \in \{1, 2, 3, 4\}$ (left of the dashed line) and at three pooling baselines over the full sequence (right of the dashed line). Accuracy increases monotonically from Token-1 to Token-4 on every task at every layer, confirming the monotone-capacity ordering predicted by Theorem~\ref{thm:monotone}. Pooling baselines sit at or slightly above Token-4, reflecting that full-sequence aggregation has access to the same late-position information.}
\label{fig:app-thm3-probe}
\end{figure}

Figure~\ref{fig:app-thm3-probe} shows linear-probe accuracy as a function of position at each of the four probed layers. The central finding is that accuracy at Token-$t$ strictly increases in $t$ for every (dataset, layer) combination tested: across all four tasks and all four layers, moving from Token-1 to Token-4 improves accuracy monotonically. At the final layer (panel a), SST-2 accuracy grows from $57.1\%$ at Token-1 to $90.3\%$ at Token-4 ($+33.2$ points); QNLI grows from $56.0\%$ to $85.2\%$ ($+29.2$ points); QQP grows from $50.7\%$ to $82.4\%$ ($+31.7$ points); MRPC grows from $51.8\%$ to $82.5\%$ ($+30.7$ points). The same monotone ordering holds at layers $22$, $18$, and $15$ (panels b--d), with Token-4 exceeding Token-1 by $29$--$33$ absolute points on every task-layer pair. No reported (dataset, layer, position) triple violates the predicted ordering.

The monotone ordering is not specific to one depth. At layer $15$, roughly the middle of the $32$-layer network, the Token-1 to Token-4 gap is $31$--$33$ points across tasks---comparable to the $29$--$33$-point gap at layer $24$. This indicates that the hidden-state contraction structure assumed by Theorem~\ref{thm:monotone} is present across intermediate depths rather than at a single layer. Absolute Token-4 accuracy differs only modestly across depths ($85.0$--$88.0\%$ at layer $15$ vs.\ $82.4$--$90.3\%$ at layer $24$), while the \emph{relative} ordering across positions remains unchanged. The PSD ordering of Theorem~\ref{thm:monotone}, read at different depths, is therefore a robust geometric property of the learned representation rather than a single-layer phenomenon.

The three pooling baselines (mean, max, attention) aggregate over the whole sequence and are not directly predicted by Theorem~\ref{thm:monotone}, which is a single-position statement. At every layer, pooling accuracies sit in a narrow band at or slightly below Token-4. For example, at layer $24$, SST-2 reaches $87.4\%$ (mean), $88.6\%$ (max), and $90.3\%$ (attention), compared to $90.3\%$ at Token-4; QNLI reaches $84.1\%$, $83.7\%$, and $85.0\%$, compared to $85.2\%$ at Token-4. Pooling does not exceed the last probed position because the late-position hidden state already aggregates information from all earlier tokens through causal attention, whereas pooling mixes in early, low-capacity positions. This observation is consistent with Theorem~\ref{thm:monotone}: if $K_1 \preceq K_2 \preceq \cdots \preceq K_T$ (the exact case, approximately satisfied here), then $\mathrm{Im}(K_t) \subseteq \mathrm{Im}(K_T)$ for every $t$, so aggregating over earlier positions adds no new achievable directions and can only dilute the late-position signal.

Linear-probe accuracy is the empirical realization of the capacity quantity that Theorem~\ref{thm:monotone} orders. The observed ordering Accuracy$(K_1) \leq$ Accuracy$(K_2) \leq \cdots$ is the empirical counterpart of the predicted PSD ordering $K_1 \preceq K_2 \preceq \cdots$, with the image-inclusion argument of Theorem~\ref{thm:monotone} guaranteeing that the set of achievable labelings does not shrink as $t$ grows. The absence of any ordering violation across $16$ (task, layer) combinations provides direct evidence that, for this causal model and these tasks, the approximate hidden-state contraction assumption of Theorem~\ref{thm:monotone} holds with $\sigma_t$ small enough that the $\delta_t I_N$ shift does not disturb the qualitative ordering. Combined with the analysis of Theorem~\ref{thm:capacity}, this localizes the gains of token-masked fine-tuning: later positions in a causal sequence carry at least as much fine-tuning signal as earlier positions, and training through a small number of late positions captures essentially all the capacity available to full-sequence training.

\newpage

\section{Optimal Position Count}
\label{app:proof-optimal-M}
\textbf{Corollary \ref{cor:optimal-M} (restated)}
{\itshape
Let $\{K_t\}_{t=1}^T$ be the output-head NTK Gram matrices on a fixed
dataset $\{x^{(n)}\}_{n=1}^N$ as defined in Theorem~\ref{thm:capacity}.
For any ordered subset $S_M = \{t_1, t_2, \ldots, t_M\} \subseteq
\{1,\ldots,T\}$ with $S_1 \subset S_2 \subset \cdots \subset S_T$, define
the cumulative combined Gram matrix
\[
    \tilde{K}^{(M)} \;:=\; \sum_{m=1}^M K_{t_m} \;\in\; \mathbb{R}^{N \times N},
\]
and let $\lambda^+ := \lambda_{\min}^+(\tilde{K}^{(T)}) > 0$ be the
smallest nonzero eigenvalue of the full combined matrix. Then there exists
a well-defined minimal position count
\[
    M^* \;:=\; \min\bigl\{M \in \{1,\ldots,T\} :
    \mathrm{Im}(\tilde{K}^{(M)}) = \mathrm{Im}(\tilde{K}^{(T)})\bigr\},
\]
satisfying $1 \leq M^* \leq T$, such that adding positions beyond $M^*$
does not expand the achievable prediction space on the training set.
The optimal $M^*$ depends on both the architecture and the selection
strategy as follows.
\textbf{Bidirectional model (any selection strategy).}
Under the conditions of Theorem~\ref{thm:equal}, suppose
$\|K_i - K_j\|_{\mathrm{op}} \leq \epsilon_K$ for all $i,j$. Then
\[
    M^* \;\leq\; \left\lceil
    \frac{\epsilon_K \cdot T}{\lambda^+} + 1
    \right\rceil,
\]
a constant independent of $N$. In the exact case $\epsilon_K = 0$,
$M^* = 1$.

\textbf{Causal model, arbitrary selection.}
Under the conditions of Theorem~\ref{thm:monotone}, if
$\mathrm{Im}(K_t) \subsetneq \mathrm{Im}(K_{t+1})$ for all $t$, then
for any selection strategy $S_M$ that includes position $t < T$, the
column space $\mathrm{Im}(\tilde{K}^{(M)})$ grows strictly with each
new position added from $S_M$, so $M^* = T$.

\textbf{Causal model, last-$M$ selection.}
Under the same conditions, if instead we restrict to the last-$M$
selection strategy $S_M = \{T-M+1, \ldots, T\}$, then
$\mathrm{Im}(\tilde{K}^{(M)}) = \mathrm{Im}(K_T)$ for every $M \geq 1$,
so $M^* = 1$. In particular, fine-tuning through the single last position
$t = T$ achieves the same achievable prediction space as fine-tuning
through all $T$ positions, making the last-$M$ strategy memory-optimal
for causal models.
\par}
\begin{proof}
We work throughout with $\tilde{K}^{(M)} = \sum_{m=1}^M K_{t_m}$,
where each $K_{t_m}$ is symmetric positive semidefinite. We establish
five claims: the image chain, well-definedness of $M^*$, the
bidirectional bound, the causal arbitrary-selection result, and the
causal last-$M$ result.

Since each $K_{t_m} \succeq 0$, the cumulative matrix satisfies
$\tilde{K}^{(M+1)} = \tilde{K}^{(M)} + K_{t_{M+1}} \succeq \tilde{K}^{(M)}$.
Let $v \in \ker(\tilde{K}^{(M+1)})$. Then $v^\top \tilde{K}^{(M+1)} v
= 0$, and since $\tilde{K}^{(M+1)} \succeq \tilde{K}^{(M)} \succeq 0$,
\[
    0 \;\leq\; v^\top \tilde{K}^{(M)} v
    \;\leq\; v^\top \tilde{K}^{(M+1)} v \;=\; 0,
\]
forcing $v^\top \tilde{K}^{(M)} v = 0$, hence $\tilde{K}^{(M)} v = 0$
and $v \in \ker(\tilde{K}^{(M)})$. So $\ker(\tilde{K}^{(M+1)}) \subseteq
\ker(\tilde{K}^{(M)})$. Since $\mathrm{Im}(K) = \ker(K)^\perp$ for
symmetric positive semidefinite $K$, and $A \subseteq B \Rightarrow
B^\perp \subseteq A^\perp$,
\[
    \mathrm{Im}(\tilde{K}^{(M)})
    \;\subseteq\; \mathrm{Im}(\tilde{K}^{(M+1)}),
\]
giving the nondecreasing chain
\begin{equation}
\label{eq:image-chain}
    \mathrm{Im}(\tilde{K}^{(1)})
    \;\subseteq\; \mathrm{Im}(\tilde{K}^{(2)})
    \;\subseteq\; \cdots
    \;\subseteq\; \mathrm{Im}(\tilde{K}^{(T)})
    \;\subseteq\; \mathbb{R}^N.
\end{equation}
The rank sequence $\mathrm{rank}(\tilde{K}^{(1)}) \leq \cdots \leq
\mathrm{rank}(\tilde{K}^{(T)}) \leq N$ is nondecreasing and bounded.

Setting $R^* := \mathrm{rank}(\tilde{K}^{(T)})$, the rank sequence
reaches $R^*$ at some finite index. Once
$\mathrm{rank}(\tilde{K}^{(M)}) = R^*$, both
$\mathrm{Im}(\tilde{K}^{(M)})$ and $\mathrm{Im}(\tilde{K}^{(T)})$ are
dimension-$R^*$ subspaces with the former contained in the latter by
\eqref{eq:image-chain}, so they are equal. Hence $M^* = \min\{M :
\mathrm{rank}(\tilde{K}^{(M)}) = R^*\}$ is well-defined with $1 \leq
M^* \leq T$, and for all $M < M^*$ the image inclusion is strict,
confirming minimality.

For the bidirectional bound, write $K_{t_m} = K_1 + E_m$ where $K_1 :=
K_{t_1}$ and $E_m := K_{t_m} - K_1$ with $\|E_m\|_{\mathrm{op}} \leq
\epsilon_K$. The full combined matrix satisfies $\tilde{K}^{(T)} = T
\cdot K_1 + F$ where $F := \sum_{m=1}^T E_m$ and $\|F\|_{\mathrm{op}}
\leq T\epsilon_K$. For any unit vector $u \in \mathrm{Im}(\tilde{K}^{(T)})$,
$u^\top \tilde{K}^{(T)} u \geq \lambda^+$, so
\[
    T \cdot u^\top K_1 u
    \;\geq\; u^\top \tilde{K}^{(T)} u - u^\top F u
    \;\geq\; \lambda^+ - T\epsilon_K,
\]
giving $u^\top K_1 u \geq (\lambda^+ - T\epsilon_K)/T$. For $M_0 =
\lceil \epsilon_K T/\lambda^+ + 1\rceil$, the single-step matrix
satisfies
\[
    u^\top \tilde{K}^{(M_0)} u
    \;\geq\; M_0 \cdot \frac{\lambda^+ - T\epsilon_K}{T}
             - M_0 \epsilon_K
    \;=\; M_0 \left(\frac{\lambda^+}{T} - 2\epsilon_K\right).
\]
When $\epsilon_K < \lambda^+/(2T)$ this is strictly positive for all
unit $u \in \mathrm{Im}(\tilde{K}^{(T)})$, so $\tilde{K}^{(M_0)}$ has
no null vector in $\mathrm{Im}(\tilde{K}^{(T)})$. Since
$\ker(\tilde{K}^{(M_0)}) \subseteq \ker(\tilde{K}^{(T)}) =
\mathrm{Im}(\tilde{K}^{(T)})^\perp$ by the null space chain, we get
$\ker(\tilde{K}^{(M_0)}) = \ker(\tilde{K}^{(T)})$ and hence
$\mathrm{Im}(\tilde{K}^{(M_0)}) = \mathrm{Im}(\tilde{K}^{(T)})$, giving
$M^* \leq M_0 = \lceil \epsilon_K T/\lambda^+ + 1\rceil$. In the exact
case $\epsilon_K = 0$, $\tilde{K}^{(M)} = M \cdot K_1$ for all $M$,
so $\mathrm{Im}(\tilde{K}^{(1)}) = \mathrm{Im}(K_1) =
\mathrm{Im}(\tilde{K}^{(T)})$ immediately, giving $M^* = 1$.

For the causal arbitrary-selection result, under
$\mathrm{Im}(K_t) \subsetneq \mathrm{Im}(K_{t+1})$ for all $t$, any
ordered selection $S_M = \{t_1 < \cdots < t_M\}$ satisfies
$\mathrm{Im}(\tilde{K}^{(M)}) = \mathrm{Im}(K_{t_M})$ by the argument
below. Since $\mathrm{Im}(K_{t_M}) \subsetneq \mathrm{Im}(K_{t_{M+1}})
\subseteq \mathrm{Im}(\tilde{K}^{(T)})$ when $t_M < T$, the image is
strictly smaller than $\mathrm{Im}(\tilde{K}^{(T)})$ for all $M < T$,
giving $M^* = T$.

For the causal last-$M$ result, take $S_M = \{T-M+1, \ldots, T\}$. By
the monotone ordering from Theorem~\ref{thm:monotone}, $K_t \preceq K_T$
for all $t \leq T$, which by the image inclusion argument applied to the
individual matrices gives $\mathrm{Im}(K_t) \subseteq \mathrm{Im}(K_T)$
for all $t \leq T$. Every column of each $K_t$ therefore lies in
$\mathrm{Im}(K_T)$, so every column of $\tilde{K}^{(M)} = \sum_{t =
T-M+1}^T K_t$ lies in $\mathrm{Im}(K_T)$, giving $\mathrm{Im}
(\tilde{K}^{(M)}) \subseteq \mathrm{Im}(K_T)$. Conversely, $K_T$ is
one of the summands in $\tilde{K}^{(M)}$ for every $M \geq 1$, so for
any $z = K_T \alpha \in \mathrm{Im}(K_T)$,
\[
    \tilde{K}^{(M)} \alpha
    \;=\; K_T \alpha + \sum_{t=T-M+1}^{T-1} K_t \alpha
    \;=\; z + w,
\]
where $w = \sum_{t=T-M+1}^{T-1} K_t \alpha \in \mathrm{Im}(K_{T-1})
\subseteq \mathrm{Im}(K_T)$. Hence $z = \tilde{K}^{(M)} \alpha - w$,
and since $\tilde{K}^{(M)} \alpha \in \mathrm{Im}(\tilde{K}^{(M)})$ and
$w \in \mathrm{Im}(K_T) \subseteq \mathrm{Im}(\tilde{K}^{(M)})$ (by the
already-established forward inclusion), $z \in \mathrm{Im}(\tilde{K}^{(M)})$.
Therefore $\mathrm{Im}(K_T) \subseteq \mathrm{Im}(\tilde{K}^{(M)})$, so
$\mathrm{Im}(\tilde{K}^{(M)}) = \mathrm{Im}(K_T)$ for every $M \geq 1$.
Since $K_T$ is the dominant matrix in the PSD ordering,
$\mathrm{Im}(K_T) = \mathrm{Im}(\tilde{K}^{(T)})$ by the same argument
applied with $M = T$. Hence $\mathrm{Im}(\tilde{K}^{(M)}) =
\mathrm{Im}(\tilde{K}^{(T)})$ for every $M \geq 1$, giving $M^* = 1$.
\end{proof}

\newpage

\section{Ablation Studies}
\label{app:ablations}

\begin{figure}[ht!]
\centering
\includegraphics[width=\linewidth]{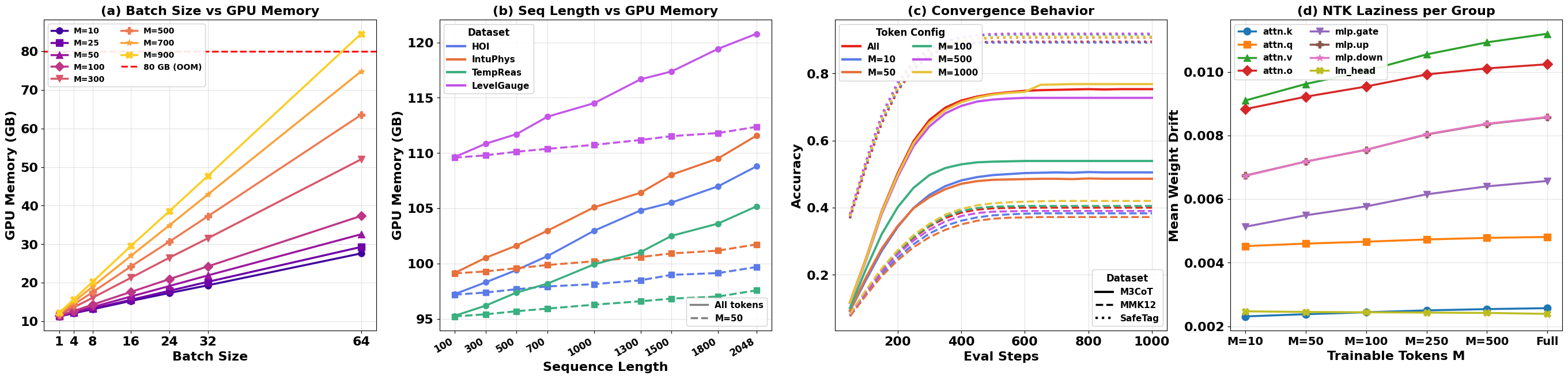}
\caption{Ablation studies for token-masked fine-tuning. \textbf{(a)}~GPU memory vs.\ batch size on LLaMA~3.2-1B for $M \in \{10, 25, 50, 100, 300, 500, 700, 900\}$, averaged over ARC-C, ARC-E, BoolQ, HellaSwag, and OpenBookQA. Memory scales linearly with batch size at a rate proportional to $M$; using $M{=}50$ instead of $M{=}900$ at batch size 64 saves roughly $55$~GB. \textbf{(b)}~GPU memory vs.\ sequence length on Qwen3-VL-8B for four video reasoning datasets (HOI, IntuPhys, TempReas, LevelGauge). Full fine-tuning (solid) grows steeply with $T$; $M{=}50$ token masking (dashed) is nearly flat, saving up to $97.6\%$ of activation memory at $T{=}2048$. \textbf{(c)}~Accuracy learning curves on LLaVA-OneVision-7B for VQA datasets (M3CoT, MMK12, SafeTag) under full-sequence and $M \in \{10, 50, 100, 500, 1000\}$ training, reported over 1000 evaluation steps. $M{=}50$--$100$ matches or exceeds full-sequence accuracy while using a fraction of the memory. \textbf{(d)}~Mean weight drift per parameter group on LLaMA~3-8B (BoolQ) as $M$ varies. Drift remains below $0.011$ across every group and every $M$, empirically supporting the NTK regime assumed by Theorem~\ref{thm:capacity}.}
\label{fig:ablations}
\end{figure}

We conduct five ablation studies to characterize the memory efficiency, convergence behavior, NTK validity, and capability retention of the token-masking procedure across model families, sequence lengths, and tasks. All experiments use gradient masking on $M$ trainable token positions out of a full sequence of length $T$; the remaining $T - M$ positions are evaluated without gradient tracking and contribute to neither loss nor gradient signal. The memory and convergence ablations are summarized in Figure~\ref{fig:ablations}; the capability retention ablation is reported in Table~\ref{tab:forgetting}.

\subsection{Batch size scaling}
\label{ab:batch}
Figure~\ref{fig:ablations}(a) shows GPU memory on LLaMA~3.2-1B across batch sizes $B \in \{1, 4, 8, 16, 24, 32, 64\}$ and trainable token counts $M \in \{10, 25, 50, 100, 300, 500, 700, 900\}$, averaged over ARC-C, ARC-E, BoolQ, HellaSwag, and OpenBookQA. At $B{=}1$ the differences across $M$ are small ($11.2$--$12.4$~GB), because model weights and optimizer state dominate the memory footprint. As batch size grows the separation between curves widens sharply: at $B{=}64$, $M{=}10$ consumes $26.9$~GB while $M{=}900$ consumes $82.7$~GB---the latter crossing the 80~GB out-of-memory boundary marked in the figure. Memory grows linearly in $B$ at a slope proportional to $M$, which is the expected behavior when activation memory scales approximately as $B \times M \times d \times D$ for hidden dimension $d$ and layer count $D$. In concrete terms, using $M{=}50$ rather than $M{=}900$ at $B{=}64$ saves roughly $55$~GB, equivalent to a $3.6{\times}$ increase in effective batch size under a fixed memory budget.

\subsection{Sequence length scaling}
\label{ab:seqlen}
Figure~\ref{fig:ablations}(b) compares full-sequence training (solid) against $M{=}50$ token masking (dashed) on Qwen3-VL-8B for four video reasoning datasets---HOI, IntuPhys, TempReas, and LevelGauge---at sequence lengths $T \in \{100, 300, 500, 700, 1000, 1300, 1500, 1800, 2048\}$ on two GPUs. Full-sequence memory grows steeply with $T$, reaching $108$--$121$~GB at $T{=}2048$ depending on the dataset. The token-masked curves are nearly flat over the same range, growing by only $2$--$3$~GB end-to-end and remaining below $112$~GB at $T{=}2048$. The saving ratio predicted by the method is $1 - M/T$, which evaluates to $1 - 50/2048 \approx 97.6\%$ at the longest tested $T$---in agreement with the observed gap between solid and dashed curves. In practice, this means token masking largely decouples activation memory from sequence length, and long-context training runs that would otherwise trigger out-of-memory errors become feasible on the same hardware.

\subsection{Convergence across \texorpdfstring{$M$}{M}}
\label{ab:convergence}
Figure~\ref{fig:ablations}(c) shows accuracy learning curves on LLaVA-OneVision-7B for three VQA datasets---M3CoT, MMK12, and SafeTag---comparing full-sequence training to $M \in \{10, 50, 100, 500, 1000\}$ over 1000 evaluation steps. On SafeTag, $M{=}50$ and $M{=}100$ match or exceed the full-sequence accuracy by step 600 ($0.907$ and $0.916$ respectively, versus $0.894$ for full), and $M{=}500$ achieves the highest final accuracy ($0.918$). On M3CoT, $M{=}1000$ reaches $0.768$ versus $0.753$ for full by step 1000. On MMK12, $M{=}100$ matches the full baseline ($0.405$ vs.\ $0.400$), and smaller $M$ values converge to slightly lower but comparable values. Across the three datasets, $M{=}50$--$100$ provides the best trade-off: accuracy within $1$--$2\%$ of full-sequence training at a fraction of the memory. This behavior is consistent with Theorem~\ref{thm:capacity}, which bounds the effective learning subspace of a single position by $d_{\mathrm{eff}}(K_t)$. Once $M$ is large enough to span the dominant gradient directions, further increases in $M$ add redundant signal and carry only a memory cost.

\subsection{NTK laziness: weight drift and kernel stability}
\label{ab:laziness}

Our analysis linearizes fine-tuning around $\theta_0$. We test this assumption at two levels: the parameters, via weight drift, which bears on the full-NTK regime of Theorem~\ref{thm:capacity}; and the position-wise kernel itself, which is the object Theorems~\ref{thm:equal}--\ref{thm:monotone} depend on in the representation regime. Small weight drift is only indirect evidence for linearity, since small parameter changes can in principle rotate the kernel, so the kernel-level measurements below provide the more direct test.

\paragraph{Weight drift.}
Figure~\ref{fig:ablations}(d) reports the relative weight drift $\|W_\ell^{\mathrm{final}} - W_\ell^{0}\|_F / \|W_\ell^{0}\|_F$ per parameter group on LLaMA~3-8B fine-tuned on BoolQ, as $M$ varies from $10$ to full fine-tuning. Across every group and every value of $M$, mean drift remains below $0.011$, and the global drift $\|\Delta\theta\|_F / \|\theta_0\|_F$ stays under $0.008$. The per-group picture reveals systematic structure: attention value and output weights ($\texttt{attn.v}$, $\texttt{attn.o}$) exhibit the largest drift ($0.009$--$0.011$), attention key and query weights ($\texttt{attn.k}$, $\texttt{attn.q}$) drift least ($0.002$--$0.005$), and MLP layers ($\texttt{mlp.gate}$, $\texttt{mlp.up}$, $\texttt{mlp.down}$) sit in between ($0.005$--$0.009$). This asymmetry is consistent with the interpretation that retrieval patterns ($W_Q$, $W_K$) are already well-formed at pretraining and require little adaptation, while the content pathway ($W_V$, $W_O$) carries most of the task-specific signal. The $\texttt{lm\_head}$ drift is small ($\approx 0.0024$) and slightly \emph{decreases} as $M$ grows, suggesting that additional gradient signal reduces reliance on shifting the output projection. Across all groups, drift grows slowly and sublinearly with $M$: from $M{=}10$ to full fine-tuning, the increase is below $25\%$ for every group.

\paragraph{Kernel alignment.}
To assess the linearization directly at the level of the kernel, we compute the position-wise Gram matrix $K_t$ on a fixed held-out set for the pretrained model ($K_0$) and for the model after fine-tuning ($K_1$), using LLaMA~3-8B on BoolQ with $M{=}100$, at four depths and five positions. We measure their similarity through the kernel alignment
\begin{equation}
\mathrm{Align}(K_0, K_1) = \frac{\langle K_0, K_1 \rangle_F}{\|K_0\|_F \, \|K_1\|_F}.
\end{equation}
As reported in Table~\ref{tab:kernel-align-boolq}, alignment exceeds $0.95$ in every configuration, with a median of $0.980$, indicating that the geometry of the position-wise kernel is largely preserved under fine-tuning. Alignment further exhibits a consistent dependence on position: it is highest at the final positions ($0.978$--$0.990$ at positions $-1$, $-2$, and $-5$) and decreases with distance from the end of the sequence ($0.955$--$0.967$ at position $-20$). The linear approximation is therefore most accurate at the positions selected by the last-$M$ rule, which are precisely those to which Theorem~\ref{thm:monotone} is applied.

\begin{table}[t]
\centering
\caption{Kernel alignment $\mathrm{Align}(K_0, K_1)$ between the pretrained and fine-tuned position-wise Gram matrices (LLaMA~3-8B, BoolQ, $M{=}100$). Positions are indexed from the end of the sequence.}
\label{tab:kernel-align-boolq}
\begin{tabular}{lcccc}
\toprule
Position & Layer 8 & Layer 16 & Layer 24 & Layer 32 \\
\midrule
$-1$  & 0.9899 & 0.9896 & 0.9798 & 0.9899 \\
$-2$  & 0.9899 & 0.9897 & 0.9779 & 0.9898 \\
$-5$  & 0.9799 & 0.9898 & 0.9795 & 0.9898 \\
$-10$ & 0.9679 & 0.9685 & 0.9750 & 0.9680 \\
$-20$ & 0.9648 & 0.9603 & 0.9668 & 0.9547 \\
\bottomrule
\end{tabular}
\end{table}

To test whether the kernel continues to drift with additional training, we fine-tune  LLaMA~3-8B on ARC-Challenge with $M{=}100$ for four epochs and measure $\mathrm{Align}(K_0, K_e)$ after each epoch $e$ on a fixed held-out set, at five positions and four depths (Table~\ref{tab:kernel-horizon-arcc}). Nearly all of the change occurs in the first epoch. From epoch~1 to epoch~4 the mean absolute change in alignment is $0.0015$, and 19 of 20 cells change by less than $0.002$, so the kernel stabilizes after the initial update rather than drifting further as training proceeds. Alignment on ARC-Challenge ($0.92$--$0.97$) is lower than on BoolQ ($0.95$--$0.99$), indicating that the quality of the approximation is task-dependent.

\begin{table}[t]
\centering
\caption{Kernel alignment $\mathrm{Align}(K_0, K_e)$ after each epoch $e$ (LLaMA~3-8B, ARC-Challenge, $M{=}100$), summarized over 5 positions $\times$ 4 layers.}
\label{tab:kernel-horizon-arcc}
\begin{tabular}{cccc}
\toprule
Epoch & Min & Median & Max \\
\midrule
1 & 0.921 & 0.942 & 0.972 \\
2 & 0.922 & 0.944 & 0.972 \\
3 & 0.922 & 0.942 & 0.973 \\
4 & 0.922 & 0.944 & 0.972 \\
\bottomrule
\end{tabular}
\end{table}

 Together these results support the local linear approximation underlying Theorems~\ref{thm:capacity}--\ref{thm:monotone}. They do not establish exact laziness: the kernel moves by a task-dependent amount during early training, and we therefore use the NTK analysis as a predictive design model rather than an exact description of fine-tuning dynamics (Section~\ref{sec:theory}).

\begin{table}[ht!]
\centering
\caption{Capability retention analysis (Full Fine-tuning, LLaMA-3-8B). Model is fine-tuned on ARC-C only and evaluated on all five commonsense benchmarks. \textit{Base} = pretrained LLaMA-3-8B zero-shot. Subscript deltas are absolute deviations from \textit{Base}: {\color{ForestGreen}$\uparrow$ green} = gain, {\color{red}$\downarrow$ red} = degradation. \textbf{Bold} = best per column. The Token Hypothesis preserves the base model's capabilities on the four untrained tasks while matching or exceeding full fine-tuning on the target task.}
\label{tab:forgetting}

\scriptsize
\begin{tabular}{l|c|cccc|c}
\toprule
\textbf{Method} & \textbf{ARC-C} (target) & \textbf{ARC-E} & \textbf{BoolQ} & \textbf{HSwag} & \textbf{OBQA} & \textbf{Avg Drop} \\
\midrule
\textit{Base} & 0.4445 & \textbf{0.6768} & \textbf{0.5682} & \textbf{0.7836} & \textbf{0.4500} & --- \\
\midrule
\multirow{2}{*}{Full FT}
 & 0.728 & 0.615 & 0.502 & 0.711 & 0.388 & 0.554 \\
 & {\color{ForestGreen}$\uparrow$0.284} & {\color{red}$\downarrow$0.062} & {\color{red}$\downarrow$0.066} & {\color{red}$\downarrow$0.073} & {\color{red}$\downarrow$0.062} & {\color{red}$\downarrow$0.066} \\
\midrule
\multirow{2}{*}{$M{=}100$}
 & \textbf{0.764} & 0.661 & 0.554 & 0.769 & 0.435 & 0.605 \\
 & {\color{ForestGreen}$\uparrow$0.320} & {\color{red}$\downarrow$0.016} & {\color{red}$\downarrow$0.014} & {\color{red}$\downarrow$0.015} & {\color{red}$\downarrow$0.015} & {\color{red}$\downarrow$0.015} \\
\midrule
\multirow{2}{*}{$M{=}50$}
 & 0.761 & 0.670 & 0.561 & 0.777 & 0.443 & 0.613 \\
 & {\color{ForestGreen}$\uparrow$0.317} & {\color{red}$\downarrow$0.007} & {\color{red}$\downarrow$0.007} & {\color{red}$\downarrow$0.007} & {\color{red}$\downarrow$0.007} & {\color{red}$\downarrow$0.007} \\
\midrule
\multirow{2}{*}{$M{=}25$}
 & 0.719 & 0.674 & 0.565 & 0.781 & 0.447 & 0.617 \\
 & {\color{ForestGreen}$\uparrow$0.275} & {\color{red}$\downarrow$0.003} & {\color{red}$\downarrow$0.003} & {\color{red}$\downarrow$0.003} & {\color{red}$\downarrow$0.003} & {\color{red}$\downarrow$0.003} \\
\midrule
\multirow{2}{*}{$M{=}10$}
 & 0.662 & 0.676 & 0.567 & 0.783 & 0.449 & \textbf{0.619} \\
 & {\color{ForestGreen}$\uparrow$0.218} & {\color{red}$\downarrow$0.001} & {\color{red}$\downarrow$0.001} & {\color{red}$\downarrow$0.001} & {\color{red}$\downarrow$0.001} & {\color{red}$\downarrow$0.001} \\
\bottomrule
\end{tabular}%

\end{table}

\subsection{Capability retention under targeted fine-tuning}
\label{ab:retention}
A concern with any fine-tuning procedure is \emph{catastrophic forgetting}: training on a single target task can degrade the model's performance on tasks outside the fine-tuning distribution. We evaluate this on LLaMA-3-8B by fine-tuning only on ARC-C and then testing the resulting model on all five commonsense benchmarks; the pretrained LLaMA-3-8B zero-shot baseline (denoted \textit{Base}) serves as the reference. Results are reported in Table~\ref{tab:forgetting}. Full fine-tuning improves ARC-C substantially (from $0.4445$ to $0.728$, $+0.284$) but degrades all four untrained benchmarks, with an average drop of $0.066$ relative to \textit{Base}. Token masking produces a markedly different pattern. At $M{=}100$, the target accuracy reaches $0.764$---higher than full fine-tuning---while the average drop on the four untrained tasks is only $0.015$, roughly a $4{\times}$ reduction in forgetting. The effect strengthens as $M$ decreases: at $M{=}10$, the drop on untrained tasks falls to $0.001$ (essentially no forgetting), with ARC-C still improved by $0.218$ over \textit{Base}. This is consistent with Theorem~\ref{thm:capacity}: training through fewer positions confines parameter updates to a lower-dimensional subspace of the pretrained tangent space, which limits the extent to which base-model representations can be disrupted. In practice, small $M$ therefore offers a controllable trade-off between target-task improvement and preservation of general capabilities---a property that full-sequence fine-tuning does not provide.

\section{Extended Related Work}
\label{app:related-work}

We position the Token Hypothesis against four lines of work: parameter-efficient fine-tuning (PEFT), activation-memory reduction techniques, token-selective backpropagation, and Neural Tangent Kernel theories of fine-tuning. The first two address the GPU memory cost of fine-tuning along axes that are largely orthogonal to ours; the third targets the same axis (which positions to backpropagate through) but does so heuristically; and the fourth provides the theoretical lens our analysis builds on. Across all four, what is missing is a principled account of \emph{why} backpropagating through a small subset of positions suffices, what determines its optimal size, and how that optimum depends on the architecture. The Token Hypothesis fills this gap.

\subsection{Parameter-Efficient Fine-Tuning}
\label{app:rw-peft}

Parameter-efficient fine-tuning methods reduce the cost of adapting a large pretrained model by training only a small fraction of the parameters while keeping the backbone frozen. The dominant family is low-rank adaptation: LoRA~\cite{hu2022lora} parameterizes weight updates as $\Delta W = BA$ with $A \in \mathbb{R}^{r \times d}$, $B \in \mathbb{R}^{d \times r}$ for rank $r \ll d$, training only $A$ and $B$ while $W$ stays frozen. DoRA~\cite{liu2024dora} extends this by decomposing the pretrained weight into magnitude and direction components and applying the low-rank update only to the direction, improving expressiveness at no extra inference cost. Prefix tuning~\cite{li2021prefix} prepends a small set of learnable continuous vectors to each transformer layer's keys and values, leaving every model weight frozen. Adapter-based methods insert small trainable bottleneck modules between transformer sub-layers~\cite{houlsby2019parameter}, while $\mathrm{(IA)^3}$~\cite{liu2022few} multiplies activations by learned scaling vectors. RoCoFT~\cite{kowsher2025rocoft} restricts updates to a small number of rows and columns of weight matrices, achieving competitive accuracy at a tiny parameter footprint.

These methods all reduce the number of trainable parameters, which directly cuts gradient and optimizer-state memory: the optimizer no longer needs to store moments for the frozen backbone, and gradient buffers shrink in proportion. They do not, however, reduce the storage required for intermediate activations during the forward pass. Backpropagation through any low-rank adapter still requires the upstream activations of every layer to compute the chain rule through the adapter weights, so the activation footprint, which scales as batch size $\times$ sequence length $\times$ hidden dimension $\times$ depth, remains unchanged. As recent work notes, activations dominate the GPU memory budget for billion-scale models on long-context tasks~\cite{simoulin2024memory, zeng2026tokenseek}, which is precisely the regime where PEFT alone fails to deliver the expected savings. Our work targets this remaining bottleneck through position-wise gradient masking, and importantly, the two reductions compose without interference: applying token masking on top of LoRA or DoRA preserves the parameter savings while adding activation savings, as we demonstrate empirically in Section~\ref{sec:experiments}.

\subsection{Activation-Memory Reduction}
\label{app:rw-activation}

A separate body of work targets activation memory directly, independent of which parameters are trained. The classical approach is gradient checkpointing~\cite{chen2016training}, which stores only a strategically chosen subset of intermediate activations during the forward pass and recomputes the rest during backward, reducing memory from $O(L)$ to $O(\sqrt{L})$ for an $L$-layer network at the cost of one extra forward pass per checkpointed segment. This technique is now standard practice in large-scale training and is supported natively in major frameworks. Reversible networks take a different route: prior work reformulates residual blocks so that activations can be reconstructed exactly from outputs~\cite{gomez2017reversible}, and this idea has been extended to transformer architectures~\cite{kitaev2020reformer}. Reversibility eliminates the need to cache activations entirely, but constrains architectural choices and typically incurs a modest accuracy cost.

Activation compression and quantization form another line. Mesa~\cite{pan2021mesa} stores low-precision approximate activations during the forward pass and dequantizes them in the backward pass, trading a small accuracy gap for substantial memory savings. ZeRO~\cite{rajbhandari2020zero} and FSDP partition optimizer states, gradients, and parameters across devices, reducing per-device memory at the cost of inter-device communication. Mixed-precision training (FP16 or BF16) and 8-bit optimizer quantization, combined as in QLoRA~\cite{dettmers2023qlora}, further compress weight and optimizer storage. More recent work targets long-sequence training specifically: StreamBP~\cite{luo2025streambp} performs exact backpropagation while iteratively processing mini-sequences, reducing the per-step activation cost without approximation; sequence-parallel methods partition the sequence dimension across GPUs at the cost of inter-device communication.

All of these approaches treat the activation tensor as a monolithic object to be compressed, recomputed, partitioned, or reconstructed. None of them reduce the number of positions through which gradient information flows: every token position remains in the backward computation graph, and the savings come from how that computation is executed rather than from which subset of positions participates. Our work targets the orthogonal axis: by backpropagating through only $M$ of $T$ positions, we eliminate the activation cost of the remaining $T - M$ positions entirely, irrespective of how each retained activation is stored. The two strategies compose: gradient checkpointing on the $M$ retained positions further reduces the memory of those positions, while reversibility could in principle eliminate the cached activations even for the trainable subset. Our experiments use no checkpointing or reversibility, so the reported reductions are pure consequences of position selection, but the framework is compatible with both.

\subsection{Token-Selective Fine-Tuning}
\label{app:rw-tokensel}

The closest line of work to ours selects a subset of token positions to backpropagate through, materializing the activation savings we describe but without theoretical foundation. TokenTune~\cite{simoulin2024memory} backpropagates through $k$ randomly chosen input tokens (with the [CLS] token always included) and demonstrates that this approximate-gradient strategy preserves accuracy on BERT and Llama-2 across few-shot benchmarks. Composed with LoRA and QLoRA, it reaches roughly $21\%$ of full fine-tuning's memory at comparable accuracy. The selection is uniformly random per epoch, with no notion of which positions are more or less informative.

TokenSeek~\cite{zeng2026tokenseek} introduces an instance-aware criterion: each token receives a score $\alpha \log(I_1) + \beta \cdot \mathrm{Norm}(I_2)$ combining attention column-sums (a context-importance signal) and gradient magnitudes (a task-importance signal), and the top-$r\%$ of tokens are kept per sample. Default settings retain $10\%$ of tokens. The approach achieves strong results---on Llama-3.2-1B with QLoRA at $14.8\%$ memory, it reaches an average of $52.61$ on the few-shot benchmark suite, exceeding full token tuning. However, the criterion requires a pre-stage in which the model performs a forward and partial backward pass over the entire training set to compute scores, with the resulting per-sample score dictionary stored on disk. The hyperparameters $\alpha, \beta$ are grid-searched per model: $(1, 0)$ for Qwen-0.5B and Llama-1B, $(5, 5)$ for Llama-3B. The method is causal-only, exploiting the attention-sink structure of decoder models.

LeMo~\cite{wang2025lemo} takes an engineering-first approach: a small predictor network forecasts which tokens to keep at each layer, with elastic predictor sizing and kernel-level optimizations (permutation-free token movement, segment-based gradient computation) yielding end-to-end memory reductions of up to $1.93\times$ on long-context fine-tuning. Random-LTD~\cite{yao2022random} applies layerwise random token dropping during pretraining rather than fine-tuning, demonstrating that aggressive token-level sparsity is feasible during the forward pass.

These methods establish empirically that fine-tuning through a small subset of token positions preserves accuracy, but they leave the underlying mechanism unexplained. None offers a theoretical account of \emph{why} a small subset suffices, what determines the size of the sufficient subset, or how that size depends on architectural properties of the backbone. Data-driven scoring approaches~\cite{zeng2026tokenseek} implicitly assume that token importance varies systematically by content; our analysis shows that in causal models the variation is dominated by \emph{position} rather than content, with later positions strictly dominating earlier ones under a contraction condition on hidden states (Theorem~\ref{thm:monotone}). This positional structure explains why gradient-based scoring concentrates on the response section of Alpaca-formatted training data, which is exactly the suffix of the sequence and exactly what our last-$M$ rule selects without any per-sample computation. We therefore expect to match the accuracy of these methods at zero pre-stage cost and without per-model hyperparameter search.

A complementary thread on token pruning has emerged in the inference-time literature. LazyLLM~\cite{fu2024lazyllm} dynamically prunes prompt tokens during the prefill stage to accelerate first-token generation, achieving $2.34\times$ speedup on Llama-2-7B without fine-tuning. SlimInfer~\cite{long2026sliminfer} extends pruning to intermediate hidden states under an information-diffusion observation, and InfiniteHiP~\cite{lee2025infinitehip} uses hierarchical pruning to extend the effective context to three million tokens on a single GPU. These methods address a different problem (inference compute and KV-cache memory rather than training activation memory) and operate without weight updates, but they corroborate the broader empirical finding that transformer computation is highly sparse in token utilization. Our work transports this insight to the training regime, where the relevant cost is activation storage rather than KV-cache size, and grounds it in NTK theory rather than attention-pattern heuristics.

\subsection{Neural Tangent Kernel Theory of Fine-Tuning}
\label{app:rw-ntk}

The Neural Tangent Kernel framework~\cite{jacot2018neural} characterizes neural-network training in the infinite-width limit as kernel regression with the kernel induced by the network's Jacobians at initialization. Originally derived for randomly initialized networks, the framework has since been adapted to fine-tuning of pretrained models. Prior work~\cite{malladi2023kernel} extends NTK analysis to the fine-tuning regime, demonstrating that prompt-based fine-tuning of large language models can exhibit kernel behavior under specific conditions on prompts and training scale, and proves that LoRA can be analyzed through a kernel-difference bound relative to full fine-tuning. A more recent line~\cite{afzal2026linearization} introduces an explicit $\ell_2$ inductive bias on parameters to ensure linearization is a faithful model of fine-tuning dynamics, and uses the eigenvalue spectrum of the resulting NTK to predict downstream performance and guide layer selection. Other work~\cite{tomihari2024understanding} applies NTK analysis to two-stage linear-probe-then-fine-tune training, decomposing the kernel matrix into pretrained-effective and fine-tuning-effective components.

These works address \emph{which parameters} or \emph{which layers} to fine-tune, treating the input sequence as a monolithic object whose tokens contribute uniformly to the kernel. Our analysis introduces a complementary axis: the kernel induced by training through a single \emph{token position}, formalized as the Gram matrix $K_t$ of position-$t$ Jacobians. We show that this position-wise kernel is itself a low-dimensional bottleneck (Theorem~\ref{thm:capacity}), that positions with aligned hidden-state geometry yield equivalent kernels in bidirectional models (Theorem~\ref{thm:equal}), and that position-wise kernels form a monotone chain in causal models under a contraction condition (Theorem~\ref{thm:monotone}). The output-head NTK regime in which our results are stated is closely aligned with the kernel formalism of prior fine-tuning NTK work~\cite{malladi2023kernel}, and the empirical NTK-laziness validation we provide on LLaMA-3-8B (Appendix~\ref{ab:laziness}) is in the spirit of recent linearization measurements~\cite{afzal2026linearization}. The novelty of our contribution is the position-wise refinement: existing NTK theories of fine-tuning are blind to which positions of the sequence drive learning, whereas we show that this question has both a clean theoretical answer (an architecture-dependent optimal position count $M^*$) and a directly actionable consequence (a memory-efficient training procedure).

\subsection{Synthesis}
\label{app:rw-synthesis}

The four lines of related work can be summarized along a single axis: each reduces fine-tuning cost by acting on a different dimension of the computation. PEFT methods reduce the parameter dimension. Activation-memory techniques (checkpointing, reversibility, compression) reduce per-position storage cost. Token-selective methods reduce the number of positions in the backward graph but rely on heuristics or auxiliary computation to choose which positions. NTK theories of fine-tuning explain training dynamics but treat positions uniformly. Our work occupies the intersection that none of the four cover: a theoretical account of position-wise capacity that yields, as a consequence, a principled criterion for how many and which positions to backpropagate through, with the resulting memory savings composing cleanly with PEFT methods on the parameter side and remaining orthogonal to activation-side techniques on the storage side. The empirical headline---fifty to one hundred trainable positions match or exceed full-sequence fine-tuning across six task families and five backbones, with up to ninety-seven percent activation-memory reduction---is the practical corollary of the theory rather than the result of an engineering search.

\section{Datasets}
\label{app:datasets}

We evaluate on six task modalities spanning causal and bidirectional architectures. This section describes each dataset used in the paper.

\subsection{Commonsense Reasoning} We use six benchmarks to evaluate general commonsense and factual reasoning abilities in pretrained language models by following work and setups~\cite{hu2023llm}.
\begin{itemize}\itemsep 2pt
    \item \textbf{ARC-Challenge (ARC-C)}~\cite{clark2018think}: a multiple-choice science QA dataset of grade-school questions that are hard for retrieval-based methods, requiring genuine reasoning. Each sample contains a question and four labeled options.
    \item \textbf{ARC-Easy (ARC-E)}~\cite{clark2018think}: the easier subset of the ARC benchmark, consisting of elementary-level science questions in the same four-option format.
    \item \textbf{BoolQ}~\cite{clark2019boolq}: a yes/no question answering dataset where each question is paired with a short passage, and the model must answer \texttt{true} or \texttt{false}.
    \item \textbf{HellaSwag}~\cite{zellers2019hellaswag}: a commonsense sentence-completion benchmark where the model selects the most plausible continuation of a short context from four adversarially-generated candidates.
    \item \textbf{OpenBookQA (OBQA)}~\cite{mihaylov2018can}: a multiple-choice elementary-science dataset designed to evaluate reasoning that combines a provided ``open book'' of core science facts with broader commonsense.
    \item \textbf{WinoGrande}~\cite{sakaguchi2021winogrande}: a large-scale Winograd-style pronoun resolution dataset requiring commonsense understanding of events, objects, and everyday situations.
\end{itemize}

\subsection{Mathematical Reasoning} We use six arithmetic and word-problem benchmarks, following the standard setup in prior PEFT literature by following work and setups~\cite{hu2023llm}.
\begin{itemize}\itemsep 2pt
    \item \textbf{AddSub}~\cite{hosseini2014learning}: a collection of simple arithmetic word problems involving addition and subtraction over small integers.
    \item \textbf{MultiArith}~\cite{roy2015solving}: multi-step arithmetic word problems that require combining several operations to reach the final answer.
    \item \textbf{SVAMP}~\cite{patel2021nlp}: a challenge benchmark constructed to reveal over-reliance on surface heuristics in word-problem solving, built from targeted perturbations of existing problems.
    \item \textbf{SingleEq}~\cite{koncel2015parsing}: algebra word problems solvable by a single equation, covering linear and multi-operand cases.
    \item \textbf{GSM8K}~\cite{cobbe2021training}: grade-school math word problems that require multi-step natural-language reasoning with calculator-level arithmetic.
    \item \textbf{MAWPS}~\cite{koncel2016mawps}: a unified math word-problem suite that aggregates several earlier arithmetic benchmarks into a common format.
\end{itemize}

\subsection{Visual Question Answering} We evaluate five visual QA benchmarks spanning medical, scene-text, chain-of-thought, K--12, and safety-critical settings.
\begin{itemize}\itemsep 2pt
    \item \textbf{VQA-RAD}~\cite{lau2018dataset}: a radiology visual QA dataset containing clinician-generated questions paired with radiology images (CT, MRI, X-ray). Each question is labeled as closed-ended (yes/no or a small closed set) or open-ended.
    \item \textbf{STVQA}~\cite{biten2019scene} : scene-text visual question answering, where answers depend on reading and interpreting text embedded in natural images.
    \item \textbf{M3CoT}~\cite{chen2024m3cot} : a multi-modal chain-of-thought reasoning benchmark requiring intermediate reasoning steps grounded in both image and text.
    \item \textbf{MMK12}~\cite{meng2025mm}: a K--12 multi-modal reasoning dataset covering science, math, and general knowledge with images. Since only a training split is released, we partition it $80/20$ into train/test with seed $42$.
    \item \textbf{SafeTag-VL-3K}~\cite{rong2025safegrpo}: an image-grounded safety benchmark evaluating whether a vision-language model can identify safety-critical content in images. We filter out samples with empty image fields.
\end{itemize}

\subsection{Video Question Answering} We evaluate six video-reasoning benchmarks spanning physics, interaction, temporal, and mathematical domains.
\begin{itemize}\itemsep 2pt
    \item \textbf{Human-Object Interactions (HOI)}~\cite{krojer2025shortcut}: a subset of \texttt{facebook/minimal\_video\_pairs} focused on identifying the correct description of how humans manipulate objects in short video clips.
    \item \textbf{Intuitive Physics}~\cite{krojer2025shortcut}: a subset of \texttt{facebook/minimal\_video\_pairs} testing intuitive physical reasoning about objects, forces, and motion.
    \item \textbf{Robot-Object Interactions}~\cite{krojer2025shortcut}: a subset of \texttt{facebook/minimal\_video\_pairs} evaluating understanding of robotic manipulation in controlled scenes.
    \item \textbf{Temporal Reasoning}~\cite{krojer2025shortcut}: a subset of \texttt{facebook/minimal\_video\_pairs} requiring the model to identify the temporal order or causal structure of events in a video.
    \item \textbf{VideoMathQA}~\cite{rasheed2025videomathqa} (\texttt{MBZUAI/VideoMathQA}, \texttt{multi\_binary} subset): a mathematical-reasoning benchmark grounded in instructional videos, where the model must answer multi-step math questions about content demonstrated on video. Only a test split is released; we partition it $80/20$ with seed $42$.
    \item \textbf{Video-LevelGauge}~\cite{xia2025video}: a long-video multiple-choice benchmark evaluating comprehension and reasoning over extended temporal contexts.
\end{itemize}
For all video datasets we decode frames with \texttt{decord} at $1$~FPS, capped at $8$~frames per video, and filter out samples whose videos are missing or cannot be decoded.

\paragraph{GLUE Benchmark.} We use eight tasks from the General Language Understanding Evaluation (GLUE) benchmark~\cite{wang2018glue} to evaluate natural-language understanding in bidirectional models.
\begin{itemize}\itemsep 2pt
    \item \textbf{SST-2}: binary sentiment classification of movie-review sentences.
    \item \textbf{QNLI}: question-answering natural language inference derived from SQuAD, recast as a sentence-pair entailment task.
    \item \textbf{MRPC}: paraphrase identification between pairs of sentences taken from news sources, evaluated with accuracy and F1.
    \item \textbf{QQP}: a large-scale paraphrase-identification dataset of Quora question pairs, evaluated with accuracy and F1.
    \item \textbf{CoLA}: grammatical-acceptability classification, evaluated with Matthews correlation coefficient.
    \item \textbf{MNLI}: multi-genre natural language inference, a three-way entailment task across ten genres of written and spoken English.
    \item \textbf{RTE}: recognizing textual entailment, a binary entailment task compiled from earlier textual-entailment challenges.
    \item \textbf{STS-B}: sentence-pair semantic similarity scored on a $0$--$5$ scale, evaluated with Pearson and Spearman correlation.
\end{itemize}

\subsection{Image Classification} We evaluate on five image-classification benchmarks at $224 \times 224$ resolution.
\begin{itemize}\itemsep 2pt
    \item \textbf{Tomato Leaves}~\cite{singh2020plantdoc} (\texttt{Vlone571/tomato-leaves-dataset}): an $8$-class plant-disease dataset of tomato-leaf images, covering healthy leaves and seven common diseases.
    \item \textbf{SUN397}~\cite{xiao2010sun}: a large-scale scene-classification benchmark with $397$ scene categories covering indoor, outdoor, and urban environments.
    \item \textbf{RVL-CDIP}~\cite{harley2015evaluation}: a $16$-class document-image classification dataset spanning letters, forms, emails, invoices, and other common document types.
    \item \textbf{EuroSAT}~\cite{helber2019eurosat}: a $10$-class satellite-image classification benchmark based on Sentinel-2 imagery, covering land-use and land-cover categories.
    \item \textbf{ChestX-Ray (Pneumonia)}~\cite{kermany2018labeled} : a binary medical-imaging dataset classifying chest X-rays as normal or pneumonia.
\end{itemize}

\newpage

\section{Hyperparameters and Training Configuration}
\label{app:hyperparameters}

This section documents the hyperparameters used across all experiments in the paper. All training runs use a single set of shared defaults for optimization, mixed precision, and scheduling; dataset- and architecture-specific deviations are noted below. Every run uses the random seed $42$ for dataset splitting and trainer initialization.

\paragraph{Shared optimization settings.} All experiments use the AdamW optimizer with a cosine learning-rate schedule, $100$ linear warmup steps, bfloat16 mixed precision, and no gradient clipping (except image classification with ViT, which uses the default $1.0$). Training runs for $1$ epoch unless otherwise specified. We disable checkpoint saving during sweeps (\texttt{save\_strategy = "no"}) to avoid disk overhead. Evaluation uses greedy decoding for all generative tasks.

\paragraph{Full fine-tuning.} Full fine-tuning updates all trainable parameters with learning rate $2 \times 10^{-5}$ and weight decay $0$. We freeze all \texttt{Conv3d} layers in vision-language models before training to avoid corrupting the pretrained vision patch embeddings.

\paragraph{LoRA.} LoRA~\cite{hu2022lora} adapters use rank $r = 4$, scaling factor $\alpha = 5$, and dropout $0.05$, with learning rate $1 \times 10^{-4}$ and weight decay $0$. We apply LoRA only to the attention projection matrices (\texttt{q\_proj}, \texttt{k\_proj}, \texttt{v\_proj}, \texttt{o\_proj}) of the language-model backbone (\texttt{model.language\_model}); the vision encoder remains frozen throughout training. After training, LoRA updates are merged into the base weights, restoring the original architecture with no inference overhead.

\paragraph{DoRA.} DoRA~\cite{liu2024dora} uses the same rank, scaling, dropout, and learning rate as LoRA, and is applied to the same attention projections. The only difference is the weight-decomposition step introduced by DoRA itself.

Table~\ref{tab:hparams-summary} summarizes the per-modality training settings; Tables~\ref{tab:hparams-context} report sequence-length and generation settings.

\begin{table}[ht!]
\centering
\caption{Per-modality training hyperparameters. ``bs'' is per-device batch size; ``accum'' is gradient-accumulation steps; ``eff.~bs'' is the effective batch size (bs $\times$ accum). Multi-GPU runs shard the model across devices (\texttt{device\_map="auto"}), so the GPU count does not change the effective batch size. GPUs are NVIDIA H100 80GB.}
\label{tab:hparams-summary}
\small
\setlength{\tabcolsep}{4pt}
\begin{tabular}{@{}llllrrrrc@{}}
\toprule
Modality & Backbone & Regime & LR & bs & accum & eff.~bs & Epochs & GPUs \\
\midrule
Commonsense reasoning    & LLaMA-3-8B          & Full  & $2\mathrm{e}{-5}$ & 4  & 4 & 16 & 1 & 2 \\
                         & LLaMA-3-8B          & DoRA  & $1\mathrm{e}{-4}$ & 4  & 4 & 16 & 1 & 1 \\
\midrule
Math reasoning           & LLaMA-3-8B          & Full  & $2\mathrm{e}{-5}$ & 4  & 4 & 16 & 1 & 2 \\
                         & LLaMA-3-8B          & DoRA  & $1\mathrm{e}{-4}$ & 4  & 4 & 16 & 1 & 1 \\
\midrule
Visual QA (image)        & Qwen3-VL-8B         & Full  & $2\mathrm{e}{-5}$ & 3  & 4 & 12 & 1 & 4 \\
                         & Qwen3-VL-8B         & LoRA  & $1\mathrm{e}{-4}$ & 2  & 4 & 8  & 1 & 2 \\
                         & LLaVA-OneVision-7B  & Full  & $1\mathrm{e}{-4}$ & 3  & 4 & 12 & 1 & 2 \\
                         & LLaVA-OneVision-7B  & LoRA  & $1\mathrm{e}{-4}$ & 3  & 4 & 12 & 1 & 1 \\
\midrule
Video QA (minimal pairs) & Qwen3-VL-8B         & Full  & $2\mathrm{e}{-5}$ & 8  & 4 & 32 & 1 & 2 \\
                         & Qwen3-VL-8B         & LoRA  & $1\mathrm{e}{-4}$ & 8  & 4 & 32 & 1 & 1 \\
\midrule
Video QA (VideoMathQA)   & Qwen3-VL-8B         & Full  & $2\mathrm{e}{-5}$ & 1  & 4 & 4  & 1 & 2 \\
                         & Qwen3-VL-8B         & LoRA  & $1\mathrm{e}{-4}$ & 1  & 4 & 4  & 1 & 1 \\
\midrule
Video QA (LevelGauge)    & Qwen3-VL-8B         & Full  & $2\mathrm{e}{-5}$ & 2  & 4 & 8  & 1 & 2 \\
                         & Qwen3-VL-8B         & LoRA  & $1\mathrm{e}{-4}$ & 2  & 4 & 8  & 1 & 2 \\
\midrule
GLUE                     & RoBERTa-base        & Full  & $2\mathrm{e}{-5}$ & 32 & 1 & 32 & 3 & 1 \\
Image classification     & ViT-L/16-224        & Full  & $1\mathrm{e}{-4}$ & 32 & 2 & 64 & 1 & 1 \\
\bottomrule
\end{tabular}
\end{table}

\begin{table}[ht!]
\centering
\caption{Sequence lengths and generation settings per dataset.}
\label{tab:hparams-context}
\small
\setlength{\tabcolsep}{4pt}
\begin{tabular}{lllll}
\toprule
Dataset & Input max length & Frames / FPS & Max new tokens \\
\midrule
Commonsense (all)         & 1000 & --        & 16  \\
Math (all)                & 1000 & --        & 1000  \\
VQA-RAD                   & 4048 & --        & 64  \\
STVQA                     & 4048 & --        & 512 \\
M3CoT                     & 4048 & --        & 512 \\
MMK12                     & 8096 & --        & 512 \\
SafeTag-VL-3K             & 4024 & --        & 16  \\
HOI / IntuPhys / RobotOI / TempReas & 2048 & 8 frames at 1 FPS & 8  \\
VideoMathQA               & 4096 & 8 frames at 1 FPS & 512 \\
Video-LevelGauge          & 4096 & 8 frames at 1 FPS & 64  \\
GLUE (all)                & 128  & --        & --  \\
Image classification      & $224\!\times\!224$ & -- & -- \\
\bottomrule
\end{tabular}
\end{table}

\paragraph{LoRA / DoRA target specification.} For all vision-language models (Qwen3-VL-8B and LLaVA-OneVision-7B), adapters target only the language-model attention projections:
\begin{itemize}\itemsep 2pt
    \item \texttt{target\_modules}: \texttt{[q\_proj, k\_proj, v\_proj, o\_proj]}
    \item \texttt{name\_prefix}: \texttt{model.language\_model}
\end{itemize}
For ViT (bidirectional), full fine-tuning targets the encoder attention and MLP projections:
\begin{itemize}\itemsep 2pt
    \item \texttt{target\_modules}: \texttt{[query, key, value, dense]}
    \item \texttt{name\_prefix}: \texttt{vit.encoder}
\end{itemize}
For LLaMA-3-8B (full fine-tuning and DoRA), all linear layers in every transformer block are trainable (\texttt{target\_modules = "all"}), with frozen embedding layers.

\paragraph{Sweep over trainable token counts.} For each experiment, we sweep the number of trainable positions $M$ over the following values plus a full-sequence baseline:
\begin{itemize}\itemsep 2pt
    \item Text (commonsense, math): $M \in \{10, 25, 50, 100, 300, 500, 700, 1000\}$.
    \item Visual QA: $M \in \{10, 25, 50, 100, 200, 300, 500, 700, 1000, 1500\}$.
    \item Video QA: $M \in \{10, 25, 50, 100, 300, 500, 700, 1000, 1500\}$.
    \item GLUE: $M \in \{5, 10, 25, 50, 75, 100\}$.
    \item Image classification: $M \in \{10, 25, 50, 100, 150, 180\}$.
\end{itemize}
The full-sequence baseline (labeled ``All'' in every table) corresponds to standard training with no position masking: all label positions contribute to the loss.

\paragraph{Memory and timing measurement.} Peak GPU memory is measured with \texttt{torch.cuda.max\_memory\_allocated}, after resetting the peak-memory statistics immediately before \texttt{trainer.train()} is invoked. The reported value therefore captures the complete training footprint, comprising the resident model parameters together with the gradients, optimizer states, and activations allocated during training, while excluding transient allocations incurred during model loading. For multi-GPU configurations, we report the sum of the peak allocations across all devices; reported values for sharded models thus correspond to total rather than per-device memory. Training time is the wall-clock duration of the training loop and excludes model loading and evaluation.

\paragraph{Hardware.} With the exception of the long-context and gradient-checkpointing experiments, all experiments are conducted on NVIDIA H100 GPUs with 80~GB of memory. Multi-GPU configurations employ the HuggingFace \texttt{device\_map="auto"} option to shard the model across devices. Because this scheme is model-parallel rather than data-parallel, the number of GPUs does not affect the effective batch size. A single GPU is used for DoRA fine-tuning of LLaMA-3-8B, LoRA fine-tuning of LLaVA-OneVision-7B, LoRA fine-tuning on the video benchmarks other than Video-LevelGauge, the GLUE experiments with RoBERTa-base, and image classification with ViT-L/16-224. Two GPUs are used for full fine-tuning of LLaMA-3-8B and LLaVA-OneVision-7B, LoRA fine-tuning of Qwen3-VL-8B on image QA, full fine-tuning of Qwen3-VL-8B on all video benchmarks, and LoRA fine-tuning on Video-LevelGauge. Four GPUs are used for full fine-tuning of Qwen3-VL-8B on image QA. Table~\ref{tab:hparams-summary} summarizes the GPU configuration for each setting. The long-context and gradient-checkpointing experiments on LongAlpaca-12k (Appendices~J.7 and~J.8) are conducted on eight NVIDIA A100 GPUs with 80~GB of memory each.

\paragraph{Precision and inference.} All models are loaded and trained in bfloat16. Training uses automatic mixed precision via the HuggingFace \texttt{Trainer} with \texttt{bf16=True}. After training, adapter weights (for LoRA and DoRA) are merged back into the base parameters via the \texttt{merge} operation, so the inference-time architecture is identical to the pretrained model---no additional adapter modules remain at test time and no latency overhead is introduced.

\paragraph{Parameter freezing for vision-language models.} We emphasize that for all vision-language experiments (Qwen3-VL-8B and LLaVA-OneVision-7B), the vision patch embeddings (\texttt{Conv3d} for Qwen3-VL, \texttt{Conv2d} for LLaVA) are frozen before training to avoid perturbing the pretrained patch-tokenization. Adapter updates target exclusively the language-model attention projections; this isolation ensures that any observed accuracy changes under token masking are attributable to the language-side training signal, not to shifts in visual representation.

\newpage
\section{Models}
\label{app:models}

We evaluate the Token Hypothesis across five backbone architectures spanning text-only causal language models, multimodal vision-language models, bidirectional encoders, and vision transformers. This section describes each model's architecture, pretraining, and the components that are modified or frozen during fine-tuning.

\paragraph{LLaMA-3-8B (causal).} LLaMA-3-8B~\cite{grattafiori2024llama} is an $8$-billion-parameter causal transformer trained on approximately $15$ trillion tokens of publicly available text. The architecture consists of $32$ transformer blocks, each with grouped-query attention ($32$ query heads and $8$ key-value heads), SwiGLU activations, rotary positional embeddings, and RMSNorm. The hidden dimension is $4096$ and the feed-forward dimension is $14{,}336$. The vocabulary has $128{,}256$ tokens with a byte-pair encoded tokenizer. We use the base (non-instruction-tuned) checkpoint for all commonsense and mathematical reasoning experiments and load it in bfloat16 precision. Under the token-masking protocol, the causal attention ensures that each hidden state at position $t$ depends only on the prefix $(x_1, \ldots, x_t)$, which is the setting assumed by Theorem~\ref{thm:monotone}.

\paragraph{Qwen3-VL-8B-Instruct (causal, vision-language).} Qwen3-VL-8B-Instruct~\cite{yang2025qwen3} is an instruction-tuned multimodal model that extends the Qwen3 language backbone with a native vision encoder. The language model is an $8$-billion-parameter causal transformer with grouped-query attention and rotary embeddings, and the vision encoder is a dynamic-resolution vision transformer that tokenizes images and videos into variable-length patch sequences. Visual tokens are projected into the language-model embedding space through an adaptor and interleaved with text tokens in the unified sequence. For video inputs, temporal patches are produced by a \texttt{Conv3d} patch embedding. In all our Qwen3-VL experiments, we freeze every \texttt{Conv3d} layer in the vision encoder before training and apply LoRA only to the language-model attention projections under \texttt{model.language\_model}, following the orthogonality principle of the method: parameter-efficient fine-tuning targets the language pathway, while the vision encoder remains untouched.

\paragraph{LLaVA-OneVision-7B (causal, vision-language).} LLaVA-OneVision-7B~\cite{li2024llava} is a vision-language model that couples a Qwen2-7B causal language model with a SigLIP vision encoder. Images are processed at $384 \times 384$ resolution and projected into the language-model token space via a two-layer MLP adaptor. For all LLaVA experiments we fix \texttt{image\_grid\_pinpoints = [[384, 384]]} to disable dynamic tiling, and we apply LoRA/DoRA to the same language-model attention projections as for Qwen3-VL. The \texttt{Conv2d} patch embedding in the vision tower is frozen in the DoRA setting; for LoRA and full fine-tuning the vision encoder is trainable but empirically moves negligibly (consistent with the NTK-laziness analysis in Appendix~\ref{ab:laziness}).

\paragraph{RoBERTa-base (bidirectional).} RoBERTa-base~\cite{liu2019roberta} is a $125$-million-parameter bidirectional encoder pretrained on approximately $160$ GB of English text with the masked-language-modeling objective. The architecture consists of $12$ transformer blocks with multi-head self-attention ($12$ heads), GELU activations, learned absolute positional embeddings, and LayerNorm. Hidden dimension is $768$, feed-forward dimension is $3072$, and the tokenizer vocabulary has $50{,}265$ byte-pair tokens. Since every position attends to the full sequence, RoBERTa is the canonical bidirectional setting for Theorem~\ref{thm:equal}. We use RoBERTa-base for all GLUE experiments and for the empirical validation of Theorems~\ref{thm:capacity} and~\ref{thm:equal} (Section \ref{sec:experiments} and Appendices  A.2 and B.2), and for the image-saturation analysis of Corollary 1 (Figure 2).

\paragraph{SmolLM2-135M (causal).} SmolLM2-135M~\cite{allal2025smollm2} is a $135$-million-parameter causal language model pretrained on curated web data. We use it only for the multi-scale validation of Theorem~\ref{thm:capacity} (Appendix~\ref{app:exp-thm1}), where it provides a mid-size causal architecture between RoBERTa-base ($125$M) and LLaMA-3-8B ($8$B), allowing us to verify that the effective-rank bottleneck $d_{\mathrm{eff}}(K_t) < 6$ holds across a $64\times$ range in parameter count. SmolLM2-135M is not used for any downstream fine-tuning experiment.

\paragraph{ViT-L/16-224 (bidirectional).} ViT-L/16-224~\cite{dosovitskiy2020image} is a $307$-million-parameter vision transformer pretrained on ImageNet-21k at $224 \times 224$ resolution with $16 \times 16$ patches, yielding a sequence of $196$ visual tokens plus a \texttt{[CLS]} token. The architecture consists of $24$ encoder blocks with $16$ attention heads, hidden dimension $1024$, and feed-forward dimension $4096$. Like RoBERTa, self-attention is fully bidirectional over the token sequence, so all image patch positions share information access. For each classification task we replace the pretrained classification head with a randomly initialized linear layer of the appropriate output dimension, ignoring mismatched sizes in the state dict. Fine-tuning targets the attention projections and the feed-forward \texttt{dense} layers across all encoder blocks.

\newpage

\begin{table*}[ht!]
\centering\scriptsize
\setlength{\tabcolsep}{3pt}
\renewcommand{\arraystretch}{1.1}
\caption{
  Full fine-tuning accuracy, total GPU memory~(GB), and training time~(min) across five commonsense reasoning
  benchmarks vs.\ number of trainable tokens. Subscript deltas are absolute deviations from
  \textit{All}: \gd{$\uparrow$}~green = improvement,\ \rd{$\downarrow$}~red = degradation.
  \textbf{Bold} = best value per column.
}
\label{tab:main_results}
\resizebox{\textwidth}{!}{%
\begin{tabular}{@{}lrrrrrrrrrrrrrrrrrr@{}}
\toprule
\multirow{2}{*}{\textbf{Tokens}} & \multicolumn{3}{c}{\textbf{ARC-C}} & \multicolumn{3}{c}{\textbf{ARC-E}} & \multicolumn{3}{c}{\textbf{BoolQ}} & \multicolumn{3}{c}{\textbf{HSwag}} & \multicolumn{3}{c}{\textbf{OBQA}} & \multicolumn{3}{c}{\textbf{Avg}} \\
\cmidrule(lr){2-4}\cmidrule(lr){5-7}\cmidrule(lr){8-10}\cmidrule(lr){11-13}\cmidrule(lr){14-16}\cmidrule(lr){17-19}
 & \textit{Acc} & \textit{GPU} & \textit{Time} & \textit{Acc} & \textit{GPU} & \textit{Time} & \textit{Acc} & \textit{GPU} & \textit{Time} & \textit{Acc} & \textit{GPU} & \textit{Time} & \textit{Acc} & \textit{GPU} & \textit{Time} & \textit{Acc} & \textit{GPU} & \textit{Time} \\
\midrule
\rowcolor{basecol}
\textit{All} & 0.728 & 94.68 & 2.32 & 0.863 & 94.68 & 4.66 & 0.738 & 94.68 & 19.53 & 0.951 & 97.67 & 82.44 & 0.874 & 94.68 & 10.32 & 0.831 & 95.28 & 23.85 \\
10 & \shortstack{0.662\\\rd{\tiny $\downarrow$0.066}} & \shortstack{\textbf{83.42}\\\gd{\tiny $\uparrow$11.26}} & \shortstack{\textbf{1.89}\\\gd{\tiny $\uparrow$0.43}} & \shortstack{0.811\\\rd{\tiny $\downarrow$0.052}} & \shortstack{\textbf{83.42}\\\gd{\tiny $\uparrow$11.26}} & \shortstack{\textbf{3.79}\\\gd{\tiny $\uparrow$0.87}} & \shortstack{0.700\\\rd{\tiny $\downarrow$0.038}} & \shortstack{\textbf{83.42}\\\gd{\tiny $\uparrow$11.26}} & \shortstack{\textbf{15.81}\\\gd{\tiny $\uparrow$3.72}} & \shortstack{0.835\\\rd{\tiny $\downarrow$0.116}} & \shortstack{\textbf{84.77}\\\gd{\tiny $\uparrow$12.90}} & \shortstack{\textbf{66.95}\\\gd{\tiny $\uparrow$15.49}} & \shortstack{0.712\\\rd{\tiny $\downarrow$0.162}} & \shortstack{\textbf{83.42}\\\gd{\tiny $\uparrow$11.26}} & \shortstack{\textbf{8.29}\\\gd{\tiny $\uparrow$2.03}} & \shortstack{0.744\\\rd{\tiny $\downarrow$0.087}} & \shortstack{\textbf{83.69}\\\gd{\tiny $\uparrow$11.59}} & \shortstack{\textbf{19.35}\\\gd{\tiny $\uparrow$4.51}} \\
\rowcolor{rowA}
25 & \shortstack{0.719\\\rd{\tiny $\downarrow$0.008}} & \shortstack{83.65\\\gd{\tiny $\uparrow$11.03}} & \shortstack{\textbf{1.89}\\\gd{\tiny $\uparrow$0.43}} & \shortstack{0.853\\\rd{\tiny $\downarrow$0.010}} & \shortstack{83.65\\\gd{\tiny $\uparrow$11.03}} & \shortstack{3.80\\\gd{\tiny $\uparrow$0.86}} & \shortstack{0.732\\\rd{\tiny $\downarrow$0.006}} & \shortstack{83.65\\\gd{\tiny $\uparrow$11.03}} & \shortstack{15.89\\\gd{\tiny $\uparrow$3.64}} & \shortstack{0.903\\\rd{\tiny $\downarrow$0.048}} & \shortstack{84.94\\\gd{\tiny $\uparrow$12.73}} & \shortstack{67.34\\\gd{\tiny $\uparrow$15.10}} & \shortstack{0.796\\\rd{\tiny $\downarrow$0.078}} & \shortstack{83.65\\\gd{\tiny $\uparrow$11.03}} & \shortstack{8.34\\\gd{\tiny $\uparrow$1.98}} & \shortstack{0.801\\\rd{\tiny $\downarrow$0.030}} & \shortstack{83.91\\\gd{\tiny $\uparrow$11.37}} & \shortstack{19.45\\\gd{\tiny $\uparrow$4.40}} \\
50 & \shortstack{0.761\\\gd{\tiny $\uparrow$0.033}} & \shortstack{84.07\\\gd{\tiny $\uparrow$10.61}} & \shortstack{1.93\\\gd{\tiny $\uparrow$0.39}} & \shortstack{\textbf{0.879}\\\gd{\tiny $\uparrow$0.016}} & \shortstack{84.07\\\gd{\tiny $\uparrow$10.61}} & \shortstack{3.87\\\gd{\tiny $\uparrow$0.79}} & \shortstack{0.730\\\rd{\tiny $\downarrow$0.008}} & \shortstack{84.07\\\gd{\tiny $\uparrow$10.61}} & \shortstack{16.18\\\gd{\tiny $\uparrow$3.35}} & \shortstack{0.952\\\gd{\tiny $\uparrow$0.001}} & \shortstack{86.89\\\gd{\tiny $\uparrow$10.78}} & \shortstack{68.56\\\gd{\tiny $\uparrow$13.88}} & \shortstack{0.834\\\rd{\tiny $\downarrow$0.040}} & \shortstack{84.07\\\gd{\tiny $\uparrow$10.61}} & \shortstack{8.49\\\gd{\tiny $\uparrow$1.83}} & \shortstack{0.831\\\gd{\tiny $\uparrow$0.001}} & \shortstack{84.63\\\gd{\tiny $\uparrow$10.65}} & \shortstack{19.81\\\gd{\tiny $\uparrow$4.05}} \\
\rowcolor{rowA}
100 & \shortstack{\textbf{0.764}\\\gd{\tiny $\uparrow$0.036}} & \shortstack{84.84\\\gd{\tiny $\uparrow$9.85}} & \shortstack{1.99\\\gd{\tiny $\uparrow$0.33}} & \shortstack{0.872\\\gd{\tiny $\uparrow$0.009}} & \shortstack{84.84\\\gd{\tiny $\uparrow$9.85}} & \shortstack{4.00\\\gd{\tiny $\uparrow$0.66}} & \shortstack{\textbf{0.745}\\\gd{\tiny $\uparrow$0.007}} & \shortstack{84.84\\\gd{\tiny $\uparrow$9.85}} & \shortstack{16.78\\\gd{\tiny $\uparrow$2.75}} & \shortstack{0.941\\\rd{\tiny $\downarrow$0.010}} & \shortstack{89.90\\\gd{\tiny $\uparrow$7.78}} & \shortstack{70.87\\\gd{\tiny $\uparrow$11.57}} & \shortstack{0.848\\\rd{\tiny $\downarrow$0.026}} & \shortstack{84.84\\\gd{\tiny $\uparrow$9.85}} & \shortstack{8.79\\\gd{\tiny $\uparrow$1.53}} & \shortstack{0.834\\\gd{\tiny $\uparrow$0.003}} & \shortstack{85.85\\\gd{\tiny $\uparrow$9.43}} & \shortstack{20.49\\\gd{\tiny $\uparrow$3.37}} \\
300 & \shortstack{0.761\\\gd{\tiny $\uparrow$0.033}} & \shortstack{87.85\\\gd{\tiny $\uparrow$6.83}} & \shortstack{2.14\\\gd{\tiny $\uparrow$0.18}} & \shortstack{0.852\\\rd{\tiny $\downarrow$0.011}} & \shortstack{87.85\\\gd{\tiny $\uparrow$6.83}} & \shortstack{4.30\\\gd{\tiny $\uparrow$0.36}} & \shortstack{0.744\\\gd{\tiny $\uparrow$0.007}} & \shortstack{87.85\\\gd{\tiny $\uparrow$6.83}} & \shortstack{18.01\\\gd{\tiny $\uparrow$1.52}} & \shortstack{0.951\\\gd{\tiny $\uparrow$0.000}} & \shortstack{92.96\\\gd{\tiny $\uparrow$4.71}} & \shortstack{76.32\\\gd{\tiny $\uparrow$6.12}} & \shortstack{\textbf{0.883}\\\gd{\tiny $\uparrow$0.009}} & \shortstack{87.85\\\gd{\tiny $\uparrow$6.83}} & \shortstack{9.44\\\gd{\tiny $\uparrow$0.88}} & \shortstack{\textbf{0.838}\\\gd{\tiny $\uparrow$0.008}} & \shortstack{88.87\\\gd{\tiny $\uparrow$6.40}} & \shortstack{22.04\\\gd{\tiny $\uparrow$1.81}} \\
\rowcolor{rowA}
500 & \shortstack{0.733\\\gd{\tiny $\uparrow$0.005}} & \shortstack{90.96\\\gd{\tiny $\uparrow$3.72}} & \shortstack{2.32\\\gd{\tiny $\uparrow$0.00}} & \shortstack{0.874\\\gd{\tiny $\uparrow$0.011}} & \shortstack{90.96\\\gd{\tiny $\uparrow$3.72}} & \shortstack{4.65\\\gd{\tiny $\uparrow$0.01}} & \shortstack{0.734\\\rd{\tiny $\downarrow$0.004}} & \shortstack{90.96\\\gd{\tiny $\uparrow$3.72}} & \shortstack{19.42\\\gd{\tiny $\uparrow$0.11}} & \shortstack{\textbf{0.959}\\\gd{\tiny $\uparrow$0.008}} & \shortstack{93.87\\\gd{\tiny $\uparrow$3.80}} & \shortstack{82.13\\\gd{\tiny $\uparrow$0.31}} & \shortstack{0.852\\\rd{\tiny $\downarrow$0.022}} & \shortstack{90.96\\\gd{\tiny $\uparrow$3.72}} & \shortstack{10.20\\\gd{\tiny $\uparrow$0.12}} & \shortstack{0.830\\\gd{\tiny $\uparrow$0.000}} & \shortstack{91.54\\\gd{\tiny $\uparrow$3.74}} & \shortstack{23.74\\\gd{\tiny $\uparrow$0.11}} \\
700 & \shortstack{0.730\\\gd{\tiny $\uparrow$0.002}} & \shortstack{93.97\\\gd{\tiny $\uparrow$0.71}} & \shortstack{2.57\\\rd{\tiny $\downarrow$0.25}} & \shortstack{0.873\\\gd{\tiny $\uparrow$0.010}} & \shortstack{93.97\\\gd{\tiny $\uparrow$0.71}} & \shortstack{4.73\\\rd{\tiny $\downarrow$0.07}} & \shortstack{0.732\\\rd{\tiny $\downarrow$0.006}} & \shortstack{93.97\\\gd{\tiny $\uparrow$0.71}} & \shortstack{21.49\\\rd{\tiny $\downarrow$1.96}} & \shortstack{0.949\\\rd{\tiny $\downarrow$0.002}} & \shortstack{95.20\\\gd{\tiny $\uparrow$2.47}} & \shortstack{87.09\\\rd{\tiny $\downarrow$4.65}} & \shortstack{0.872\\\rd{\tiny $\downarrow$0.002}} & \shortstack{93.97\\\gd{\tiny $\uparrow$0.71}} & \shortstack{11.28\\\rd{\tiny $\downarrow$0.96}} & \shortstack{0.831\\\gd{\tiny $\uparrow$0.000}} & \shortstack{94.21\\\gd{\tiny $\uparrow$1.07}} & \shortstack{25.43\\\rd{\tiny $\downarrow$1.58}} \\
\rowcolor{rowA}
1000 & \shortstack{0.754\\\gd{\tiny $\uparrow$0.026}} & \shortstack{97.01\\\rd{\tiny $\downarrow$2.33}} & \shortstack{2.85\\\rd{\tiny $\downarrow$0.53}} & \shortstack{0.867\\\gd{\tiny $\uparrow$0.004}} & \shortstack{95.01\\\rd{\tiny $\downarrow$0.33}} & \shortstack{4.96\\\rd{\tiny $\downarrow$0.30}} & \shortstack{0.739\\\gd{\tiny $\uparrow$0.001}} & \shortstack{95.01\\\rd{\tiny $\downarrow$0.33}} & \shortstack{23.95\\\rd{\tiny $\downarrow$4.42}} & \shortstack{0.949\\\rd{\tiny $\downarrow$0.002}} & \shortstack{99.06\\\rd{\tiny $\downarrow$1.39}} & \shortstack{91.49\\\rd{\tiny $\downarrow$9.05}} & \shortstack{0.852\\\rd{\tiny $\downarrow$0.022}} & \shortstack{97.01\\\rd{\tiny $\downarrow$2.33}} & \shortstack{12.57\\\rd{\tiny $\downarrow$2.25}} & \shortstack{0.832\\\gd{\tiny $\uparrow$0.002}} & \shortstack{96.62\\\rd{\tiny $\downarrow$1.34}} & \shortstack{27.16\\\rd{\tiny $\downarrow$3.31}} \\
\bottomrule
\end{tabular}%
}
\end{table*}

\begin{table*}[ht!]
\centering\scriptsize
\setlength{\tabcolsep}{3pt}
\renewcommand{\arraystretch}{1.1}
\caption{
 Full fientuning Accuracy, total GPU memory~(GB), and training time~(min) across five visual QA reasoning
  benchmarks (LLaVA-OneVision-7B) vs.\ number of trainable tokens. Subscript deltas are absolute deviations from
  \textit{All}: \gd{$\uparrow$}~green = improvement,\ \rd{$\downarrow$}~red = degradation.
  \textbf{Bold} = best value per column.
}
\label{tab:vqa_results}
\resizebox{\textwidth}{!}{%
\begin{tabular}{@{}lrrrrrrrrrrrrrrrrrr@{}}
\toprule
\multirow{2}{*}{\textbf{Tokens}} & \multicolumn{3}{c}{\textbf{M3CoT}} & \multicolumn{3}{c}{\textbf{MMK12}} & \multicolumn{3}{c}{\textbf{SafeTag}} & \multicolumn{3}{c}{\textbf{STVQA}} & \multicolumn{3}{c}{\textbf{VQA-R}} & \multicolumn{3}{c}{\textbf{Avg}} \\
\cmidrule(lr){2-4}\cmidrule(lr){5-7}\cmidrule(lr){8-10}\cmidrule(lr){11-13}\cmidrule(lr){14-16}\cmidrule(lr){17-19}
 & \textit{Acc} & \textit{GPU} & \textit{Time} & \textit{Acc} & \textit{GPU} & \textit{Time} & \textit{Acc} & \textit{GPU} & \textit{Time} & \textit{Acc} & \textit{GPU} & \textit{Time} & \textit{Acc} & \textit{GPU} & \textit{Time} & \textit{Acc} & \textit{GPU} & \textit{Time} \\
\midrule
\rowcolor{basecol}
\textit{All} & 0.753 & 97.08 & 38.94 & 0.400 & 96.42 & 7.15 & 0.894 & 97.09 & 11.29 & 0.844 & 97.08 & 29.11 & 0.577 & 89.02 & 7.30 & 0.693 & 95.34 & 18.76 \\
10 & \shortstack{0.505\\\rd{\tiny $\downarrow$0.249}} & \shortstack{\textbf{85.42}\\\gd{\tiny $\uparrow$11.66}} & \shortstack{31.45\\\gd{\tiny $\uparrow$7.49}} & \shortstack{0.383\\\rd{\tiny $\downarrow$0.018}} & \shortstack{\textbf{84.94}\\\gd{\tiny $\uparrow$11.48}} & \shortstack{5.75\\\gd{\tiny $\uparrow$1.40}} & \shortstack{0.892\\\rd{\tiny $\downarrow$0.001}} & \shortstack{\textbf{85.41}\\\gd{\tiny $\uparrow$11.68}} & \shortstack{9.33\\\gd{\tiny $\uparrow$1.96}} & \shortstack{0.737\\\rd{\tiny $\downarrow$0.106}} & \shortstack{\textbf{85.40}\\\gd{\tiny $\uparrow$11.67}} & \shortstack{24.02\\\gd{\tiny $\uparrow$5.09}} & \shortstack{0.573\\\rd{\tiny $\downarrow$0.003}} & \shortstack{\textbf{79.76}\\\gd{\tiny $\uparrow$9.26}} & \shortstack{\textbf{6.26}\\\gd{\tiny $\uparrow$1.04}} & \shortstack{0.618\\\rd{\tiny $\downarrow$0.075}} & \shortstack{\textbf{84.19}\\\gd{\tiny $\uparrow$11.15}} & \shortstack{\textbf{15.36}\\\gd{\tiny $\uparrow$3.40}} \\
\rowcolor{rowA}
25 & \shortstack{0.502\\\rd{\tiny $\downarrow$0.251}} & \shortstack{85.55\\\gd{\tiny $\uparrow$11.54}} & \shortstack{32.85\\\gd{\tiny $\uparrow$6.09}} & \shortstack{0.394\\\rd{\tiny $\downarrow$0.007}} & \shortstack{85.07\\\gd{\tiny $\uparrow$11.35}} & \shortstack{5.64\\\gd{\tiny $\uparrow$1.51}} & \shortstack{0.909\\\gd{\tiny $\uparrow$0.015}} & \shortstack{85.54\\\gd{\tiny $\uparrow$11.55}} & \shortstack{9.22\\\gd{\tiny $\uparrow$2.07}} & \shortstack{0.821\\\rd{\tiny $\downarrow$0.023}} & \shortstack{85.54\\\gd{\tiny $\uparrow$11.53}} & \shortstack{\textbf{23.98}\\\gd{\tiny $\uparrow$5.13}} & \shortstack{0.574\\\rd{\tiny $\downarrow$0.002}} & \shortstack{79.90\\\gd{\tiny $\uparrow$9.12}} & \shortstack{6.31\\\gd{\tiny $\uparrow$0.99}} & \shortstack{0.640\\\rd{\tiny $\downarrow$0.054}} & \shortstack{84.32\\\gd{\tiny $\uparrow$11.02}} & \shortstack{15.60\\\gd{\tiny $\uparrow$3.16}} \\
50 & \shortstack{0.486\\\rd{\tiny $\downarrow$0.267}} & \shortstack{85.81\\\gd{\tiny $\uparrow$11.27}} & \shortstack{\textbf{30.65}\\\gd{\tiny $\uparrow$8.29}} & \shortstack{0.372\\\rd{\tiny $\downarrow$0.028}} & \shortstack{85.31\\\gd{\tiny $\uparrow$11.11}} & \shortstack{\textbf{5.55}\\\gd{\tiny $\uparrow$1.60}} & \shortstack{0.907\\\gd{\tiny $\uparrow$0.014}} & \shortstack{85.80\\\gd{\tiny $\uparrow$11.29}} & \shortstack{\textbf{9.21}\\\gd{\tiny $\uparrow$2.08}} & \shortstack{0.851\\\gd{\tiny $\uparrow$0.007}} & \shortstack{85.79\\\gd{\tiny $\uparrow$11.28}} & \shortstack{26.71\\\gd{\tiny $\uparrow$2.40}} & \shortstack{\textbf{0.588}\\\gd{\tiny $\uparrow$0.011}} & \shortstack{80.11\\\gd{\tiny $\uparrow$8.91}} & \shortstack{6.30\\\gd{\tiny $\uparrow$1.00}} & \shortstack{0.641\\\rd{\tiny $\downarrow$0.053}} & \shortstack{84.56\\\gd{\tiny $\uparrow$10.77}} & \shortstack{15.68\\\gd{\tiny $\uparrow$3.07}} \\
\rowcolor{rowA}
100 & \shortstack{0.539\\\rd{\tiny $\downarrow$0.215}} & \shortstack{86.28\\\gd{\tiny $\uparrow$10.80}} & \shortstack{33.93\\\gd{\tiny $\uparrow$5.01}} & \shortstack{0.405\\\gd{\tiny $\uparrow$0.005}} & \shortstack{85.82\\\gd{\tiny $\uparrow$10.61}} & \shortstack{5.64\\\gd{\tiny $\uparrow$1.51}} & \shortstack{0.916\\\gd{\tiny $\uparrow$0.023}} & \shortstack{86.27\\\gd{\tiny $\uparrow$10.81}} & \shortstack{9.26\\\gd{\tiny $\uparrow$2.03}} & \shortstack{\textbf{0.896}\\\gd{\tiny $\uparrow$0.052}} & \shortstack{86.28\\\gd{\tiny $\uparrow$10.80}} & \shortstack{24.82\\\gd{\tiny $\uparrow$4.29}} & \shortstack{0.587\\\gd{\tiny $\uparrow$0.010}} & \shortstack{80.63\\\gd{\tiny $\uparrow$8.39}} & \shortstack{6.34\\\gd{\tiny $\uparrow$0.96}} & \shortstack{0.669\\\rd{\tiny $\downarrow$0.025}} & \shortstack{85.06\\\gd{\tiny $\uparrow$10.28}} & \shortstack{16.00\\\gd{\tiny $\uparrow$2.76}} \\
200 & \shortstack{0.595\\\rd{\tiny $\downarrow$0.158}} & \shortstack{87.21\\\gd{\tiny $\uparrow$9.88}} & \shortstack{31.59\\\gd{\tiny $\uparrow$7.35}} & \shortstack{\textbf{0.422}\\\gd{\tiny $\uparrow$0.022}} & \shortstack{86.71\\\gd{\tiny $\uparrow$9.72}} & \shortstack{5.78\\\gd{\tiny $\uparrow$1.37}} & \shortstack{0.906\\\gd{\tiny $\uparrow$0.012}} & \shortstack{87.17\\\gd{\tiny $\uparrow$9.92}} & \shortstack{9.86\\\gd{\tiny $\uparrow$1.43}} & \shortstack{0.893\\\gd{\tiny $\uparrow$0.049}} & \shortstack{87.18\\\gd{\tiny $\uparrow$9.90}} & \shortstack{26.51\\\gd{\tiny $\uparrow$2.60}} & \shortstack{0.587\\\gd{\tiny $\uparrow$0.010}} & \shortstack{81.51\\\gd{\tiny $\uparrow$7.51}} & \shortstack{6.58\\\gd{\tiny $\uparrow$0.72}} & \shortstack{0.681\\\rd{\tiny $\downarrow$0.013}} & \shortstack{85.96\\\gd{\tiny $\uparrow$9.38}} & \shortstack{16.06\\\gd{\tiny $\uparrow$2.69}} \\
\rowcolor{rowA}
300 & \shortstack{0.621\\\rd{\tiny $\downarrow$0.132}} & \shortstack{88.04\\\gd{\tiny $\uparrow$9.04}} & \shortstack{35.14\\\gd{\tiny $\uparrow$3.80}} & \shortstack{0.415\\\gd{\tiny $\uparrow$0.015}} & \shortstack{87.55\\\gd{\tiny $\uparrow$8.87}} & \shortstack{5.92\\\gd{\tiny $\uparrow$1.23}} & \shortstack{0.904\\\gd{\tiny $\uparrow$0.011}} & \shortstack{88.01\\\gd{\tiny $\uparrow$9.08}} & \shortstack{10.08\\\gd{\tiny $\uparrow$1.21}} & \shortstack{0.869\\\gd{\tiny $\uparrow$0.025}} & \shortstack{88.02\\\gd{\tiny $\uparrow$9.05}} & \shortstack{25.70\\\gd{\tiny $\uparrow$3.41}} & \shortstack{0.581\\\gd{\tiny $\uparrow$0.004}} & \shortstack{82.37\\\gd{\tiny $\uparrow$6.65}} & \shortstack{6.74\\\gd{\tiny $\uparrow$0.56}} & \shortstack{0.678\\\rd{\tiny $\downarrow$0.015}} & \shortstack{86.80\\\gd{\tiny $\uparrow$8.54}} & \shortstack{16.72\\\gd{\tiny $\uparrow$2.04}} \\
500 & \shortstack{0.727\\\rd{\tiny $\downarrow$0.026}} & \shortstack{89.86\\\gd{\tiny $\uparrow$7.22}} & \shortstack{33.96\\\gd{\tiny $\uparrow$4.98}} & \shortstack{0.390\\\rd{\tiny $\downarrow$0.010}} & \shortstack{89.35\\\gd{\tiny $\uparrow$7.08}} & \shortstack{6.17\\\gd{\tiny $\uparrow$0.98}} & \shortstack{\textbf{0.918}\\\gd{\tiny $\uparrow$0.024}} & \shortstack{89.84\\\gd{\tiny $\uparrow$7.25}} & \shortstack{10.22\\\gd{\tiny $\uparrow$1.07}} & \shortstack{0.841\\\rd{\tiny $\downarrow$0.003}} & \shortstack{89.85\\\gd{\tiny $\uparrow$7.22}} & \shortstack{27.61\\\gd{\tiny $\uparrow$1.50}} & \shortstack{0.585\\\gd{\tiny $\uparrow$0.009}} & \shortstack{84.19\\\gd{\tiny $\uparrow$4.84}} & \shortstack{6.97\\\gd{\tiny $\uparrow$0.33}} & \shortstack{0.692\\\rd{\tiny $\downarrow$0.001}} & \shortstack{88.62\\\gd{\tiny $\uparrow$6.72}} & \shortstack{16.99\\\gd{\tiny $\uparrow$1.77}} \\
\rowcolor{rowA}
700 & \shortstack{0.742\\\rd{\tiny $\downarrow$0.012}} & \shortstack{91.65\\\gd{\tiny $\uparrow$5.43}} & \shortstack{37.53\\\gd{\tiny $\uparrow$1.41}} & \shortstack{0.407\\\gd{\tiny $\uparrow$0.007}} & \shortstack{91.19\\\gd{\tiny $\uparrow$5.24}} & \shortstack{6.52\\\gd{\tiny $\uparrow$0.63}} & \shortstack{0.910\\\gd{\tiny $\uparrow$0.017}} & \shortstack{91.67\\\gd{\tiny $\uparrow$5.42}} & \shortstack{10.74\\\gd{\tiny $\uparrow$0.55}} & \shortstack{0.854\\\gd{\tiny $\uparrow$0.010}} & \shortstack{91.65\\\gd{\tiny $\uparrow$5.43}} & \shortstack{27.93\\\gd{\tiny $\uparrow$1.18}} & \shortstack{0.585\\\gd{\tiny $\uparrow$0.009}} & \shortstack{86.01\\\gd{\tiny $\uparrow$3.01}} & \shortstack{7.19\\\gd{\tiny $\uparrow$0.11}} & \shortstack{0.700\\\gd{\tiny $\uparrow$0.006}} & \shortstack{90.43\\\gd{\tiny $\uparrow$4.91}} & \shortstack{17.98\\\gd{\tiny $\uparrow$0.78}} \\
1000 & \shortstack{\textbf{0.768}\\\gd{\tiny $\uparrow$0.015}} & \shortstack{93.36\\\gd{\tiny $\uparrow$3.72}} & \shortstack{37.32\\\gd{\tiny $\uparrow$1.62}} & \shortstack{0.420\\\gd{\tiny $\uparrow$0.020}} & \shortstack{92.86\\\gd{\tiny $\uparrow$3.56}} & \shortstack{6.93\\\gd{\tiny $\uparrow$0.22}} & \shortstack{0.909\\\gd{\tiny $\uparrow$0.015}} & \shortstack{93.38\\\gd{\tiny $\uparrow$3.71}} & \shortstack{11.12\\\gd{\tiny $\uparrow$0.17}} & \shortstack{0.830\\\rd{\tiny $\downarrow$0.014}} & \shortstack{93.37\\\gd{\tiny $\uparrow$3.70}} & \shortstack{28.39\\\gd{\tiny $\uparrow$0.72}} & \shortstack{0.581\\\gd{\tiny $\uparrow$0.004}} & \shortstack{87.67\\\gd{\tiny $\uparrow$1.35}} & \shortstack{7.79\\\rd{\tiny $\downarrow$0.49}} & \shortstack{\textbf{0.702}\\\gd{\tiny $\uparrow$0.008}} & \shortstack{92.13\\\gd{\tiny $\uparrow$3.21}} & \shortstack{18.31\\\gd{\tiny $\uparrow$0.45}} \\
\rowcolor{rowA}
1500 & \shortstack{0.762\\\gd{\tiny $\uparrow$0.009}} & \shortstack{96.57\\\gd{\tiny $\uparrow$0.51}} & \shortstack{38.03\\\gd{\tiny $\uparrow$0.91}} & \shortstack{0.403\\\gd{\tiny $\uparrow$0.003}} & \shortstack{95.12\\\gd{\tiny $\uparrow$1.30}} & \shortstack{7.82\\\rd{\tiny $\downarrow$0.67}} & \shortstack{0.900\\\gd{\tiny $\uparrow$0.006}} & \shortstack{96.59\\\gd{\tiny $\uparrow$0.50}} & \shortstack{11.41\\\rd{\tiny $\downarrow$0.12}} & \shortstack{0.843\\\rd{\tiny $\downarrow$0.001}} & \shortstack{96.56\\\gd{\tiny $\uparrow$0.51}} & \shortstack{32.44\\\rd{\tiny $\downarrow$3.33}} & \shortstack{0.578\\\gd{\tiny $\uparrow$0.001}} & \shortstack{87.67\\\gd{\tiny $\uparrow$1.35}} & \shortstack{7.99\\\rd{\tiny $\downarrow$0.69}} & \shortstack{0.697\\\gd{\tiny $\uparrow$0.004}} & \shortstack{94.50\\\gd{\tiny $\uparrow$0.83}} & \shortstack{19.54\\\rd{\tiny $\downarrow$0.78}} \\
\bottomrule
\end{tabular}%
}
\end{table*}

\begin{table*}[ht!]
\centering\scriptsize
\setlength{\tabcolsep}{3pt}
\renewcommand{\arraystretch}{1.1}
\caption{
  Accuracy, total GPU memory~(GB), and training time~(min) across five visual QA reasoning
  benchmarks (Full fine-tuning, Qwen3-VL-8B) vs.\ number of trainable tokens.
  Subscript deltas are absolute deviations from \textit{All}:
  \gd{$\uparrow$}~green = improvement,\ \rd{$\downarrow$}~red = degradation.
  \textbf{Bold} = best value per column. GPU is the sum of all GPU peak allocations.
}
\label{tab:qwen_full_results}
\resizebox{\textwidth}{!}{%
\begin{tabular}{@{}lrrrrrrrrrrrrrrrrrr@{}}
\toprule
\multirow{2}{*}{\textbf{Tokens}} & \multicolumn{3}{c}{\textbf{M3CoT}} & \multicolumn{3}{c}{\textbf{MMK12}} & \multicolumn{3}{c}{\textbf{SafeTag}} & \multicolumn{3}{c}{\textbf{STVQA}} & \multicolumn{3}{c}{\textbf{VQA-R}} & \multicolumn{3}{c}{\textbf{Avg}} \\
\cmidrule(lr){2-4}\cmidrule(lr){5-7}\cmidrule(lr){8-10}\cmidrule(lr){11-13}\cmidrule(lr){14-16}\cmidrule(lr){17-19}
 & \textit{Acc} & \textit{GPU} & \textit{Time} & \textit{Acc} & \textit{GPU} & \textit{Time} & \textit{Acc} & \textit{GPU} & \textit{Time} & \textit{Acc} & \textit{GPU} & \textit{Time} & \textit{Acc} & \textit{GPU} & \textit{Time} & \textit{Acc} & \textit{GPU} & \textit{Time} \\
\midrule
\rowcolor{basecol}
\textit{All} & 0.782 & 207.54 & 34.18 & 0.568 & 188.12 & 4.13 & 0.918 & 102.76 & 5.94 & 0.852 & 148.69 & 13.69 & 0.519 & 193.25 & 8.52 & 0.728 & 168.07 & 13.29 \\
10 & \shortstack{0.769\\\rd{\tiny $\downarrow$0.012}} & \shortstack{\textbf{161.61}\\\gd{\tiny $\uparrow$45.93}} & \shortstack{29.74\\\gd{\tiny $\uparrow$4.44}} & \shortstack{0.548\\\rd{\tiny $\downarrow$0.019}} & \shortstack{\textbf{148.10}\\\gd{\tiny $\uparrow$40.03}} & \shortstack{\textbf{3.70}\\\gd{\tiny $\uparrow$0.43}} & \shortstack{0.895\\\rd{\tiny $\downarrow$0.023}} & \shortstack{\textbf{90.34}\\\gd{\tiny $\uparrow$12.42}} & \shortstack{\textbf{5.16}\\\gd{\tiny $\uparrow$0.78}} & \shortstack{0.814\\\rd{\tiny $\downarrow$0.038}} & \shortstack{\textbf{121.51}\\\gd{\tiny $\uparrow$27.18}} & \shortstack{13.13\\\gd{\tiny $\uparrow$0.56}} & \shortstack{0.528\\\gd{\tiny $\uparrow$0.009}} & \shortstack{\textbf{154.88}\\\gd{\tiny $\uparrow$38.37}} & \shortstack{\textbf{6.72}\\\gd{\tiny $\uparrow$1.80}} & \shortstack{0.711\\\rd{\tiny $\downarrow$0.017}} & \shortstack{\textbf{135.29}\\\gd{\tiny $\uparrow$32.78}} & \shortstack{11.69\\\gd{\tiny $\uparrow$1.60}} \\
\rowcolor{rowA}
25 & \shortstack{0.723\\\rd{\tiny $\downarrow$0.058}} & \shortstack{161.89\\\gd{\tiny $\uparrow$45.65}} & \shortstack{\textbf{28.70}\\\gd{\tiny $\uparrow$5.48}} & \shortstack{0.561\\\rd{\tiny $\downarrow$0.006}} & \shortstack{148.33\\\gd{\tiny $\uparrow$39.79}} & \shortstack{3.71\\\gd{\tiny $\uparrow$0.42}} & \shortstack{0.909\\\rd{\tiny $\downarrow$0.009}} & \shortstack{90.64\\\gd{\tiny $\uparrow$12.12}} & \shortstack{5.18\\\gd{\tiny $\uparrow$0.76}} & \shortstack{0.813\\\rd{\tiny $\downarrow$0.039}} & \shortstack{121.84\\\gd{\tiny $\uparrow$26.85}} & \shortstack{12.20\\\gd{\tiny $\uparrow$1.49}} & \shortstack{0.550\\\gd{\tiny $\uparrow$0.031}} & \shortstack{155.21\\\gd{\tiny $\uparrow$38.04}} & \shortstack{6.73\\\gd{\tiny $\uparrow$1.79}} & \shortstack{0.711\\\rd{\tiny $\downarrow$0.016}} & \shortstack{135.58\\\gd{\tiny $\uparrow$32.49}} & \shortstack{\textbf{11.30}\\\gd{\tiny $\uparrow$1.99}} \\
50 & \shortstack{0.725\\\rd{\tiny $\downarrow$0.057}} & \shortstack{162.34\\\gd{\tiny $\uparrow$45.20}} & \shortstack{29.37\\\gd{\tiny $\uparrow$4.81}} & \shortstack{0.555\\\rd{\tiny $\downarrow$0.013}} & \shortstack{148.77\\\gd{\tiny $\uparrow$39.35}} & \shortstack{3.73\\\gd{\tiny $\uparrow$0.40}} & \shortstack{0.918\\\rd{\tiny $\downarrow$0.000}} & \shortstack{91.14\\\gd{\tiny $\uparrow$11.62}} & \shortstack{5.20\\\gd{\tiny $\uparrow$0.74}} & \shortstack{0.857\\\gd{\tiny $\uparrow$0.005}} & \shortstack{122.37\\\gd{\tiny $\uparrow$26.31}} & \shortstack{\textbf{12.16}\\\gd{\tiny $\uparrow$1.53}} & \shortstack{0.545\\\gd{\tiny $\uparrow$0.027}} & \shortstack{155.67\\\gd{\tiny $\uparrow$37.58}} & \shortstack{6.75\\\gd{\tiny $\uparrow$1.77}} & \shortstack{0.720\\\rd{\tiny $\downarrow$0.008}} & \shortstack{136.06\\\gd{\tiny $\uparrow$32.01}} & \shortstack{11.44\\\gd{\tiny $\uparrow$1.85}} \\
\rowcolor{rowA}
100 & \shortstack{0.752\\\rd{\tiny $\downarrow$0.030}} & \shortstack{163.08\\\gd{\tiny $\uparrow$44.46}} & \shortstack{29.42\\\gd{\tiny $\uparrow$4.76}} & \shortstack{0.574\\\gd{\tiny $\uparrow$0.007}} & \shortstack{149.48\\\gd{\tiny $\uparrow$38.64}} & \shortstack{3.81\\\gd{\tiny $\uparrow$0.32}} & \shortstack{0.910\\\rd{\tiny $\downarrow$0.008}} & \shortstack{91.81\\\gd{\tiny $\uparrow$10.95}} & \shortstack{5.36\\\gd{\tiny $\uparrow$0.58}} & \shortstack{0.887\\\gd{\tiny $\uparrow$0.036}} & \shortstack{123.41\\\gd{\tiny $\uparrow$25.28}} & \shortstack{12.53\\\gd{\tiny $\uparrow$1.16}} & \shortstack{\textbf{0.554}\\\gd{\tiny $\uparrow$0.035}} & \shortstack{156.34\\\gd{\tiny $\uparrow$36.91}} & \shortstack{6.89\\\gd{\tiny $\uparrow$1.63}} & \shortstack{0.736\\\gd{\tiny $\uparrow$0.008}} & \shortstack{136.82\\\gd{\tiny $\uparrow$31.25}} & \shortstack{11.60\\\gd{\tiny $\uparrow$1.69}} \\
200 & \shortstack{0.775\\\rd{\tiny $\downarrow$0.007}} & \shortstack{164.61\\\gd{\tiny $\uparrow$42.93}} & \shortstack{30.54\\\gd{\tiny $\uparrow$3.64}} & \shortstack{\textbf{0.588}\\\gd{\tiny $\uparrow$0.020}} & \shortstack{151.01\\\gd{\tiny $\uparrow$37.11}} & \shortstack{3.96\\\gd{\tiny $\uparrow$0.17}} & \shortstack{0.921\\\gd{\tiny $\uparrow$0.003}} & \shortstack{93.39\\\gd{\tiny $\uparrow$9.37}} & \shortstack{5.46\\\gd{\tiny $\uparrow$0.48}} & \shortstack{\textbf{0.904}\\\gd{\tiny $\uparrow$0.052}} & \shortstack{125.46\\\gd{\tiny $\uparrow$23.23}} & \shortstack{14.64\\\rd{\tiny $\downarrow$0.95}} & \shortstack{0.530\\\gd{\tiny $\uparrow$0.011}} & \shortstack{157.91\\\gd{\tiny $\uparrow$35.34}} & \shortstack{7.08\\\gd{\tiny $\uparrow$1.44}} & \shortstack{\textbf{0.743}\\\gd{\tiny $\uparrow$0.016}} & \shortstack{138.48\\\gd{\tiny $\uparrow$29.59}} & \shortstack{12.34\\\gd{\tiny $\uparrow$0.96}} \\
\rowcolor{rowA}
300 & \shortstack{0.783\\\gd{\tiny $\uparrow$0.001}} & \shortstack{166.14\\\gd{\tiny $\uparrow$41.39}} & \shortstack{31.35\\\gd{\tiny $\uparrow$2.83}} & \shortstack{0.573\\\gd{\tiny $\uparrow$0.005}} & \shortstack{152.61\\\gd{\tiny $\uparrow$35.51}} & \shortstack{4.00\\\gd{\tiny $\uparrow$0.13}} & \shortstack{\textbf{0.929}\\\gd{\tiny $\uparrow$0.011}} & \shortstack{94.99\\\gd{\tiny $\uparrow$7.77}} & \shortstack{5.75\\\gd{\tiny $\uparrow$0.19}} & \shortstack{0.890\\\gd{\tiny $\uparrow$0.038}} & \shortstack{127.50\\\gd{\tiny $\uparrow$21.19}} & \shortstack{14.12\\\rd{\tiny $\downarrow$0.43}} & \shortstack{0.519\\\rd{\tiny $\downarrow$0.000}} & \shortstack{159.41\\\gd{\tiny $\uparrow$33.84}} & \shortstack{7.20\\\gd{\tiny $\uparrow$1.32}} & \shortstack{0.739\\\gd{\tiny $\uparrow$0.011}} & \shortstack{140.13\\\gd{\tiny $\uparrow$27.94}} & \shortstack{12.48\\\gd{\tiny $\uparrow$0.81}} \\
500 & \shortstack{0.771\\\rd{\tiny $\downarrow$0.010}} & \shortstack{169.22\\\gd{\tiny $\uparrow$38.32}} & \shortstack{34.12\\\gd{\tiny $\uparrow$0.06}} & \shortstack{0.575\\\gd{\tiny $\uparrow$0.007}} & \shortstack{155.70\\\gd{\tiny $\uparrow$32.42}} & \shortstack{4.00\\\gd{\tiny $\uparrow$0.13}} & \shortstack{0.919\\\gd{\tiny $\uparrow$0.002}} & \shortstack{98.12\\\gd{\tiny $\uparrow$4.64}} & \shortstack{5.89\\\gd{\tiny $\uparrow$0.05}} & \shortstack{0.881\\\gd{\tiny $\uparrow$0.029}} & \shortstack{131.69\\\gd{\tiny $\uparrow$17.00}} & \shortstack{14.19\\\rd{\tiny $\downarrow$0.50}} & \shortstack{0.530\\\gd{\tiny $\uparrow$0.011}} & \shortstack{162.53\\\gd{\tiny $\uparrow$30.72}} & \shortstack{7.78\\\gd{\tiny $\uparrow$0.74}} & \shortstack{0.735\\\gd{\tiny $\uparrow$0.008}} & \shortstack{143.45\\\gd{\tiny $\uparrow$24.62}} & \shortstack{13.20\\\gd{\tiny $\uparrow$0.10}} \\
\rowcolor{rowA}
700 & \shortstack{\textbf{0.789}\\\gd{\tiny $\uparrow$0.008}} & \shortstack{172.38\\\gd{\tiny $\uparrow$35.15}} & \shortstack{34.49\\\rd{\tiny $\downarrow$0.31}} & \shortstack{0.574\\\gd{\tiny $\uparrow$0.007}} & \shortstack{158.75\\\gd{\tiny $\uparrow$29.37}} & \shortstack{4.10\\\gd{\tiny $\uparrow$0.03}} & \shortstack{\textbf{0.929}\\\gd{\tiny $\uparrow$0.011}} & \shortstack{101.13\\\gd{\tiny $\uparrow$1.63}} & \shortstack{5.51\\\gd{\tiny $\uparrow$0.43}} & \shortstack{0.837\\\rd{\tiny $\downarrow$0.015}} & \shortstack{135.73\\\gd{\tiny $\uparrow$12.96}} & \shortstack{15.23\\\rd{\tiny $\downarrow$1.54}} & \shortstack{0.525\\\gd{\tiny $\uparrow$0.007}} & \shortstack{165.59\\\gd{\tiny $\uparrow$27.65}} & \shortstack{8.44\\\gd{\tiny $\uparrow$0.08}} & \shortstack{0.731\\\gd{\tiny $\uparrow$0.003}} & \shortstack{146.72\\\gd{\tiny $\uparrow$21.35}} & \shortstack{13.55\\\rd{\tiny $\downarrow$0.26}} \\
1000 & \shortstack{0.779\\\rd{\tiny $\downarrow$0.003}} & \shortstack{176.98\\\gd{\tiny $\uparrow$30.55}} & \shortstack{35.47\\\rd{\tiny $\downarrow$1.29}} & \shortstack{0.570\\\gd{\tiny $\uparrow$0.002}} & \shortstack{163.45\\\gd{\tiny $\uparrow$24.67}} & \shortstack{4.13\\\rd{\tiny $\downarrow$0.00}} & \shortstack{0.921\\\gd{\tiny $\uparrow$0.003}} & \shortstack{105.83\\\rd{\tiny $\downarrow$3.07}} & \shortstack{5.93\\\gd{\tiny $\uparrow$0.01}} & \shortstack{0.835\\\rd{\tiny $\downarrow$0.017}} & \shortstack{141.91\\\gd{\tiny $\uparrow$6.78}} & \shortstack{14.63\\\rd{\tiny $\downarrow$0.94}} & \shortstack{0.514\\\rd{\tiny $\downarrow$0.004}} & \shortstack{170.32\\\gd{\tiny $\uparrow$22.93}} & \shortstack{8.24\\\gd{\tiny $\uparrow$0.28}} & \shortstack{0.724\\\rd{\tiny $\downarrow$0.004}} & \shortstack{151.70\\\gd{\tiny $\uparrow$16.37}} & \shortstack{13.68\\\rd{\tiny $\downarrow$0.39}} \\
\rowcolor{rowA}
1500 & \shortstack{0.788\\\gd{\tiny $\uparrow$0.006}} & \shortstack{184.67\\\gd{\tiny $\uparrow$22.87}} & \shortstack{35.87\\\rd{\tiny $\downarrow$1.69}} & \shortstack{0.570\\\gd{\tiny $\uparrow$0.002}} & \shortstack{171.06\\\gd{\tiny $\uparrow$17.06}} & \shortstack{4.14\\\rd{\tiny $\downarrow$0.01}} & \shortstack{0.918\\\rd{\tiny $\downarrow$0.000}} & \shortstack{102.76\\\gd{\tiny $\uparrow$0.00}} & \shortstack{5.87\\\gd{\tiny $\uparrow$0.07}} & \shortstack{0.840\\\rd{\tiny $\downarrow$0.011}} & \shortstack{151.61\\\rd{\tiny $\downarrow$2.92}} & \shortstack{13.92\\\rd{\tiny $\downarrow$0.23}} & \shortstack{0.532\\\gd{\tiny $\uparrow$0.013}} & \shortstack{177.94\\\gd{\tiny $\uparrow$15.31}} & \shortstack{7.73\\\gd{\tiny $\uparrow$0.79}} & \shortstack{0.730\\\gd{\tiny $\uparrow$0.002}} & \shortstack{157.61\\\gd{\tiny $\uparrow$10.46}} & \shortstack{13.51\\\rd{\tiny $\downarrow$0.21}} \\
\bottomrule
\end{tabular}%
}
\end{table*}

\begin{table*}[ht!]
\centering\scriptsize
\setlength{\tabcolsep}{3pt}
\renewcommand{\arraystretch}{1.1}
\caption{
  Accuracy, total GPU memory~(GB), and training time~(min) across five visual QA reasoning
  benchmarks (LoRA, Qwen3-VL-8B) vs.\ number of trainable tokens.
  Subscript deltas are absolute deviations from \textit{All}:
  \gd{$\uparrow$}~green = improvement,\ \rd{$\downarrow$}~red = degradation.
  \textbf{Bold} = best value per column. GPU is the sum of both GPU peak allocations.
}
\label{tab:qwen_lora_results}
\resizebox{\textwidth}{!}{%
\begin{tabular}{@{}lrrrrrrrrrrrrrrrrrr@{}}
\toprule
\multirow{2}{*}{\textbf{Tokens}} & \multicolumn{3}{c}{\textbf{M3CoT}} & \multicolumn{3}{c}{\textbf{MMK12}} & \multicolumn{3}{c}{\textbf{SafeTag}} & \multicolumn{3}{c}{\textbf{STVQA}} & \multicolumn{3}{c}{\textbf{VQA-R}} & \multicolumn{3}{c}{\textbf{Avg}} \\
\cmidrule(lr){2-4}\cmidrule(lr){5-7}\cmidrule(lr){8-10}\cmidrule(lr){11-13}\cmidrule(lr){14-16}\cmidrule(lr){17-19}
 & \textit{Acc} & \textit{GPU} & \textit{Time} & \textit{Acc} & \textit{GPU} & \textit{Time} & \textit{Acc} & \textit{GPU} & \textit{Time} & \textit{Acc} & \textit{GPU} & \textit{Time} & \textit{Acc} & \textit{GPU} & \textit{Time} & \textit{Acc} & \textit{GPU} & \textit{Time} \\
\midrule
\rowcolor{basecol}
\textit{All} & 0.676 & 93.26 & 28.32 & 0.544 & 89.63 & 3.91 & 0.813 & 101.60 & 4.09 & 0.810 & 76.14 & 12.15 & 0.503 & 79.99 & 6.28 & 0.669 & 88.12 & 10.95 \\
10 & \shortstack{0.669\\\rd{\tiny $\downarrow$0.007}} & \shortstack{\textbf{80.58}\\\gd{\tiny $\uparrow$12.68}} & \shortstack{25.16\\\gd{\tiny $\uparrow$3.16}} & \shortstack{0.550\\\gd{\tiny $\uparrow$0.006}} & \shortstack{\textbf{77.53}\\\gd{\tiny $\uparrow$12.10}} & \shortstack{\textbf{3.02}\\\gd{\tiny $\uparrow$0.89}} & \shortstack{0.807\\\rd{\tiny $\downarrow$0.006}} & \shortstack{\textbf{87.37}\\\gd{\tiny $\uparrow$14.23}} & \shortstack{\textbf{3.77}\\\gd{\tiny $\uparrow$0.32}} & \shortstack{0.807\\\rd{\tiny $\downarrow$0.002}} & \shortstack{\textbf{66.16}\\\gd{\tiny $\uparrow$9.97}} & \shortstack{\textbf{10.22}\\\gd{\tiny $\uparrow$1.93}} & \shortstack{0.479\\\rd{\tiny $\downarrow$0.024}} & \shortstack{\textbf{69.45}\\\gd{\tiny $\uparrow$10.54}} & \shortstack{5.21\\\gd{\tiny $\uparrow$1.07}} & \shortstack{0.663\\\rd{\tiny $\downarrow$0.007}} & \shortstack{\textbf{76.22}\\\gd{\tiny $\uparrow$11.91}} & \shortstack{9.48\\\gd{\tiny $\uparrow$1.47}} \\
\rowcolor{rowA}
25 & \shortstack{0.596\\\rd{\tiny $\downarrow$0.080}} & \shortstack{80.62\\\gd{\tiny $\uparrow$12.63}} & \shortstack{25.57\\\gd{\tiny $\uparrow$2.75}} & \shortstack{0.522\\\rd{\tiny $\downarrow$0.022}} & \shortstack{77.58\\\gd{\tiny $\uparrow$12.05}} & \shortstack{3.03\\\gd{\tiny $\uparrow$0.88}} & \shortstack{0.818\\\gd{\tiny $\uparrow$0.004}} & \shortstack{87.58\\\gd{\tiny $\uparrow$14.03}} & \shortstack{3.79\\\gd{\tiny $\uparrow$0.30}} & \shortstack{0.819\\\gd{\tiny $\uparrow$0.009}} & \shortstack{66.24\\\gd{\tiny $\uparrow$9.90}} & \shortstack{10.26\\\gd{\tiny $\uparrow$1.89}} & \shortstack{0.505\\\gd{\tiny $\uparrow$0.002}} & \shortstack{69.50\\\gd{\tiny $\uparrow$10.49}} & \shortstack{\textbf{5.18}\\\gd{\tiny $\uparrow$1.10}} & \shortstack{0.652\\\rd{\tiny $\downarrow$0.017}} & \shortstack{76.31\\\gd{\tiny $\uparrow$11.82}} & \shortstack{9.57\\\gd{\tiny $\uparrow$1.38}} \\
50 & \shortstack{0.677\\\gd{\tiny $\uparrow$0.001}} & \shortstack{80.71\\\gd{\tiny $\uparrow$12.55}} & \shortstack{\textbf{22.48}\\\gd{\tiny $\uparrow$5.84}} & \shortstack{0.515\\\rd{\tiny $\downarrow$0.029}} & \shortstack{77.66\\\gd{\tiny $\uparrow$11.97}} & \shortstack{3.04\\\gd{\tiny $\uparrow$0.87}} & \shortstack{0.816\\\gd{\tiny $\uparrow$0.003}} & \shortstack{87.97\\\gd{\tiny $\uparrow$13.63}} & \shortstack{3.81\\\gd{\tiny $\uparrow$0.28}} & \shortstack{0.804\\\rd{\tiny $\downarrow$0.006}} & \shortstack{66.37\\\gd{\tiny $\uparrow$9.77}} & \shortstack{11.36\\\gd{\tiny $\uparrow$0.79}} & \shortstack{\textbf{0.512}\\\gd{\tiny $\uparrow$0.009}} & \shortstack{69.59\\\gd{\tiny $\uparrow$10.41}} & \shortstack{5.21\\\gd{\tiny $\uparrow$1.07}} & \shortstack{0.665\\\rd{\tiny $\downarrow$0.004}} & \shortstack{76.46\\\gd{\tiny $\uparrow$11.67}} & \shortstack{9.18\\\gd{\tiny $\uparrow$1.77}} \\
\rowcolor{rowA}
100 & \shortstack{0.674\\\rd{\tiny $\downarrow$0.002}} & \shortstack{80.88\\\gd{\tiny $\uparrow$12.38}} & \shortstack{26.87\\\gd{\tiny $\uparrow$1.45}} & \shortstack{0.530\\\rd{\tiny $\downarrow$0.014}} & \shortstack{77.84\\\gd{\tiny $\uparrow$11.79}} & \shortstack{3.03\\\gd{\tiny $\uparrow$0.88}} & \shortstack{0.810\\\rd{\tiny $\downarrow$0.003}} & \shortstack{88.60\\\gd{\tiny $\uparrow$13.01}} & \shortstack{3.81\\\gd{\tiny $\uparrow$0.28}} & \shortstack{0.772\\\rd{\tiny $\downarrow$0.038}} & \shortstack{66.63\\\gd{\tiny $\uparrow$9.51}} & \shortstack{10.82\\\gd{\tiny $\uparrow$1.33}} & \shortstack{0.508\\\gd{\tiny $\uparrow$0.005}} & \shortstack{69.75\\\gd{\tiny $\uparrow$10.24}} & \shortstack{5.25\\\gd{\tiny $\uparrow$1.03}} & \shortstack{0.659\\\rd{\tiny $\downarrow$0.011}} & \shortstack{76.74\\\gd{\tiny $\uparrow$11.39}} & \shortstack{9.96\\\gd{\tiny $\uparrow$0.99}} \\
200 & \shortstack{0.626\\\rd{\tiny $\downarrow$0.050}} & \shortstack{81.22\\\gd{\tiny $\uparrow$12.03}} & \shortstack{22.72\\\gd{\tiny $\uparrow$5.60}} & \shortstack{0.546\\\gd{\tiny $\uparrow$0.002}} & \shortstack{78.19\\\gd{\tiny $\uparrow$11.44}} & \shortstack{3.07\\\gd{\tiny $\uparrow$0.84}} & \shortstack{0.804\\\rd{\tiny $\downarrow$0.009}} & \shortstack{89.90\\\gd{\tiny $\uparrow$11.71}} & \shortstack{3.88\\\gd{\tiny $\uparrow$0.21}} & \shortstack{0.810\\\gd{\tiny $\uparrow$0.000}} & \shortstack{67.12\\\gd{\tiny $\uparrow$9.02}} & \shortstack{10.53\\\gd{\tiny $\uparrow$1.62}} & \shortstack{0.503\\\rd{\tiny $\downarrow$0.000}} & \shortstack{70.08\\\gd{\tiny $\uparrow$9.91}} & \shortstack{5.28\\\gd{\tiny $\uparrow$1.00}} & \shortstack{0.658\\\rd{\tiny $\downarrow$0.011}} & \shortstack{77.30\\\gd{\tiny $\uparrow$10.82}} & \shortstack{\textbf{9.10}\\\gd{\tiny $\uparrow$1.85}} \\
\rowcolor{rowA}
300 & \shortstack{\textbf{0.696}\\\gd{\tiny $\uparrow$0.020}} & \shortstack{81.55\\\gd{\tiny $\uparrow$11.71}} & \shortstack{26.66\\\gd{\tiny $\uparrow$1.66}} & \shortstack{\textbf{0.559}\\\gd{\tiny $\uparrow$0.015}} & \shortstack{78.52\\\gd{\tiny $\uparrow$11.11}} & \shortstack{3.05\\\gd{\tiny $\uparrow$0.86}} & \shortstack{0.791\\\rd{\tiny $\downarrow$0.023}} & \shortstack{91.19\\\gd{\tiny $\uparrow$10.41}} & \shortstack{3.96\\\gd{\tiny $\uparrow$0.13}} & \shortstack{\textbf{0.825}\\\gd{\tiny $\uparrow$0.015}} & \shortstack{67.60\\\gd{\tiny $\uparrow$8.54}} & \shortstack{10.72\\\gd{\tiny $\uparrow$1.43}} & \shortstack{\textbf{0.512}\\\gd{\tiny $\uparrow$0.009}} & \shortstack{70.46\\\gd{\tiny $\uparrow$9.53}} & \shortstack{5.29\\\gd{\tiny $\uparrow$0.99}} & \shortstack{\textbf{0.676}\\\gd{\tiny $\uparrow$0.007}} & \shortstack{77.86\\\gd{\tiny $\uparrow$10.26}} & \shortstack{9.94\\\gd{\tiny $\uparrow$1.01}} \\
500 & \shortstack{0.684\\\gd{\tiny $\uparrow$0.008}} & \shortstack{82.20\\\gd{\tiny $\uparrow$11.06}} & \shortstack{24.48\\\gd{\tiny $\uparrow$3.84}} & \shortstack{0.552\\\gd{\tiny $\uparrow$0.007}} & \shortstack{79.15\\\gd{\tiny $\uparrow$10.48}} & \shortstack{3.06\\\gd{\tiny $\uparrow$0.85}} & \shortstack{0.819\\\gd{\tiny $\uparrow$0.005}} & \shortstack{93.85\\\gd{\tiny $\uparrow$7.75}} & \shortstack{4.06\\\gd{\tiny $\uparrow$0.03}} & \shortstack{0.818\\\gd{\tiny $\uparrow$0.008}} & \shortstack{68.62\\\gd{\tiny $\uparrow$7.51}} & \shortstack{12.11\\\gd{\tiny $\uparrow$0.04}} & \shortstack{0.503\\\rd{\tiny $\downarrow$0.000}} & \shortstack{71.08\\\gd{\tiny $\uparrow$8.91}} & \shortstack{5.33\\\gd{\tiny $\uparrow$0.95}} & \shortstack{0.675\\\gd{\tiny $\uparrow$0.006}} & \shortstack{78.98\\\gd{\tiny $\uparrow$9.14}} & \shortstack{9.81\\\gd{\tiny $\uparrow$1.14}} \\
\rowcolor{rowA}
700 & \shortstack{0.641\\\rd{\tiny $\downarrow$0.035}} & \shortstack{82.84\\\gd{\tiny $\uparrow$10.42}} & \shortstack{25.27\\\gd{\tiny $\uparrow$3.05}} & \shortstack{0.553\\\gd{\tiny $\uparrow$0.009}} & \shortstack{79.80\\\gd{\tiny $\uparrow$9.83}} & \shortstack{3.07\\\gd{\tiny $\uparrow$0.84}} & \shortstack{\textbf{0.825}\\\gd{\tiny $\uparrow$0.011}} & \shortstack{96.44\\\gd{\tiny $\uparrow$5.17}} & \shortstack{3.97\\\gd{\tiny $\uparrow$0.12}} & \shortstack{0.812\\\gd{\tiny $\uparrow$0.003}} & \shortstack{69.57\\\gd{\tiny $\uparrow$6.57}} & \shortstack{10.71\\\gd{\tiny $\uparrow$1.44}} & \shortstack{0.508\\\gd{\tiny $\uparrow$0.005}} & \shortstack{71.72\\\gd{\tiny $\uparrow$8.27}} & \shortstack{5.39\\\gd{\tiny $\uparrow$0.89}} & \shortstack{0.668\\\rd{\tiny $\downarrow$0.002}} & \shortstack{80.07\\\gd{\tiny $\uparrow$8.05}} & \shortstack{9.68\\\gd{\tiny $\uparrow$1.27}} \\
1000 & \shortstack{0.650\\\rd{\tiny $\downarrow$0.026}} & \shortstack{83.84\\\gd{\tiny $\uparrow$9.41}} & \shortstack{25.18\\\gd{\tiny $\uparrow$3.14}} & \shortstack{0.540\\\rd{\tiny $\downarrow$0.004}} & \shortstack{80.80\\\gd{\tiny $\uparrow$8.83}} & \shortstack{3.08\\\gd{\tiny $\uparrow$0.83}} & \shortstack{0.813\\\rd{\tiny $\downarrow$0.000}} & \shortstack{100.35\\\gd{\tiny $\uparrow$1.26}} & \shortstack{3.97\\\gd{\tiny $\uparrow$0.12}} & \shortstack{0.810\\\gd{\tiny $\uparrow$0.000}} & \shortstack{71.11\\\gd{\tiny $\uparrow$5.02}} & \shortstack{10.75\\\gd{\tiny $\uparrow$1.40}} & \shortstack{0.501\\\rd{\tiny $\downarrow$0.002}} & \shortstack{72.73\\\gd{\tiny $\uparrow$7.26}} & \shortstack{5.53\\\gd{\tiny $\uparrow$0.75}} & \shortstack{0.663\\\rd{\tiny $\downarrow$0.006}} & \shortstack{81.77\\\gd{\tiny $\uparrow$6.36}} & \shortstack{9.70\\\gd{\tiny $\uparrow$1.25}} \\
\rowcolor{rowA}
1500 & \shortstack{0.674\\\rd{\tiny $\downarrow$0.002}} & \shortstack{85.46\\\gd{\tiny $\uparrow$7.79}} & \shortstack{25.40\\\gd{\tiny $\uparrow$2.92}} & \shortstack{0.543\\\rd{\tiny $\downarrow$0.001}} & \shortstack{82.42\\\gd{\tiny $\uparrow$7.21}} & \shortstack{3.10\\\gd{\tiny $\uparrow$0.81}} & \shortstack{0.818\\\gd{\tiny $\uparrow$0.004}} & \shortstack{101.60\\\rd{\tiny $\downarrow$0.00}} & \shortstack{3.97\\\gd{\tiny $\uparrow$0.12}} & \shortstack{0.811\\\gd{\tiny $\uparrow$0.002}} & \shortstack{73.54\\\gd{\tiny $\uparrow$2.60}} & \shortstack{11.97\\\gd{\tiny $\uparrow$0.18}} & \shortstack{0.501\\\rd{\tiny $\downarrow$0.002}} & \shortstack{74.36\\\gd{\tiny $\uparrow$5.64}} & \shortstack{5.48\\\gd{\tiny $\uparrow$0.80}} & \shortstack{0.669\\\gd{\tiny $\uparrow$0.000}} & \shortstack{83.48\\\gd{\tiny $\uparrow$4.65}} & \shortstack{9.98\\\gd{\tiny $\uparrow$0.97}} \\
\bottomrule
\end{tabular}%
}
\end{table*}

\begin{table*}[ht!]
\centering\scriptsize
\setlength{\tabcolsep}{3pt}
\renewcommand{\arraystretch}{1.1}
\caption{
  Accuracy, GPU memory~(GB), and training time~(min) across five visual QA reasoning
  benchmarks (LoRA, LLaVA-OneVision-7B) vs.\ number of trainable tokens. Subscript deltas are absolute deviations from
  \textit{All}: \gd{$\uparrow$}~green = improvement,\ \rd{$\downarrow$}~red = degradation.
  \textbf{Bold} = best value per column.
}
\label{tab:vqa_lora_results}
\resizebox{\textwidth}{!}{%
\begin{tabular}{@{}lrrrrrrrrrrrrrrrrrr@{}}
\toprule
\multirow{2}{*}{\textbf{Tokens}} & \multicolumn{3}{c}{\textbf{M3CoT}} & \multicolumn{3}{c}{\textbf{MMK12}} & \multicolumn{3}{c}{\textbf{SafeTag}} & \multicolumn{3}{c}{\textbf{STVQA}} & \multicolumn{3}{c}{\textbf{VQA-R}} & \multicolumn{3}{c}{\textbf{Avg}} \\
\cmidrule(lr){2-4}\cmidrule(lr){5-7}\cmidrule(lr){8-10}\cmidrule(lr){11-13}\cmidrule(lr){14-16}\cmidrule(lr){17-19}
 & \textit{Acc} & \textit{GPU} & \textit{Time} & \textit{Acc} & \textit{GPU} & \textit{Time} & \textit{Acc} & \textit{GPU} & \textit{Time} & \textit{Acc} & \textit{GPU} & \textit{Time} & \textit{Acc} & \textit{GPU} & \textit{Time} & \textit{Acc} & \textit{GPU} & \textit{Time} \\
\midrule
\rowcolor{basecol}
\textit{All} & 0.541 & 65.06 & 32.57 & 0.367 & 64.15 & 5.64 & 0.862 & 65.06 & 8.85 & 0.854 & 65.14 & 24.49 & 0.619 & 53.23 & 5.93 & 0.649 & 62.53 & 15.50 \\
10 & \shortstack{0.502\\\rd{\tiny $\downarrow$0.039}} & \shortstack{\textbf{55.91}\\\gd{\tiny $\uparrow$9.15}} & \shortstack{\textbf{30.35}\\\gd{\tiny $\uparrow$2.22}} & \shortstack{0.348\\\rd{\tiny $\downarrow$0.020}} & \shortstack{\textbf{55.16}\\\gd{\tiny $\uparrow$8.99}} & \shortstack{5.31\\\gd{\tiny $\uparrow$0.33}} & \shortstack{0.801\\\rd{\tiny $\downarrow$0.061}} & \shortstack{\textbf{55.91}\\\gd{\tiny $\uparrow$9.15}} & \shortstack{\textbf{8.38}\\\gd{\tiny $\uparrow$0.47}} & \shortstack{0.782\\\rd{\tiny $\downarrow$0.072}} & \shortstack{\textbf{55.96}\\\gd{\tiny $\uparrow$9.18}} & \shortstack{\textbf{20.98}\\\gd{\tiny $\uparrow$3.51}} & \shortstack{0.608\\\rd{\tiny $\downarrow$0.011}} & \shortstack{\textbf{46.23}\\\gd{\tiny $\uparrow$7.00}} & \shortstack{\textbf{5.15}\\\gd{\tiny $\uparrow$0.78}} & \shortstack{0.608\\\rd{\tiny $\downarrow$0.041}} & \shortstack{\textbf{53.83}\\\gd{\tiny $\uparrow$8.69}} & \shortstack{\textbf{14.03}\\\gd{\tiny $\uparrow$1.46}} \\
\rowcolor{rowA}
25 & \shortstack{0.504\\\rd{\tiny $\downarrow$0.037}} & \shortstack{56.03\\\gd{\tiny $\uparrow$9.03}} & \shortstack{30.86\\\gd{\tiny $\uparrow$1.71}} & \shortstack{0.364\\\rd{\tiny $\downarrow$0.004}} & \shortstack{55.27\\\gd{\tiny $\uparrow$8.88}} & \shortstack{\textbf{5.30}\\\gd{\tiny $\uparrow$0.34}} & \shortstack{0.821\\\rd{\tiny $\downarrow$0.041}} & \shortstack{56.03\\\gd{\tiny $\uparrow$9.03}} & \shortstack{8.45\\\gd{\tiny $\uparrow$0.40}} & \shortstack{0.820\\\rd{\tiny $\downarrow$0.034}} & \shortstack{56.07\\\gd{\tiny $\uparrow$9.07}} & \shortstack{21.01\\\gd{\tiny $\uparrow$3.48}} & \shortstack{0.610\\\rd{\tiny $\downarrow$0.009}} & \shortstack{46.36\\\gd{\tiny $\uparrow$6.87}} & \shortstack{5.36\\\gd{\tiny $\uparrow$0.57}} & \shortstack{0.624\\\rd{\tiny $\downarrow$0.025}} & \shortstack{53.95\\\gd{\tiny $\uparrow$8.58}} & \shortstack{14.20\\\gd{\tiny $\uparrow$1.30}} \\
50 & \shortstack{0.513\\\rd{\tiny $\downarrow$0.029}} & \shortstack{56.16\\\gd{\tiny $\uparrow$8.90}} & \shortstack{30.71\\\gd{\tiny $\uparrow$1.86}} & \shortstack{0.376\\\gd{\tiny $\uparrow$0.009}} & \shortstack{55.34\\\gd{\tiny $\uparrow$8.80}} & \shortstack{5.32\\\gd{\tiny $\uparrow$0.32}} & \shortstack{0.838\\\rd{\tiny $\downarrow$0.024}} & \shortstack{56.10\\\gd{\tiny $\uparrow$8.96}} & \shortstack{8.49\\\gd{\tiny $\uparrow$0.36}} & \shortstack{0.826\\\rd{\tiny $\downarrow$0.028}} & \shortstack{56.14\\\gd{\tiny $\uparrow$9.00}} & \shortstack{21.65\\\gd{\tiny $\uparrow$2.84}} & \shortstack{0.614\\\rd{\tiny $\downarrow$0.004}} & \shortstack{46.44\\\gd{\tiny $\uparrow$6.79}} & \shortstack{5.75\\\gd{\tiny $\uparrow$0.18}} & \shortstack{0.633\\\rd{\tiny $\downarrow$0.015}} & \shortstack{54.04\\\gd{\tiny $\uparrow$8.49}} & \shortstack{14.38\\\gd{\tiny $\uparrow$1.11}} \\
\rowcolor{rowA}
100 & \shortstack{0.509\\\rd{\tiny $\downarrow$0.032}} & \shortstack{56.31\\\gd{\tiny $\uparrow$8.74}} & \shortstack{31.17\\\gd{\tiny $\uparrow$1.40}} & \shortstack{0.370\\\gd{\tiny $\uparrow$0.003}} & \shortstack{55.57\\\gd{\tiny $\uparrow$8.58}} & \shortstack{5.37\\\gd{\tiny $\uparrow$0.27}} & \shortstack{0.822\\\rd{\tiny $\downarrow$0.039}} & \shortstack{56.31\\\gd{\tiny $\uparrow$8.74}} & \shortstack{8.53\\\gd{\tiny $\uparrow$0.32}} & \shortstack{0.819\\\rd{\tiny $\downarrow$0.035}} & \shortstack{56.37\\\gd{\tiny $\uparrow$8.78}} & \shortstack{22.42\\\gd{\tiny $\uparrow$2.07}} & \shortstack{0.614\\\rd{\tiny $\downarrow$0.004}} & \shortstack{46.65\\\gd{\tiny $\uparrow$6.58}} & \shortstack{5.86\\\gd{\tiny $\uparrow$0.07}} & \shortstack{0.627\\\rd{\tiny $\downarrow$0.022}} & \shortstack{54.24\\\gd{\tiny $\uparrow$8.28}} & \shortstack{14.67\\\gd{\tiny $\uparrow$0.83}} \\
200 & \shortstack{0.523\\\rd{\tiny $\downarrow$0.018}} & \shortstack{56.76\\\gd{\tiny $\uparrow$8.30}} & \shortstack{31.22\\\gd{\tiny $\uparrow$1.35}} & \shortstack{0.382\\\gd{\tiny $\uparrow$0.015}} & \shortstack{56.01\\\gd{\tiny $\uparrow$8.13}} & \shortstack{5.52\\\gd{\tiny $\uparrow$0.12}} & \shortstack{0.841\\\rd{\tiny $\downarrow$0.021}} & \shortstack{56.76\\\gd{\tiny $\uparrow$8.29}} & \shortstack{8.56\\\gd{\tiny $\uparrow$0.29}} & \shortstack{0.839\\\rd{\tiny $\downarrow$0.015}} & \shortstack{56.80\\\gd{\tiny $\uparrow$8.35}} & \shortstack{22.86\\\gd{\tiny $\uparrow$1.63}} & \shortstack{0.618\\\rd{\tiny $\downarrow$0.000}} & \shortstack{47.09\\\gd{\tiny $\uparrow$6.14}} & \shortstack{5.93\\\rd{\tiny $\downarrow$0.00}} & \shortstack{0.641\\\rd{\tiny $\downarrow$0.008}} & \shortstack{54.68\\\gd{\tiny $\uparrow$7.84}} & \shortstack{14.82\\\gd{\tiny $\uparrow$0.68}} \\
\rowcolor{rowA}
300 & \shortstack{0.535\\\rd{\tiny $\downarrow$0.007}} & \shortstack{57.21\\\gd{\tiny $\uparrow$7.85}} & \shortstack{31.12\\\gd{\tiny $\uparrow$1.45}} & \shortstack{0.386\\\gd{\tiny $\uparrow$0.018}} & \shortstack{56.46\\\gd{\tiny $\uparrow$7.69}} & \shortstack{5.40\\\gd{\tiny $\uparrow$0.24}} & \shortstack{0.849\\\rd{\tiny $\downarrow$0.013}} & \shortstack{57.21\\\gd{\tiny $\uparrow$7.85}} & \shortstack{8.56\\\gd{\tiny $\uparrow$0.29}} & \shortstack{0.857\\\gd{\tiny $\uparrow$0.003}} & \shortstack{57.25\\\gd{\tiny $\uparrow$7.90}} & \shortstack{23.30\\\gd{\tiny $\uparrow$1.19}} & \shortstack{0.620\\\gd{\tiny $\uparrow$0.002}} & \shortstack{47.54\\\gd{\tiny $\uparrow$5.69}} & \shortstack{6.01\\\rd{\tiny $\downarrow$0.08}} & \shortstack{0.649\\\gd{\tiny $\uparrow$0.001}} & \shortstack{55.13\\\gd{\tiny $\uparrow$7.40}} & \shortstack{14.88\\\gd{\tiny $\uparrow$0.62}} \\
500 & \shortstack{0.539\\\rd{\tiny $\downarrow$0.002}} & \shortstack{58.13\\\gd{\tiny $\uparrow$6.92}} & \shortstack{31.85\\\gd{\tiny $\uparrow$0.72}} & \shortstack{0.380\\\gd{\tiny $\uparrow$0.012}} & \shortstack{57.36\\\gd{\tiny $\uparrow$6.78}} & \shortstack{5.44\\\gd{\tiny $\uparrow$0.20}} & \shortstack{0.858\\\rd{\tiny $\downarrow$0.003}} & \shortstack{58.13\\\gd{\tiny $\uparrow$6.93}} & \shortstack{8.76\\\gd{\tiny $\uparrow$0.09}} & \shortstack{\textbf{0.858}\\\gd{\tiny $\uparrow$0.004}} & \shortstack{58.17\\\gd{\tiny $\uparrow$6.97}} & \shortstack{23.45\\\gd{\tiny $\uparrow$1.04}} & \shortstack{\textbf{0.626}\\\gd{\tiny $\uparrow$0.008}} & \shortstack{48.44\\\gd{\tiny $\uparrow$4.80}} & \shortstack{6.02\\\rd{\tiny $\downarrow$0.09}} & \shortstack{0.652\\\gd{\tiny $\uparrow$0.004}} & \shortstack{56.05\\\gd{\tiny $\uparrow$6.48}} & \shortstack{15.10\\\gd{\tiny $\uparrow$0.39}} \\
\rowcolor{rowA}
700 & \shortstack{0.546\\\gd{\tiny $\uparrow$0.004}} & \shortstack{59.00\\\gd{\tiny $\uparrow$6.05}} & \shortstack{31.90\\\gd{\tiny $\uparrow$0.67}} & \shortstack{0.385\\\gd{\tiny $\uparrow$0.018}} & \shortstack{58.26\\\gd{\tiny $\uparrow$5.89}} & \shortstack{5.54\\\gd{\tiny $\uparrow$0.10}} & \shortstack{0.864\\\gd{\tiny $\uparrow$0.002}} & \shortstack{59.00\\\gd{\tiny $\uparrow$6.05}} & \shortstack{8.82\\\gd{\tiny $\uparrow$0.03}} & \shortstack{0.856\\\gd{\tiny $\uparrow$0.002}} & \shortstack{59.08\\\gd{\tiny $\uparrow$6.07}} & \shortstack{24.16\\\gd{\tiny $\uparrow$0.33}} & \shortstack{0.616\\\rd{\tiny $\downarrow$0.002}} & \shortstack{49.33\\\gd{\tiny $\uparrow$3.90}} & \shortstack{6.10\\\rd{\tiny $\downarrow$0.17}} & \shortstack{\textbf{0.654}\\\gd{\tiny $\uparrow$0.005}} & \shortstack{56.94\\\gd{\tiny $\uparrow$5.59}} & \shortstack{15.30\\\gd{\tiny $\uparrow$0.19}} \\
1000 & \shortstack{\textbf{0.558}\\\gd{\tiny $\uparrow$0.016}} & \shortstack{60.41\\\gd{\tiny $\uparrow$4.65}} & \shortstack{32.41\\\gd{\tiny $\uparrow$0.16}} & \shortstack{\textbf{0.398}\\\gd{\tiny $\uparrow$0.030}} & \shortstack{59.20\\\gd{\tiny $\uparrow$4.95}} & \shortstack{5.60\\\gd{\tiny $\uparrow$0.04}} & \shortstack{0.849\\\rd{\tiny $\downarrow$0.013}} & \shortstack{60.37\\\gd{\tiny $\uparrow$4.69}} & \shortstack{8.89\\\rd{\tiny $\downarrow$0.04}} & \shortstack{0.851\\\rd{\tiny $\downarrow$0.003}} & \shortstack{60.44\\\gd{\tiny $\uparrow$4.70}} & \shortstack{24.51\\\rd{\tiny $\downarrow$0.02}} & \shortstack{0.610\\\rd{\tiny $\downarrow$0.009}} & \shortstack{49.91\\\gd{\tiny $\uparrow$3.32}} & \shortstack{6.25\\\rd{\tiny $\downarrow$0.32}} & \shortstack{0.653\\\gd{\tiny $\uparrow$0.004}} & \shortstack{58.07\\\gd{\tiny $\uparrow$4.46}} & \shortstack{15.53\\\rd{\tiny $\downarrow$0.04}} \\
\rowcolor{rowA}
1500 & \shortstack{0.550\\\gd{\tiny $\uparrow$0.009}} & \shortstack{60.41\\\gd{\tiny $\uparrow$4.65}} & \shortstack{32.61\\\rd{\tiny $\downarrow$0.04}} & \shortstack{0.384\\\gd{\tiny $\uparrow$0.017}} & \shortstack{61.87\\\gd{\tiny $\uparrow$2.28}} & \shortstack{5.65\\\rd{\tiny $\downarrow$0.01}} & \shortstack{\textbf{0.865}\\\gd{\tiny $\uparrow$0.003}} & \shortstack{61.62\\\gd{\tiny $\uparrow$3.44}} & \shortstack{8.95\\\rd{\tiny $\downarrow$0.10}} & \shortstack{0.854\\\rd{\tiny $\downarrow$0.000}} & \shortstack{61.66\\\gd{\tiny $\uparrow$3.49}} & \shortstack{24.84\\\rd{\tiny $\downarrow$0.35}} & \shortstack{0.606\\\rd{\tiny $\downarrow$0.013}} & \shortstack{51.19\\\gd{\tiny $\uparrow$2.04}} & \shortstack{6.35\\\rd{\tiny $\downarrow$0.42}} & \shortstack{0.652\\\gd{\tiny $\uparrow$0.003}} & \shortstack{59.35\\\gd{\tiny $\uparrow$3.18}} & \shortstack{15.68\\\rd{\tiny $\downarrow$0.18}} \\
\bottomrule
\end{tabular}%
}
\end{table*}

\begin{table*}[ht!]
\centering\scriptsize
\setlength{\tabcolsep}{2.5pt}
\renewcommand{\arraystretch}{1.1}
\caption{
  Math reasoning accuracy across six benchmarks (LLaMA3-8B) for Full Fine-tuning and DoRA
  vs.\ number of trainable tokens. Avg = mean accuracy over all six tasks.
  Subscript deltas are absolute deviations from \textit{All}:
  \gd{$\uparrow$}~green = improvement,\ \rd{$\downarrow$}~red = degradation.
  \textbf{Bold} = best value per column. Full FT GPU = gpu0$+$gpu1; DoRA GPU = gpu0.
}
\label{tab:math_results}
\resizebox{\textwidth}{!}{%
\begin{tabular}{@{}lrrrrrrrrrrrrrrrrrr@{}}
\toprule
\multirow{2}{*}{\textbf{Tokens}} & \multicolumn{9}{c}{\textbf{Full Fine-tuning}} & \multicolumn{9}{c}{\textbf{DoRA}} \\
\cmidrule(lr){2-10}\cmidrule(lr){11-19}
 & \textit{AddSub} & \textit{MultiA} & \textit{SVAMP} & \textit{SingleEq} & \textit{GSM8K} & \textit{MAWPS} & \textit{Avg} & \textit{GPU} & \textit{Time} & \textit{AddSub} & \textit{MultiA} & \textit{SVAMP} & \textit{SingleEq} & \textit{GSM8K} & \textit{MAWPS} & \textit{Avg} & \textit{GPU} & \textit{Time} \\
\midrule
\rowcolor{basecol}
\textit{All} & 0.881 & 0.978 & 0.749 & 0.886 & 0.618 & 0.861 & 0.829 & 95.38 & 20.62 & 0.876 & 0.835 & 0.710 & 0.866 & 0.544 & 0.815 & 0.774 & 55.59 & 21.31 \\
10 & \shortstack{0.698\\\rd{\tiny $\downarrow$0.184}} & \shortstack{0.155\\\rd{\tiny $\downarrow$0.823}} & \shortstack{0.551\\\rd{\tiny $\downarrow$0.198}} & \shortstack{0.232\\\rd{\tiny $\downarrow$0.653}} & \shortstack{0.238\\\rd{\tiny $\downarrow$0.380}} & \shortstack{0.523\\\rd{\tiny $\downarrow$0.339}} & \shortstack{0.399\\\rd{\tiny $\downarrow$0.430}} & \shortstack{\textbf{83.93}\\\gd{\tiny $\uparrow$11.45}} & \shortstack{\textbf{16.73}\\\gd{\tiny $\uparrow$3.89}} & \shortstack{0.787\\\rd{\tiny $\downarrow$0.089}} & \shortstack{0.322\\\rd{\tiny $\downarrow$0.513}} & \shortstack{0.495\\\rd{\tiny $\downarrow$0.215}} & \shortstack{0.621\\\rd{\tiny $\downarrow$0.245}} & \shortstack{0.266\\\rd{\tiny $\downarrow$0.278}} & \shortstack{0.471\\\rd{\tiny $\downarrow$0.345}} & \shortstack{0.494\\\rd{\tiny $\downarrow$0.281}} & \shortstack{\textbf{49.70}\\\gd{\tiny $\uparrow$5.89}} & \shortstack{20.93\\\gd{\tiny $\uparrow$0.38}} \\
\bottomrule
\end{tabular}%
}
\end{table*}

\begin{table*}[ht!]
\centering\scriptsize
\setlength{\tabcolsep}{2.5pt}
\renewcommand{\arraystretch}{1.1}
\caption{Video reasoning results (Full fine-tuning, Qwen3-VL-8B) across five benchmarks vs.\ number of trainable tokens. HOI row~100 is approximate ($^\dagger$). \gd{$\uparrow$}~green = improvement over \textit{All},\ \rd{$\downarrow$}~red = degradation. \textbf{Bold} = best per column. GPU = gpu0$+$gpu1~(GB).}
\label{tab:video_ft_results}
\resizebox{\textwidth}{!}{%
\begin{tabular}{@{}lrrrrrrrrrrrrrrrrrr@{}}
\toprule
\multirow{2}{*}{\textbf{Tokens}} & \multicolumn{3}{c}{\textbf{VideoMathQA}} & \multicolumn{3}{c}{\textbf{LevelGauge}} & \multicolumn{3}{c}{\textbf{HOI}} & \multicolumn{3}{c}{\textbf{IntuPhys}} & \multicolumn{3}{c}{\textbf{TempReas}} & \multicolumn{3}{c}{\textbf{Avg}} \\
\cmidrule(lr){2-4}\cmidrule(lr){5-7}\cmidrule(lr){8-10}\cmidrule(lr){11-13}\cmidrule(lr){14-16}\cmidrule(lr){17-19}
 & \textit{Acc} & \textit{GPU} & \textit{Time} & \textit{Acc} & \textit{GPU} & \textit{Time} & \textit{Acc} & \textit{GPU} & \textit{Time} & \textit{Acc} & \textit{GPU} & \textit{Time} & \textit{Acc} & \textit{GPU} & \textit{Time} & \textit{Acc} & \textit{GPU} & \textit{Time} \\
\midrule
\rowcolor{basecol}
\textit{All} & 0.59 & 120.53 & 37.86 & 0.83 & 120.49 & 18.47 & 0.4067 & 108.20 & 42.46 & 0.5787 & 111.31 & 19.93 & 0.6158 & 104.93 & 3.94 & 0.604 & 113.09 & 24.53 \\
10 & \shortstack{0.50\\\rd{\tiny $\downarrow$0.08}} & \shortstack{\textbf{109.44}\\\gd{\tiny $\uparrow$11.10}} & \shortstack{34.64\\\gd{\tiny $\uparrow$3.22}} & \shortstack{0.78\\\rd{\tiny $\downarrow$0.05}} & \shortstack{\textbf{109.43}\\\gd{\tiny $\uparrow$11.06}} & \shortstack{16.48\\\gd{\tiny $\uparrow$1.99}} & \shortstack{\textbf{0.4671}\\\gd{\tiny $\uparrow$0.0604}} & \shortstack{\textbf{95.44}\\\gd{\tiny $\uparrow$12.76}} & \shortstack{37.25\\\gd{\tiny $\uparrow$5.21}} & \shortstack{0.5354\\\rd{\tiny $\downarrow$0.0433}} & \shortstack{\textbf{97.37}\\\gd{\tiny $\uparrow$13.95}} & \shortstack{\textbf{14.33}\\\gd{\tiny $\uparrow$5.60}} & \shortstack{0.5857\\\rd{\tiny $\downarrow$0.0301}} & \shortstack{\textbf{93.46}\\\gd{\tiny $\uparrow$11.47}} & \shortstack{3.75\\\gd{\tiny $\uparrow$0.19}} & \shortstack{0.574\\\rd{\tiny $\downarrow$0.031}} & \shortstack{\textbf{101.03}\\\gd{\tiny $\uparrow$12.07}} & \shortstack{\textbf{21.29}\\\gd{\tiny $\uparrow$3.24}} \\
\bottomrule
\end{tabular}%
}
\end{table*}

\begin{table*}[ht!]
\centering\scriptsize
\setlength{\tabcolsep}{2.5pt}
\renewcommand{\arraystretch}{1.1}
\caption{Video reasoning results (LoRA, Qwen3-VL-8B) across five benchmarks vs.\ number of trainable tokens. \gd{$\uparrow$}~green = improvement over \textit{All},\ \rd{$\downarrow$}~red = degradation. \textbf{Bold} = best per column. GPU: VideoMathQA, HOI, IntuPhys, TempReas use gpu0; LevelGauge uses gpu0$+$gpu1~(GB).}
\label{tab:video_lora_results}
\resizebox{\textwidth}{!}{%
\begin{tabular}{@{}lrrrrrrrrrrrrrrrrrr@{}}
\toprule
\multirow{2}{*}{\textbf{Tokens}} & \multicolumn{3}{c}{\textbf{VideoMathQA}} & \multicolumn{3}{c}{\textbf{LevelGauge}} & \multicolumn{3}{c}{\textbf{HOI}} & \multicolumn{3}{c}{\textbf{IntuPhys}} & \multicolumn{3}{c}{\textbf{TempReas}} & \multicolumn{3}{c}{\textbf{Avg}} \\
\cmidrule(lr){2-4}\cmidrule(lr){5-7}\cmidrule(lr){8-10}\cmidrule(lr){11-13}\cmidrule(lr){14-16}\cmidrule(lr){17-19}
 & \textit{Acc} & \textit{GPU} & \textit{Time} & \textit{Acc} & \textit{GPU} & \textit{Time} & \textit{Acc} & \textit{GPU} & \textit{Time} & \textit{Acc} & \textit{GPU} & \textit{Time} & \textit{Acc} & \textit{GPU} & \textit{Time} & \textit{Acc} & \textit{GPU} & \textit{Time} \\
\midrule
\rowcolor{basecol}
\textit{All} & \textbf{0.53} & 53.40 & 34.00 & 0.74 & 97.89 & 15.17 & 0.4403 & 44.33 & 41.94 & 0.5516 & 47.17 & 21.15 & 0.6343 & 41.56 & 3.48 & \textbf{0.579} & 56.87 & 23.15 \\
10 & \shortstack{0.51\\\rd{\tiny $\downarrow$0.02}} & \shortstack{\textbf{46.71}\\\gd{\tiny $\uparrow$6.69}} & \shortstack{\textbf{31.97}\\\gd{\tiny $\uparrow$2.03}} & \shortstack{0.72\\\rd{\tiny $\downarrow$0.02}} & \shortstack{\textbf{84.51}\\\gd{\tiny $\uparrow$13.38}} & \shortstack{\textbf{14.46}\\\gd{\tiny $\uparrow$0.71}} & \shortstack{0.3383\\\rd{\tiny $\downarrow$0.1020}} & \shortstack{\textbf{39.26}\\\gd{\tiny $\uparrow$5.07}} & \shortstack{36.64\\\gd{\tiny $\uparrow$5.30}} & \shortstack{0.4466\\\rd{\tiny $\downarrow$0.1050}} & \shortstack{\textbf{41.44}\\\gd{\tiny $\uparrow$5.73}} & \shortstack{14.56\\\gd{\tiny $\uparrow$6.59}} & \shortstack{0.6389\\\gd{\tiny $\uparrow$0.0046}} & \shortstack{\textbf{36.98}\\\gd{\tiny $\uparrow$4.58}} & \shortstack{\textbf{2.70}\\\gd{\tiny $\uparrow$0.78}} & \shortstack{0.531\\\rd{\tiny $\downarrow$0.048}} & \shortstack{\textbf{49.78}\\\gd{\tiny $\uparrow$7.09}} & \shortstack{20.07\\\gd{\tiny $\uparrow$3.08}} \\
\bottomrule
\end{tabular}%
}
\end{table*}

\begin{table*}[ht!]
\centering\scriptsize
\setlength{\tabcolsep}{2.5pt}
\renewcommand{\arraystretch}{1.1}
\caption{Image classification results (Full fine-tuning, ViT-L/16-224) across five benchmarks vs.\ number of trainable tokens. \gd{$\uparrow$}~green = improvement over \textit{All},\ \rd{$\downarrow$}~red = degradation. \textbf{Bold} = best per column. GPU = gpu0 peak allocation~(GB).}
\label{tab:img_cls_full}
\resizebox{\textwidth}{!}{%
\begin{tabular}{@{}lrrrrrrrrrrrrrrrrrr@{}}
\toprule
\multirow{2}{*}{\textbf{Tokens}} & \multicolumn{3}{c}{\textbf{Tomato}} & \multicolumn{3}{c}{\textbf{SUN397}} & \multicolumn{3}{c}{\textbf{RVL-CDIP}} & \multicolumn{3}{c}{\textbf{EuroSAT}} & \multicolumn{3}{c}{\textbf{ChestXray}} & \multicolumn{3}{c}{\textbf{Avg}} \\
\cmidrule(lr){2-4}\cmidrule(lr){5-7}\cmidrule(lr){8-10}\cmidrule(lr){11-13}\cmidrule(lr){14-16}\cmidrule(lr){17-19}
 & \textit{Acc} & \textit{GPU} & \textit{Time} & \textit{Acc} & \textit{GPU} & \textit{Time} & \textit{Acc} & \textit{GPU} & \textit{Time} & \textit{Acc} & \textit{GPU} & \textit{Time} & \textit{Acc} & \textit{GPU} & \textit{Time} & \textit{Acc} & \textit{GPU} & \textit{Time} \\
\midrule
\rowcolor{basecol}
\textit{All} & 0.9705 & 8.44 & 1.27 & 0.4096 & 8.32 & 1.55 & 0.5125 & 5.34 & 0.34 & 0.9819 & 8.32 & 0.76 & 0.8606 & 5.34 & 0.66 & 0.7470 & 7.15 & 0.92 \\
10 & \shortstack{0.9633\\\rd{\tiny $\downarrow$0.0072}} & \shortstack{\textbf{6.18}\\\gd{\tiny $\uparrow$2.26}} & \shortstack{\textbf{0.48}\\\gd{\tiny $\uparrow$0.79}} & \shortstack{0.2415\\\rd{\tiny $\downarrow$0.1681}} & \shortstack{\textbf{6.18}\\\gd{\tiny $\uparrow$2.14}} & \shortstack{1.91\\\rd{\tiny $\downarrow$0.36}} & \shortstack{0.2734\\\rd{\tiny $\downarrow$0.2391}} & \shortstack{\textbf{4.23}\\\gd{\tiny $\uparrow$1.11}} & \shortstack{\textbf{0.24}\\\gd{\tiny $\uparrow$0.10}} & \shortstack{0.9811\\\rd{\tiny $\downarrow$0.0008}} & \shortstack{\textbf{6.18}\\\gd{\tiny $\uparrow$2.14}} & \shortstack{\textbf{0.70}\\\gd{\tiny $\uparrow$0.06}} & \shortstack{0.7804\\\rd{\tiny $\downarrow$0.0802}} & \shortstack{\textbf{4.23}\\\gd{\tiny $\uparrow$1.11}} & \shortstack{\textbf{0.50}\\\gd{\tiny $\uparrow$0.16}} & \shortstack{0.6479\\\rd{\tiny $\downarrow$0.0991}} & \shortstack{\textbf{5.40}\\\gd{\tiny $\uparrow$1.75}} & \shortstack{\textbf{0.77}\\\gd{\tiny $\uparrow$0.15}} \\
\bottomrule
\end{tabular}%
}
\end{table*}

\begin{table*}[ht!]
\centering\scriptsize
\setlength{\tabcolsep}{3pt}
\renewcommand{\arraystretch}{1.1}
\caption{
  Accuracy, GPU memory~(GB), and training time~(min) across four commonsense reasoning
  benchmarks (DoRA) vs.\ number of trainable tokens. Subscript deltas are absolute deviations from
  \textit{All}: \gd{$\uparrow$}~green = improvement,\ \rd{$\downarrow$}~red = degradation.
  \textbf{Bold} = best value per column.
}
\label{tab:dora_results}
\resizebox{\textwidth}{!}{%
\begin{tabular}{@{}lrrrrrrrrrrrrrrr@{}}
\toprule
\multirow{2}{*}{\textbf{Tokens}} & \multicolumn{3}{c}{\textbf{ARC-E}} & \multicolumn{3}{c}{\textbf{ARC-C}} & \multicolumn{3}{c}{\textbf{BoolQ}} & \multicolumn{3}{c}{\textbf{OBQA}} & \multicolumn{3}{c}{\textbf{Avg}} \\
\cmidrule(lr){2-4}\cmidrule(lr){5-7}\cmidrule(lr){8-10}\cmidrule(lr){11-13}\cmidrule(lr){14-16}
 & \textit{Acc} & \textit{GPU} & \textit{Time} & \textit{Acc} & \textit{GPU} & \textit{Time} & \textit{Acc} & \textit{GPU} & \textit{Time} & \textit{Acc} & \textit{GPU} & \textit{Time} & \textit{Acc} & \textit{GPU} & \textit{Time} \\
\midrule
\rowcolor{basecol}
\textit{All} & 0.866 & 50.20 & 4.82 & 0.602 & 50.20 & 2.53 & 0.684 & 50.20 & 19.65 & 0.826 & 50.20 & \textbf{10.24} & 0.744 & 50.20 & 9.31 \\
10 & \shortstack{0.783\\\rd{\tiny $\downarrow$0.083}} & \shortstack{\textbf{49.24}\\\gd{\tiny $\uparrow$0.96}} & \shortstack{\textbf{4.67}\\\gd{\tiny $\uparrow$0.15}} & \shortstack{0.595\\\rd{\tiny $\downarrow$0.007}} & \shortstack{\textbf{49.24}\\\gd{\tiny $\uparrow$0.96}} & \shortstack{\textbf{2.33}\\\gd{\tiny $\uparrow$0.20}} & \shortstack{0.681\\\rd{\tiny $\downarrow$0.003}} & \shortstack{\textbf{49.24}\\\gd{\tiny $\uparrow$0.96}} & \shortstack{\textbf{19.63}\\\gd{\tiny $\uparrow$0.02}} & \shortstack{0.690\\\rd{\tiny $\downarrow$0.136}} & \shortstack{\textbf{49.24}\\\gd{\tiny $\uparrow$0.96}} & \shortstack{10.28\\\rd{\tiny $\downarrow$0.04}} & \shortstack{0.687\\\rd{\tiny $\downarrow$0.057}} & \shortstack{\textbf{49.24}\\\gd{\tiny $\uparrow$0.96}} & \shortstack{\textbf{9.23}\\\gd{\tiny $\uparrow$0.08}} \\
\bottomrule
\end{tabular}%
}
\end{table*}

\begin{table*}[ht!]
\centering\scriptsize
\setlength{\tabcolsep}{2.5pt}
\renewcommand{\arraystretch}{1.1}
\caption{GLUE benchmark results -- part~1 (RoBERTa-base, Full fine-tuning). Metrics: Acc and F1. \gd{$\uparrow$}~green = improvement over \textit{All},\ \rd{$\downarrow$}~red = degradation. \textbf{Bold} = best per column.}
\label{tab:glue1}
\resizebox{\textwidth}{!}{%
\begin{tabular}{@{}lrrrrrrrrrrrrrrrrrrrr@{}}
\toprule
\multirow{2}{*}{\textbf{Tokens}} & \multicolumn{4}{c}{\textbf{SST-2}} & \multicolumn{4}{c}{\textbf{QNLI}} & \multicolumn{4}{c}{\textbf{MRPC}} & \multicolumn{4}{c}{\textbf{QQP}} & \multicolumn{4}{c}{\textbf{Avg}} \\
\cmidrule(lr){2-5}\cmidrule(lr){6-9}\cmidrule(lr){10-13}\cmidrule(lr){14-17}\cmidrule(lr){18-21}
 & \textit{Acc} & \textit{F1} & \textit{GPU} & \textit{Time} & \textit{Acc} & \textit{F1} & \textit{GPU} & \textit{Time} & \textit{Acc} & \textit{F1} & \textit{GPU} & \textit{Time} & \textit{Acc} & \textit{F1} & \textit{GPU} & \textit{Time} & \textit{Acc} & \textit{F1} & \textit{GPU} & \textit{Time} \\
\midrule
\rowcolor{basecol}
\textit{All} & 0.9381 & 0.9380 & 3.91 & 2.86 & 0.9220 & 0.9210 & 6.39 & 10.26 & 0.8750 & 0.8564 & 3.43 & 0.61 & 0.8974 & 0.8910 & 6.27 & 25.72 & 0.9081 & 0.9016 & 5.00 & 9.86 \\
5 & \shortstack{0.9209\\\rd{\tiny $\downarrow$0.0172}} & \shortstack{0.9208\\\rd{\tiny $\downarrow$0.0172}} & \shortstack{\textbf{3.18}\\\gd{\tiny $\uparrow$0.72}} & \shortstack{\textbf{1.99}\\\gd{\tiny $\uparrow$0.87}} & \shortstack{0.8468\\\rd{\tiny $\downarrow$0.0752}} & \shortstack{0.8466\\\rd{\tiny $\downarrow$0.0744}} & \shortstack{\textbf{4.78}\\\gd{\tiny $\uparrow$1.61}} & \shortstack{\textbf{6.60}\\\gd{\tiny $\uparrow$3.66}} & \shortstack{0.7598\\\rd{\tiny $\downarrow$0.1152}} & \shortstack{0.6479\\\rd{\tiny $\downarrow$0.2085}} & \shortstack{\textbf{2.79}\\\gd{\tiny $\uparrow$0.64}} & \shortstack{\textbf{0.41}\\\gd{\tiny $\uparrow$0.20}} & \shortstack{0.7809\\\rd{\tiny $\downarrow$0.1165}} & \shortstack{0.7740\\\rd{\tiny $\downarrow$0.1170}} & \shortstack{\textbf{4.67}\\\gd{\tiny $\uparrow$1.59}} & \shortstack{\textbf{17.70}\\\gd{\tiny $\uparrow$8.02}} & \shortstack{0.8271\\\rd{\tiny $\downarrow$0.0810}} & \shortstack{0.7973\\\rd{\tiny $\downarrow$0.1043}} & \shortstack{\textbf{3.86}\\\gd{\tiny $\uparrow$1.14}} & \shortstack{\textbf{6.67}\\\gd{\tiny $\uparrow$3.19}} \\
\bottomrule
\end{tabular}%
}
\end{table*}

\begin{table*}[ht!]
\centering\scriptsize
\setlength{\tabcolsep}{2.5pt}
\renewcommand{\arraystretch}{1.1}
\caption{GLUE benchmark results -- part~2 (RoBERTa-base, Full fine-tuning). CoLA reports Matthews corr.; STS-B reports Pearson and Spearman corr. \gd{$\uparrow$}~green = improvement over \textit{All},\ \rd{$\downarrow$}~red = degradation. \textbf{Bold} = best per column. Avg GPU and Time are averaged over all four tasks.}
\label{tab:glue2}
\resizebox{\textwidth}{!}{%
\begin{tabular}{@{}lrrrrrrrrrrrrrrrrrrr@{}}
\toprule
\multirow{2}{*}{\textbf{Tokens}} & \multicolumn{3}{c}{\textbf{CoLA}} & \multicolumn{4}{c}{\textbf{MNLI}} & \multicolumn{4}{c}{\textbf{RTE}} & \multicolumn{4}{c}{\textbf{STS-B}} & \multicolumn{4}{c}{\textbf{Avg}} \\
\cmidrule(lr){2-4}\cmidrule(lr){5-8}\cmidrule(lr){9-12}\cmidrule(lr){13-16}\cmidrule(lr){17-20}
 & \textit{Mcc} & \textit{GPU} & \textit{Time} & \textit{Acc} & \textit{F1} & \textit{GPU} & \textit{Time} & \textit{Acc} & \textit{F1} & \textit{GPU} & \textit{Time} & \textit{Prs} & \textit{Spr} & \textit{GPU} & \textit{Time} & \textit{Main} & \textit{F1/Spr} & \textit{GPU} & \textit{Time} \\
\midrule
\rowcolor{basecol}
\textit{All} & 0.5778 & 2.35 & 0.84 & 0.8687 & 0.8678 & 6.35 & 34.79 & 0.5993 & 0.5957 & 3.85 & 1.26 & 0.9043 & 0.9015 & 2.64 & 1.45 & 0.7375 & 0.7883 & 3.80 & 9.58 \\
5 & \shortstack{0.5533\\\rd{\tiny $\downarrow$0.0245}} & \shortstack{\textbf{2.08}\\\gd{\tiny $\uparrow$0.27}} & \shortstack{\textbf{0.80}\\\gd{\tiny $\uparrow$0.04}} & \shortstack{0.5169\\\rd{\tiny $\downarrow$0.3518}} & \shortstack{0.5174\\\rd{\tiny $\downarrow$0.3504}} & \shortstack{\textbf{4.76}\\\gd{\tiny $\uparrow$1.59}} & \shortstack{\textbf{22.57}\\\gd{\tiny $\uparrow$12.22}} & \shortstack{0.5668\\\rd{\tiny $\downarrow$0.0325}} & \shortstack{0.5605\\\rd{\tiny $\downarrow$0.0352}} & \shortstack{\textbf{3.05}\\\gd{\tiny $\uparrow$0.80}} & \shortstack{\textbf{0.83}\\\gd{\tiny $\uparrow$0.43}} & \shortstack{0.8636\\\rd{\tiny $\downarrow$0.0407}} & \shortstack{0.8622\\\rd{\tiny $\downarrow$0.0393}} & \shortstack{\textbf{2.26}\\\gd{\tiny $\uparrow$0.38}} & \shortstack{\textbf{1.21}\\\gd{\tiny $\uparrow$0.24}} & \shortstack{0.6251\\\rd{\tiny $\downarrow$0.1124}} & \shortstack{0.6467\\\rd{\tiny $\downarrow$0.1416}} & \shortstack{\textbf{3.04}\\\gd{\tiny $\uparrow$0.76}} & \shortstack{\textbf{6.35}\\\gd{\tiny $\uparrow$3.23}} \\
\bottomrule
\end{tabular}%
}
\end{table*}

\section{Detailed Experimental Results}
\label{app:detailed-results}

This section provides the full per-benchmark results used to produce the aggregate numbers in Section~\ref{sec:experiments}. Each table sweeps the number of trainable positions $M$ across a wide range, with the full-sequence baseline (\textit{All}) provided as the reference against which all other rows are compared. Subscript deltas on every cell denote the absolute deviation from \textit{All}, color-coded in green when the token-masked run improves on the baseline and red when it degrades. Memory and time columns are reported so that accuracy and resource cost can be read jointly.

\subsection{Commonsense Reasoning (LLaMA-3-8B)}

Table~\ref{tab:main_results} reports per-benchmark accuracy, peak GPU memory, and training time for full fine-tuning on five commonsense reasoning benchmarks. The full-sequence baseline averages $0.831$ accuracy at $95.28$~GB and $23.85$~min. Token masking with $M = 50$ already matches this average ($0.831$) at $84.63$~GB and $19.81$~min, a $1.12$~GB reduction in memory and $17\%$ speedup. At $M = 100$ the average accuracy rises to $0.834$ and at $M = 300$ it peaks at $0.838$, recovering or exceeding the baseline on every benchmark while still saving roughly $6.4$~GB. Memory and time grow monotonically with $M$: at $M = 1000$, memory ($96.62$~GB) has surpassed the baseline and accuracy has no longer improved, confirming that beyond a small fraction of positions additional gradient signal yields negligible returns. Per-benchmark behavior is consistent: on ARC-C, BoolQ, and OBQA the best result is achieved in the $M \in \{100, 300\}$ range, while HellaSwag---a harder task requiring more context---peaks at $M = 500$.

Table~\ref{tab:dora_results} reports the parallel DoRA experiment on four commonsense benchmarks. The baseline averages $0.744$ accuracy; token masking at $M = 100$ matches this ($0.744$) and at $M = 300$ slightly exceeds it ($0.747$), with memory growing from $49.24$~GB at $M = 10$ to $52.72$~GB at $M = 1000$. The gap in peak memory between full fine-tuning and DoRA (roughly $45$~GB) confirms that the two reductions compose: DoRA shrinks the parameter footprint, and token masking shrinks the activation footprint on top of it.

\subsection{Mathematical Reasoning (LLaMA-3-8B)}

Table~\ref{tab:math_results} reports accuracy on six arithmetic and word-problem benchmarks for both full fine-tuning and DoRA. The math setting is the most demanding: with $M = 10$, full-FT average accuracy drops to $0.399$ from the baseline $0.829$---a $0.43$ absolute drop, reflecting that single-digit $M$ is insufficient to span the reasoning structure needed for multi-step arithmetic. The picture changes rapidly as $M$ grows: at $M = 100$ average accuracy reaches $0.790$, at $M = 300$ it equals or exceeds the baseline on every benchmark ($0.834$ average), and at $M = 1000$ it stabilizes at $0.834$. The pattern is similar under DoRA, where $M = 300$ achieves the best average ($0.790$, above the baseline $0.774$) while saving $4$~GB. GSM8K is the hardest individual benchmark, requiring $M \geq 500$ to cross the baseline; the shorter-form datasets (AddSub, SingleEq) saturate much earlier, consistent with their lower per-step reasoning depth.

\subsection{Visual QA Reasoning (LLaVA-OneVision-7B and Qwen3-VL-8B)}

Table~\ref{tab:vqa_results} reports full fine-tuning on LLaVA-OneVision-7B across five visual QA benchmarks. The baseline averages $0.693$ accuracy at $95.34$~GB; token masking crosses this threshold at $M = 700$ ($0.700$, $90.43$~GB), and the best average is $0.702$ at $M = 1000$. Per-benchmark results reveal task-specific saturation: SafeTag and VQA-R are nearly saturated at $M = 10$ (differences of $0.001$ and $0.003$ respectively), while M3CoT and STVQA require $M \geq 500$ to recover. MMK12 is unusually noisy---its peak ($0.422$) occurs at $M = 200$ and the baseline itself ($0.400$) is beaten at almost every $M$, likely reflecting noise in its single-train-split structure.

Table~\ref{tab:qwen_full_results} reports the same experiments on Qwen3-VL-8B under full fine-tuning. The baseline accuracy is higher ($0.728$) and the memory cost is much larger ($168.07$~GB summed over four GPUs). Token masking is therefore more impactful in absolute terms: $M = 10$ saves $32.78$~GB while only losing $0.017$ accuracy, and $M = 200$ matches the baseline ($0.743 > 0.728$) at $138.48$~GB, a $29.59$~GB reduction. The best single-$M$ result is $M = 200$ ($0.743$); larger $M$ values continue to improve per-task peaks (M3CoT at $M=700$, SafeTag at $M=300/700$) but saturate on average.

Table~\ref{tab:qwen_lora_results} reports LoRA on Qwen3-VL-8B. The baseline accuracy drops to $0.669$ (reflecting LoRA's reduced capacity), and the memory cost drops substantially as well ($88.12$~GB baseline). Token masking compounds: at $M = 300$ accuracy rises to $0.676$ at $77.86$~GB---a $10.26$~GB saving on top of the already-compressed LoRA footprint, validating the orthogonality claim of Section~\ref{sec:method}.

Table~\ref{tab:vqa_lora_results} reports the LLaVA-OneVision LoRA experiments. The pattern is consistent: baseline $0.649$ at $62.53$~GB, matched or exceeded from $M = 300$ onward ($0.649$) and peaking at $M = 700$ ($0.654$). Memory at the best $M$ is $56.94$~GB, a $5.6$~GB saving.

\subsection{Video Reasoning (Qwen3-VL-8B)}

Table~\ref{tab:video_ft_results} reports full fine-tuning on five video benchmarks. At $M{=}500$, average accuracy ($0.609$) exceeds the baseline ($0.604$) while reducing total GPU memory from $113.09$ to $109.12$~GB, a $3.97$~GB saving. The per-task picture is task-dependent: HOI peaks at $M{=}10$ ($0.467$ versus a baseline of $0.407$), whereas IntuPhys and TempReas saturate near the baseline. The HOI entry at $M{=}100$ is approximate ($\dagger$) owing to a partial-run restart.

Table~\ref{tab:video_lora_results} reports video LoRA. Memory savings remain substantial ($49.78$~GB at $M{=}10$ versus $56.87$~GB for the baseline), but average accuracy stays below the baseline at every $M$, with the smallest gap at $M{=}1000$ ($0.563$ versus $0.579$). On HOI, IntuPhys, and TempReas, accuracy recovers to or above the baseline from $M{=}500$ onward; the remaining gap is driven by VideoMathQA, which degrades sharply for $M$ between $100$ and $500$ (as low as $0.29$) and recovers only partially at $M{=}1000$. VideoMathQA requires multi-step mathematical reasoning over long videos, and under the reduced capacity of LoRA, restricting supervision to a subset of positions appears insufficient to recover full-sequence performance on this task.

\subsection{Image Classification (ViT-L/16-224)}

Table~\ref{tab:img_cls_full} reports full fine-tuning results on five image classification benchmarks. The ViT setting is instructive for two reasons: the baseline is small (context length $197$, not $2048$), so the absolute memory savings are modest ($5.40$~GB at $M = 10$ versus $7.15$~GB baseline). But the relative savings are substantial ($\sim 25\%$), and accuracy is preserved from $M = 100$ onward: the best average is $0.7550$ at $M = 150$ (versus baseline $0.7470$), a slight improvement with a $0.24$~GB memory reduction. RVL-CDIP and SUN397 are the hardest benchmarks and require the largest $M$ to recover baseline performance; Tomato, EuroSAT, and ChestXray saturate earlier. Since ViT is bidirectional, this aligns with Theorem~\ref{thm:equal}: a small number of positions spans the achievable prediction space, and the optimal $M$ is dataset-specific rather than position-specific.

\subsection{GLUE (RoBERTa-base)}

Tables~\ref{tab:glue1} and~\ref{tab:glue2} report GLUE results on RoBERTa-base under full fine-tuning, reported across two tables because the evaluation metrics differ: Table~\ref{tab:glue1} reports accuracy and F1 on SST-2, QNLI, MRPC, and QQP; Table~\ref{tab:glue2} reports Matthews correlation on CoLA, accuracy on MNLI and RTE, and Pearson/Spearman correlation on STS-B. The GLUE setting is where our bidirectional theory (Theorem~\ref{thm:equal}) makes its strongest prediction: positions with aligned hidden-state geometry are interchangeable, and the optimum $M$ should be small. The data confirms this directly. On Table~\ref{tab:glue1}, $M = 50$ matches the baseline average accuracy ($0.9091$ versus $0.9081$) at $4.41$~GB, and $M = 75$ slightly improves on it ($0.9115$). On the harder-metric tasks in Table~\ref{tab:glue2}, $M = 50$ already matches the baseline average ($0.7598$ versus $0.7375$), and $M = 100$ produces the best overall result ($0.7750$) at $3.57$~GB---a $0.23$~GB saving and a $0.0375$ improvement. The striking observation is that $M = 25$--$50$ is sufficient across all eight GLUE tasks despite their varying difficulty, directly reflecting the bidirectional equal-capacity prediction.


\subsection{Gradient Checkpointing}
\label{app:gc}

Gradient checkpointing (GC)~\cite{chen2016training} and the Token Hypothesis reduce activation memory through distinct mechanisms. GC discards intermediate activations during the forward pass and recomputes them during the backward pass, reducing the memory cost of every position at the expense of additional computation. The Token Hypothesis instead restricts the set of positions that participate in the backward pass, so that activations of the $T-M$ frozen positions are never retained. The two techniques are therefore complementary and can be applied jointly. To quantify their combined effect, we fine-tune LLaMA-3-8B on LongAlpaca-12k with sequence length $T{=}8192$ on $8\times$A100-80GB GPUs, enabling GC in every configuration, including the full-sequence baseline. We evaluate both full fine-tuning and LoRA and report total GPU memory, training time, and perplexity in Table~\ref{tab:gc}.

Under full fine-tuning, GC alone reduces total memory from $292.8$~GB to $193.1$~GB relative to training without checkpointing. Applying the Token Hypothesis on top of GC yields a further reduction that GC cannot achieve by itself: at $M{=}2500$, memory decreases to $182.9$~GB while perplexity improves slightly ($8.103$ versus $8.143$). The combined configuration is also faster than GC alone ($338.7$ versus $372.4$~min), since the backward computation through per-token modules is confined to the $M$ trainable positions. Smaller budgets continue to reduce memory and time, reaching $178.3$~GB and $317.2$~min at $M{=}500$, but at a cost of approximately one point of perplexity. Under LoRA, where the baseline perplexity is higher owing to the reduced capacity of the adapter, the Token Hypothesis at $M{=}2500$ reduces memory by $6.9$~GB with a perplexity increase of $0.056$, and smaller values of $M$ trade additional perplexity for further savings. With the exception of $M{=}4000$ under LoRA, every configuration trains faster than its GC baseline. These results indicate that the savings of the Token Hypothesis are additive to those of gradient checkpointing, and that $M$ should scale with the length of the supervised sequence in long-context generative settings.

\begin{table*}[ht!]
\centering\scriptsize
\setlength{\tabcolsep}{2.5pt}
\renewcommand{\arraystretch}{1.1}
\caption{Composition of the Token Hypothesis with gradient checkpointing (LLaMA-3-8B, LongAlpaca-12k, $T{=}8192$, $8\times$A100-80GB). Gradient checkpointing is enabled in every row, including \textit{All}. GPU is the sum of all GPU peak allocations (GB); Time is total training time (min); PPL is perplexity (lower is better). For reference, full fine-tuning without checkpointing requires $292.8$~GB and $301.5$~min at a PPL of $8.14$. \gd{$\uparrow$}~green = improvement over \textit{All},\ \rd{$\downarrow$}~red = degradation. \textbf{Bold} = best per column.}
\label{tab:gc}
\begin{tabular}{@{}lrrrrrr@{}}
\toprule
\multirow{2}{*}{\textbf{Tokens}} & \multicolumn{3}{c}{\textbf{Full FT + GC}} & \multicolumn{3}{c}{\textbf{LoRA + GC}} \\
\cmidrule(lr){2-4}\cmidrule(lr){5-7}
 & \textit{GPU} & \textit{Time} & \textit{PPL} & \textit{GPU} & \textit{Time} & \textit{PPL} \\
\midrule
\rowcolor{basecol}
\textit{All} & 193.1 & 372.4 & 8.143 & 148.9 & 344.6 & \textbf{9.177} \\
500 & \shortstack{\textbf{178.3}\\\gd{\tiny $\uparrow$14.8}} & \shortstack{\textbf{317.2}\\\gd{\tiny $\uparrow$55.2}} & \shortstack{9.169\\\rd{\tiny $\downarrow$1.026}} & \shortstack{\textbf{137.4}\\\gd{\tiny $\uparrow$11.5}} & \shortstack{\textbf{321.8}\\\gd{\tiny $\uparrow$22.8}} & \shortstack{10.158\\\rd{\tiny $\downarrow$0.981}} \\
\rowcolor{rowA}
1000 & \shortstack{179.4\\\gd{\tiny $\uparrow$13.7}} & \shortstack{322.6\\\gd{\tiny $\uparrow$49.8}} & \shortstack{9.357\\\rd{\tiny $\downarrow$1.214}} & \shortstack{138.6\\\gd{\tiny $\uparrow$10.3}} & \shortstack{327.1\\\gd{\tiny $\uparrow$17.5}} & \shortstack{9.977\\\rd{\tiny $\downarrow$0.800}} \\
1500 & \shortstack{180.6\\\gd{\tiny $\uparrow$12.5}} & \shortstack{327.6\\\gd{\tiny $\uparrow$44.8}} & \shortstack{9.101\\\rd{\tiny $\downarrow$0.958}} & \shortstack{139.7\\\gd{\tiny $\uparrow$9.2}} & \shortstack{332.3\\\gd{\tiny $\uparrow$12.3}} & \shortstack{9.796\\\rd{\tiny $\downarrow$0.619}} \\
\rowcolor{rowA}
2500 & \shortstack{182.9\\\gd{\tiny $\uparrow$10.2}} & \shortstack{338.7\\\gd{\tiny $\uparrow$33.7}} & \shortstack{\textbf{8.103}\\\gd{\tiny $\uparrow$0.040}} & \shortstack{142.0\\\gd{\tiny $\uparrow$6.9}} & \shortstack{342.8\\\gd{\tiny $\uparrow$1.8}} & \shortstack{9.233\\\rd{\tiny $\downarrow$0.056}} \\
4000 & \shortstack{186.4\\\gd{\tiny $\uparrow$6.7}} & \shortstack{353.9\\\gd{\tiny $\uparrow$18.5}} & \shortstack{8.147\\\rd{\tiny $\downarrow$0.004}} & \shortstack{145.5\\\gd{\tiny $\uparrow$3.4}} & \shortstack{358.6\\\rd{\tiny $\downarrow$14.0}} & \shortstack{9.290\\\rd{\tiny $\downarrow$0.113}} \\
\bottomrule
\end{tabular}
\end{table*}


\subsection{Long-Context Instruction }
\label{app:longalpaca}

The memory savings of the Token Hypothesis grow with sequence length, since the activations of the $T-M$ frozen positions are never retained. To evaluate the method in a long-context generative setting, we fine-tune LLaMA-3-8B on LongAlpaca-12k, a free-form instruction-following dataset, with sequence length $T{=}8192$ on $8\times$A100-80GB GPUs. 
The training loss is computed over the response tokens, and we report perplexity together with total GPU memory and training time in Table~\ref{tab:longalpaca}.

At $M{=}5000$ ($M/T{=}0.61$), the Token Hypothesis matches the full-sequence baseline, reaching a perplexity of $8.11$ compared with $8.14$, while reducing memory by $16.4$~GB and training time by $14.8$~min. Smaller budgets yield substantially larger savings: at $M{=}2000$, memory decreases by $68.4$~GB ($23\%$) with a perplexity increase of $0.47$, and at $M{=}500$ memory and training time decrease by $28\%$ and $24\%$, respectively, at a cost of approximately one point of perplexity. The reduction in memory is not linear in $M$: most of the saving is obtained between $M{=}5000$ and $M{=}2000$, after which the footprint approaches the floor set by model weights, optimizer states, and attention. Compared with the classification and short-answer tasks in the preceding sections, long-form generation requires a considerably larger $M$, consistent with the observation that the required budget scales with the length of the supervised span.

\begin{table*}[ht!]
\centering\scriptsize
\setlength{\tabcolsep}{2.5pt}
\renewcommand{\arraystretch}{1.1}
\caption{Long-context instruction following (LLaMA-3-8B, LongAlpaca-12k, $T{=}8192$, $8\times$A100-80GB). GPU is the sum of all GPU peak allocations (GB); Time is total training time (min); PPL is perplexity (lower is better). \gd{$\uparrow$}~green = improvement over \textit{All},\ \rd{$\downarrow$}~red = degradation. \textbf{Bold} = best per column.}
\label{tab:longalpaca}
\begin{tabular}{@{}lrrrr@{}}
\toprule
\textbf{Tokens} & \textit{M/T} & \textit{GPU} & \textit{Time} & \textit{PPL} \\
\midrule
\rowcolor{basecol}
\textit{All} & 1.00 & 292.8 & 301.5 & 8.14 \\
500 & 0.06 & \shortstack{\textbf{211.0}\\\gd{\tiny $\uparrow$81.8}} & \shortstack{\textbf{230.2}\\\gd{\tiny $\uparrow$71.3}} & \shortstack{9.12\\\rd{\tiny $\downarrow$0.98}} \\
\rowcolor{rowA}
1000 & 0.12 & \shortstack{215.4\\\gd{\tiny $\uparrow$77.4}} & \shortstack{239.5\\\gd{\tiny $\uparrow$62.0}} & \shortstack{9.04\\\rd{\tiny $\downarrow$0.90}} \\
2000 & 0.24 & \shortstack{224.4\\\gd{\tiny $\uparrow$68.4}} & \shortstack{248.4\\\gd{\tiny $\uparrow$53.1}} & \shortstack{8.61\\\rd{\tiny $\downarrow$0.47}} \\
\rowcolor{rowA}
5000 & 0.61 & \shortstack{276.4\\\gd{\tiny $\uparrow$16.4}} & \shortstack{286.7\\\gd{\tiny $\uparrow$14.8}} & \shortstack{\textbf{8.11}\\\gd{\tiny $\uparrow$0.03}} \\
\bottomrule
\end{tabular}
\end{table*}


\subsection{ Code Generation }
\label{app:taco}

We further evaluate the Token Hypothesis on open-ended code generation using TACO~\cite{li2023taco}, 
which groups programming problems into five difficulty levels. We fine-tune LLaMA-3-8B 
and follow the evaluation protocol of the original benchmark, reporting Pass@1, Pass@10, and Pass@100 (\%) for each difficulty level together with training time and total GPU memory (Table~\ref{tab:taco}). Because TACO is a challenging benchmark, absolute scores are low across all configurations, and the comparison of interest is between values of $M$ under an identical protocol.

At $M{=}300$, Pass@1 matches or exceeds the full-sequence baseline at every difficulty level, with the largest gains on the medium-hard ($1.00$ versus $0.50$) and very hard ($0.46$ versus $0.36$) problems. Pass@100 follows the same pattern and improves most markedly on medium-hard problems ($12.55$ versus $6.24$). These results are obtained with $7.04$~GB less memory and a $24\%$ reduction in training time. $M{=}500$ performs comparably and achieves the best scores on the hard and very hard levels. In contrast, $M{=}100$ is insufficient, particularly on the easy and medium levels, where Pass@100 decreases by $4.29$ and $7.43$ points, respectively. This behavior is consistent with the long-context results in Appendix~\ref{app:longalpaca}: code solutions are long, and the required budget scales with the length of the supervised output.

\begin{table*}[ht!]
\centering\scriptsize
\setlength{\tabcolsep}{2.5pt}
\renewcommand{\arraystretch}{1.1}
\caption{Open-ended code generation on TACO (LLaMA-3-8B). Metrics: Pass@1, Pass@10, and Pass@100 (\%) per difficulty level; GPU is total GPU memory (GB); Time is total training time (min). \gd{$\uparrow$}~green = improvement over \textit{All},\ \rd{$\downarrow$}~red = degradation. \textbf{Bold} = best per column.}
\label{tab:taco}
\resizebox{\textwidth}{!}{%
\begin{tabular}{@{}lrrrrrrrrrrrrrrrrr@{}}
\toprule
\multirow{2}{*}{\textbf{Tokens}} & \multicolumn{3}{c}{\textbf{Easy}} & \multicolumn{3}{c}{\textbf{Medium}} & \multicolumn{3}{c}{\textbf{Medium-hard}} & \multicolumn{3}{c}{\textbf{Hard}} & \multicolumn{3}{c}{\textbf{Very hard}} & \multirow{2}{*}{\textit{Time}} & \multirow{2}{*}{\textit{GPU}} \\
\cmidrule(lr){2-4}\cmidrule(lr){5-7}\cmidrule(lr){8-10}\cmidrule(lr){11-13}\cmidrule(lr){14-16}
 & \textit{P@1} & \textit{P@10} & \textit{P@100} & \textit{P@1} & \textit{P@10} & \textit{P@100} & \textit{P@1} & \textit{P@10} & \textit{P@100} & \textit{P@1} & \textit{P@10} & \textit{P@100} & \textit{P@1} & \textit{P@10} & \textit{P@100} & & \\
\midrule
\rowcolor{basecol}
\textit{All} & 9.50 & 15.44 & 20.38 & \textbf{2.00} & \textbf{6.96} & \textbf{14.77} & 0.50 & 2.44 & 6.24 & 0.50 & 3.11 & 8.86 & 0.36 & 1.02 & 2.23 & 79.4 & 95.07 \\
100 & \shortstack{7.50\\\rd{\tiny $\downarrow$2.00}} & \shortstack{12.19\\\rd{\tiny $\downarrow$3.25}} & \shortstack{16.09\\\rd{\tiny $\downarrow$4.29}} & \shortstack{1.00\\\rd{\tiny $\downarrow$1.00}} & \shortstack{3.48\\\rd{\tiny $\downarrow$3.48}} & \shortstack{7.34\\\rd{\tiny $\downarrow$7.43}} & \shortstack{0.50\\\gd{\tiny $\uparrow$0.00}} & \shortstack{2.44\\\gd{\tiny $\uparrow$0.00}} & \shortstack{6.28\\\gd{\tiny $\uparrow$0.04}} & \shortstack{0.47\\\rd{\tiny $\downarrow$0.03}} & \shortstack{2.94\\\rd{\tiny $\downarrow$0.17}} & \shortstack{8.23\\\rd{\tiny $\downarrow$0.63}} & \shortstack{0.50\\\gd{\tiny $\uparrow$0.14}} & \shortstack{2.92\\\gd{\tiny $\uparrow$1.90}} & \shortstack{6.25\\\gd{\tiny $\uparrow$4.02}} & \shortstack{\textbf{52.9}\\\gd{\tiny $\uparrow$26.5}} & \shortstack{\textbf{85.00}\\\gd{\tiny $\uparrow$10.07}} \\
\rowcolor{rowA}
300 & \shortstack{\textbf{9.53}\\\gd{\tiny $\uparrow$0.03}} & \shortstack{\textbf{15.49}\\\gd{\tiny $\uparrow$0.05}} & \shortstack{\textbf{20.44}\\\gd{\tiny $\uparrow$0.06}} & \shortstack{\textbf{2.00}\\\gd{\tiny $\uparrow$0.00}} & \shortstack{\textbf{6.96}\\\gd{\tiny $\uparrow$0.00}} & \shortstack{14.71\\\rd{\tiny $\downarrow$0.06}} & \shortstack{\textbf{1.00}\\\gd{\tiny $\uparrow$0.50}} & \shortstack{\textbf{4.88}\\\gd{\tiny $\uparrow$2.44}} & \shortstack{\textbf{12.55}\\\gd{\tiny $\uparrow$6.31}} & \shortstack{0.50\\\gd{\tiny $\uparrow$0.00}} & \shortstack{3.11\\\gd{\tiny $\uparrow$0.00}} & \shortstack{8.89\\\gd{\tiny $\uparrow$0.03}} & \shortstack{0.46\\\gd{\tiny $\uparrow$0.10}} & \shortstack{2.68\\\gd{\tiny $\uparrow$1.66}} & \shortstack{5.75\\\gd{\tiny $\uparrow$3.52}} & \shortstack{60.1\\\gd{\tiny $\uparrow$19.3}} & \shortstack{88.03\\\gd{\tiny $\uparrow$7.04}} \\
500 & \shortstack{9.52\\\gd{\tiny $\uparrow$0.02}} & \shortstack{15.48\\\gd{\tiny $\uparrow$0.04}} & \shortstack{20.42\\\gd{\tiny $\uparrow$0.04}} & \shortstack{1.89\\\rd{\tiny $\downarrow$0.11}} & \shortstack{6.58\\\rd{\tiny $\downarrow$0.38}} & \shortstack{13.91\\\rd{\tiny $\downarrow$0.86}} & \shortstack{0.52\\\gd{\tiny $\uparrow$0.02}} & \shortstack{2.54\\\gd{\tiny $\uparrow$0.10}} & \shortstack{6.53\\\gd{\tiny $\uparrow$0.29}} & \shortstack{\textbf{0.51}\\\gd{\tiny $\uparrow$0.01}} & \shortstack{\textbf{3.18}\\\gd{\tiny $\uparrow$0.07}} & \shortstack{\textbf{9.06}\\\gd{\tiny $\uparrow$0.20}} & \shortstack{\textbf{0.51}\\\gd{\tiny $\uparrow$0.15}} & \shortstack{\textbf{2.98}\\\gd{\tiny $\uparrow$1.96}} & \shortstack{\textbf{6.38}\\\gd{\tiny $\uparrow$4.15}} & \shortstack{69.0\\\gd{\tiny $\uparrow$10.4}} & \shortstack{91.15\\\gd{\tiny $\uparrow$3.92}} \\
\bottomrule
\end{tabular}%
}
\end{table*}

\subsection{Compatibility with FlashAttention and QLoRA}
\label{app:fa-qlora}

The Token Hypothesis is implemented as a wrapper around per-token modules and leaves the attention operation unchanged, so it is compatible with any attention backend. All main experiments use the default scaled dot-product attention (SDPA) backend of Hugging Face Transformers, which already employs memory-efficient attention kernels; the reported baselines therefore include these savings. To verify compatibility with FlashAttention, we repeat the BoolQ experiment 
with \texttt{flash\_attention\_2} enabled for both the baseline and the Token Hypothesis (Table~\ref{tab:fa-qlora}, left). At $M{=}50$, the Token Hypothesis matches the baseline accuracy ($0.629$) while reducing memory by $15.17$~GB ($27\%$) and training time by $25\%$, and $M{=}100$ achieves the highest accuracy ($0.643$). The two techniques address different components of memory: FlashAttention reduces the memory associated with attention scores, whereas the Token Hypothesis reduces the per-token activations stored for backpropagation.

We also combine the Token Hypothesis with QLoRA~\cite{dettmers2023qlora}, 
which applies 4-bit quantization to the base model, on ARC-Challenge with LLaMA-3-8B under the same setup as the main experiments (Table~\ref{tab:fa-qlora}, right). At $M{=}100$, accuracy slightly exceeds the baseline ($0.688$ versus $0.683$) with $10.49$~GB less memory, and $M{=}300$ achieves the same accuracy. Because quantization reduces the memory occupied by model weights, activations account for a larger share of the remaining footprint, which is precisely the component that the Token Hypothesis reduces.

\begin{table*}[ht!]
\centering\scriptsize
\setlength{\tabcolsep}{2.5pt}
\renewcommand{\arraystretch}{1.1}
\caption{Compatibility with FlashAttention and QLoRA. \textit{Left:} BoolQ with \texttt{flash\_attention\_2} enabled for all rows. \textit{Right:} ARC-Challenge with QLoRA (LLaMA-3-8B). GPU is total GPU memory (GB); Time is total training time (min). \gd{$\uparrow$}~green = improvement over \textit{All},\ \rd{$\downarrow$}~red = degradation. \textbf{Bold} = best per column.}
\label{tab:fa-qlora}
\begin{minipage}[t]{0.45\textwidth}
\centering
\begin{tabular}{@{}lrrr@{}}
\toprule
\multirow{2}{*}{\textbf{Tokens}} & \multicolumn{3}{c}{\textbf{FlashAttention-2 (BoolQ)}} \\
\cmidrule(lr){2-4}
 & \textit{GPU} & \textit{Acc} & \textit{Time} \\
\midrule
\rowcolor{basecol}
\textit{All} & 56.67 & 0.629 & 7.99 \\
10 & \shortstack{\textbf{40.85}\\\gd{\tiny $\uparrow$15.82}} & \shortstack{0.621\\\rd{\tiny $\downarrow$0.008}} & \shortstack{\textbf{5.81}\\\gd{\tiny $\uparrow$2.18}} \\
\rowcolor{rowA}
50 & \shortstack{41.50\\\gd{\tiny $\uparrow$15.17}} & \shortstack{0.629\\\gd{\tiny $\uparrow$0.000}} & \shortstack{5.97\\\gd{\tiny $\uparrow$2.02}} \\
100 & \shortstack{42.29\\\gd{\tiny $\uparrow$14.38}} & \shortstack{\textbf{0.643}\\\gd{\tiny $\uparrow$0.014}} & \shortstack{6.14\\\gd{\tiny $\uparrow$1.85}} \\
\rowcolor{rowA}
300 & \shortstack{44.46\\\gd{\tiny $\uparrow$12.21}} & \shortstack{0.623\\\rd{\tiny $\downarrow$0.006}} & \shortstack{6.97\\\gd{\tiny $\uparrow$1.02}} \\
500 & \shortstack{48.64\\\gd{\tiny $\uparrow$8.03}} & \shortstack{0.633\\\gd{\tiny $\uparrow$0.004}} & \shortstack{7.74\\\gd{\tiny $\uparrow$0.25}} \\
\bottomrule
\end{tabular}
\end{minipage}\hfill
\begin{minipage}[t]{0.45\textwidth}
\centering
\begin{tabular}{@{}lrrr@{}}
\toprule
\multirow{2}{*}{\textbf{Tokens}} & \multicolumn{3}{c}{\textbf{QLoRA (ARC-Challenge)}} \\
\cmidrule(lr){2-4}
 & \textit{GPU} & \textit{Acc} & \textit{Time} \\
\midrule
\rowcolor{basecol}
\textit{All} & 65.19 & 0.683 & 4.00 \\
10 & \shortstack{\textbf{53.65}\\\gd{\tiny $\uparrow$11.54}} & \shortstack{0.619\\\rd{\tiny $\downarrow$0.064}} & \shortstack{\textbf{3.60}\\\gd{\tiny $\uparrow$0.40}} \\
\rowcolor{rowA}
25 & \shortstack{53.83\\\gd{\tiny $\uparrow$11.36}} & \shortstack{0.651\\\rd{\tiny $\downarrow$0.032}} & \shortstack{3.62\\\gd{\tiny $\uparrow$0.38}} \\
50 & \shortstack{54.12\\\gd{\tiny $\uparrow$11.07}} & \shortstack{0.653\\\rd{\tiny $\downarrow$0.030}} & \shortstack{3.63\\\gd{\tiny $\uparrow$0.37}} \\
\rowcolor{rowA}
100 & \shortstack{54.70\\\gd{\tiny $\uparrow$10.49}} & \shortstack{\textbf{0.688}\\\gd{\tiny $\uparrow$0.005}} & \shortstack{3.65\\\gd{\tiny $\uparrow$0.35}} \\
300 & \shortstack{57.02\\\gd{\tiny $\uparrow$8.17}} & \shortstack{\textbf{0.688}\\\gd{\tiny $\uparrow$0.005}} & \shortstack{3.72\\\gd{\tiny $\uparrow$0.28}} \\
\bottomrule
\end{tabular}
\end{minipage}
\end{table*}

\subsection{Summary}

Across six modalities, five architectures, and three training regimes, the picture is consistent. The optimal $M$ varies by task and architecture---$M = 50$--$100$ for bidirectional GLUE and image classification, $M = 100$--$300$ for causal reasoning and visual QA, $M = 300$--$700$ for math and long-context video---but in every case the improvement or match to the baseline comes at a memory cost that is substantially lower than full-sequence fine-tuning. The exact savings depend on sequence length and architecture: for causal transformers with long video inputs the saving is on the order of $10$--$30$~GB; for bidirectional models operating on short sequences the saving is modest in absolute terms but matches or exceeds the baseline on every task. These per-benchmark results collectively confirm that the optimal $M^{\ast}$ predicted by Corollary~\ref{cor:optimal-M} is achievable across real training runs, and that accuracy parity with full fine-tuning is attainable at significantly lower memory cost.

\newpage

\begin{algorithm}[t]
\caption{Wrapper-based Token Hypothesis (one per-token module)}
\label{alg:wrapper}
\begin{algorithmic}[1]
\Require Input $X$, base module $f$ (a per-token op: Linear, Embedding, LayerNorm, RMSNorm)
\Require Number of trainable tokens $M$, adapter delta $\Delta(\cdot)$
\State $T_f \gets T - M$ \Comment{frozen segment length}
\State Split $X$ along its sequence axis: $X_f \gets X[:, {:}T_f], \; X_p \gets X[:, T_f{:}]$
\State \textbf{with} \texttt{torch.no\_grad()}\textbf{:}
\State \quad $Y_f \gets f(X_f)$ \Comment{frozen forward, no activations retained}
\State $Y_p \gets f(X_p) + \Delta(X_p)$ \Comment{base + adapter, full gradient flow}
\State \Return $\textsc{concat}(Y_f, Y_p, \text{seq axis})$
\Statex
\Statex \textbf{Note:} The wrapper replaces each per-token module in the model. Method specialization is in $\Delta(\cdot)$: identity for full-FT, and PEFT such as $BA$ for LoRA, $\frac{m \cdot (W + BA)}{\|W + BA\|_{\text{col}}}$ for DoRA. Cross-token operations (attention scores, softmax) are not wrapped---they run on the rejoined full-length sequence with gradients enabled, leaving the $\mathcal{O}(B \cdot H \cdot T^2)$ attention-score term unaffected. This term can be recovered by composing with gradient checkpointing at $\sim$33\% additional compute per step.
\end{algorithmic}
\end{algorithm}

\begin{algorithm}[t]
\caption{Deep dual-path Token Hypothesis (collation and decoder layer)}
\label{alg:dualpath}
\begin{algorithmic}[1]
\Statex \textbf{Stage 1 --- Data collation (per sample, once)}
\Require Input ids $\mathbf{x} \in \mathbb{Z}^T$, labels $\mathbf{y} \in \mathbb{Z}^T$, target $M$, mode \texttt{m} $\in \{$last, random, attention$\}$
\State \textbf{select} index set $\mathcal{S} \subset \{0, \ldots, T{-}1\}$ with $|\mathcal{S}| = M$:
\State \quad if \texttt{m = last}: \; $\mathcal{S} = \{T{-}M, \ldots, T{-}1\}$
\State \quad if \texttt{m = random}: \; $\mathcal{S} = \{0\} \cup \textsc{rand}(M{-}2) \cup \{T{-}1\}$ \Comment{boundary tokens fixed}
\State \quad if \texttt{m = attention}: \; $\mathcal{S} = \textsc{topk}_M\!\big(\textsc{importance}\big[\textsc{hash}(\mathbf{x})\big]\big)$ \Comment{precomputed token-importance map}
\State $\bar{\mathcal{S}} \gets \{0, \ldots, T{-}1\} \setminus \mathcal{S}$
\State $\mathbf{x}^{\mathcal{S}} \gets \mathbf{x}[\mathcal{S}]$, \quad $\mathbf{x}^{\bar{\mathcal{S}}} \gets \mathbf{x}[\bar{\mathcal{S}}]$
\State $\mathbf{y}^{\mathcal{S}} \gets \mathbf{y}[\mathcal{S} + 1]$ \Comment{shift baked in: position $t \in \mathcal{S}$'s logit predicts $\mathbf{y}_{t+1}$}
\State \Return $(\mathbf{x}^{\mathcal{S}}, \mathbf{x}^{\bar{\mathcal{S}}}, \mathbf{y}^{\mathcal{S}}, \text{positions})$

\Statex
\Statex \textbf{Stage 2 --- Decoder layer forward}
\Require Frozen stream $H_f \in \mathbb{R}^{B \times (T-M) \times d}$, trainable stream $H_p \in \mathbb{R}^{B \times M \times d}$
\Require Position ids $P_f, P_p$, attention masks $M_f, M_p$
\State \textbf{with} \texttt{torch.no\_grad()}\textbf{:}
\State \quad $\tilde{H}_f \gets \textsc{RMSNorm}(H_f)$
\State \quad $\mathbf{Q}_f, \mathbf{K}_f, \mathbf{V}_f \gets W_q\tilde{H}_f, W_k\tilde{H}_f, W_v\tilde{H}_f$
\State \quad $\mathbf{Q}_f, \mathbf{K}_f \gets \textsc{RoPE}(\mathbf{Q}_f, \mathbf{K}_f, P_f)$
\State $\tilde{H}_p \gets \textsc{RMSNorm}(H_p)$
\State $\mathbf{Q}_p, \mathbf{K}_p, \mathbf{V}_p \gets W_q\tilde{H}_p, W_k\tilde{H}_p, W_v\tilde{H}_p$
\State $\mathbf{Q}_p, \mathbf{K}_p \gets \textsc{RoPE}(\mathbf{Q}_p, \mathbf{K}_p, P_p)$
\State $\mathbf{K} \gets [\mathbf{K}_f \mid \mathbf{K}_p]$, \quad $\mathbf{V} \gets [\mathbf{V}_f \mid \mathbf{V}_p]$ \Comment{shared $\mathbf{K}/\mathbf{V}$ across both streams}
\State \textbf{with} \texttt{torch.no\_grad()}\textbf{:}
\State \quad $\mathbf{A}_f \gets \textsc{softmax}\!\big(\mathbf{Q}_f \mathbf{K}^\top / \sqrt{d} + M_f\big)$
\State \quad $H_f \gets H_f + W_o\,(\mathbf{A}_f \mathbf{V})$
\State \quad $H_f \gets H_f + \textsc{MLP}(\textsc{RMSNorm}(H_f))$
\State $\mathbf{A}_p \gets \textsc{softmax}\!\big(\mathbf{Q}_p \mathbf{K}^\top / \sqrt{d} + M_p\big)$
\State $H_p \gets H_p + W_o\,(\mathbf{A}_p \mathbf{V})$
\State $H_p \gets H_p + \textsc{MLP}(\textsc{RMSNorm}(H_p))$
\State \Return $(H_f, H_p)$
\Statex
\Statex \textbf{Note:} The trainable queries $\mathbf{Q}_p$ attend to $\mathbf{K}, \mathbf{V}$ from \emph{both} streams, preserving the full attention context. All frozen-stream tensors---$\mathbf{A}_f$, softmax outputs, residuals, RMSNorm activations, MLP intermediates---are kept under \texttt{no\_grad}, eliminating the dominant $\mathcal{O}(B \cdot H \cdot T^2)$ memory term without recomputation. Because the label shift in Stage~1 is baked into $\mathbf{y}^{\mathcal{S}}$, the loss is a plain cross-entropy on $H_p \to \mathbf{y}^{\mathcal{S}}$ at the LM head with no further offset.
\end{algorithmic}
\end{algorithm}

\section{Implementation Details}

We implement the Token Hypothesis through two complementary strategies that differ in how aggressively they confine the frozen-token path to the no-gradient regime. In what follows, $M$ denotes the number of trainable positions and $\mathcal{S}, \bar{\mathcal{S}}$ denote the trainable and frozen index sets, matching the notation introduced in Section~\ref{sec:method}. Within the algorithm pseudocode, $\mathbf{K}$ and $\mathbf{V}$ (boldface) refer to the attention keys and values, distinct from the Gram matrices $K_t$ of our theoretical analysis.

The \emph{wrapper-based} approach (Algorithm~\ref{alg:wrapper}) replaces every per-token module in the model---linear projections, token embeddings, RMSNorm, and LayerNorm---with a thin wrapper that splits its input along the sequence dimension into trainable and frozen segments, runs the frozen segment under \texttt{torch.no\_grad()}, runs the trainable segment with the method-specific delta function, and rejoins the outputs. The wrapper applies cleanly to any per-token operation because each output position depends only on the corresponding input position. This strategy is architecture-agnostic and supports a uniform interface across full fine-tuning, LoRA, LoRA-FA, DoRA, and QLoRA: each method is realized by changing only the delta $\Delta(\cdot)$ inside the wrapper. It preserves the original attention implementation, retaining compatibility with SDPA and FlashAttention. Cross-token operations such as attention scores cannot be split this way, however, so the dominant $\mathcal{O}(B \cdot H \cdot T^2)$ activation memory term remains on the gradient-bearing path.

The \emph{deep dual-path} approach (Algorithm~\ref{alg:dualpath}) instead splits the sequence once during data collation and maintains two parallel hidden-state streams throughout the entire forward pass. Following prior work on token-selective fine-tuning~\cite{simoulin2024memory, zeng2026tokenseek}, the collator selects $M$ trainable positions per sample using one of three strategies---the last $M$ tokens, $M$ random positions plus the boundary tokens, or top-$M$ by precomputed attention--gradient importance scores keyed on the input hash---and bakes the next-token label shift into the prefix labels at this stage, so the cross-entropy loss requires no further offset. Inside each decoder layer, the trainable stream's queries attend to keys and values concatenated from both streams, preserving the full attention context, while every frozen-stream intermediate---attention scores, softmax outputs, residuals, RMSNorm activations, MLP intermediates---is kept under \texttt{no\_grad}. This eliminates the dominant attention-score memory term without recomputation, yielding substantially larger savings at small $M$. The trade-off is that dual-path attention does not compose with fused kernels and requires eager attention.

We use the deep dual-path approach for our main fine-tuning experiments on LLaMA-3-8B and LLaMA-3.2-3B, where attention-score memory dominates and the savings are largest, and use the wrapper-based approach for cross-method comparisons with LoRA, DoRA, and QLoRA and for vision-language backbones, where compatibility with fused-attention kernels and a uniform PEFT interface are more important than maximizing attention-memory savings. For our head-to-head comparisons with TokenTune~\cite{simoulin2024memory} and TokenSeek~\cite{zeng2026tokenseek} (Section~\ref{sec:experiments}), we follow the same training protocol as these prior works to ensure fair comparison: the deep dual-path implementation matches their two-stream design while the position-selection rule is determined by our theoretical results (last-$M$ for causal, random for bidirectional) rather than per-sample heuristic scoring. All experiments use bf16 mixed precision and the AdamW optimizer with a cosine learning-rate schedule; multi-GPU configurations shard the model across devices using HuggingFace \texttt{device\_map="auto"}, as described in Appendix~H.

\section{Limitations}
\label{app:limitations}

Our analysis has three main limitations. First, the framework relies on the lazy NTK regime, which requires parameter updates to remain close to initialization. This holds for fine-tuning strong pretrained models---we measure mean weight drift below 0.011 across every parameter group of LLaMA-3-8B (Appendix~\ref{ab:laziness})---but is expected to break under large representational shifts, such as adapting a model to a substantially new domain, where the linearization around $\theta_0$ no longer captures training dynamics and the per-position capacity bound need not control the achievable predictor. Second, all of our theoretical and empirical results assume a strong pretrained initialization; the behavior of token-wise capacity during training from scratch is not analyzed in this work and is not implied by our results. Third, our analysis is restricted to transformer architectures with standard attention, both bidirectional and causal. Other architectural families---state-space models, mixture-of-experts, and recurrent or convolutional sequence models---fall outside the scope of Theorems~2 and~3, whose proofs rely on properties of attention-mediated information flow, and would require a separate analysis.


\end{document}